\documentclass[10pt,twocolumn,letterpaper]{article}

\usepackage[pagenumbers]{cvpr} % To force page numbers, e.g. for an arXiv version
\usepackage{graphicx}
\usepackage{xcolor}    % for colored “Good/Bad” text (optional)
\usepackage{colortbl}  % for \cellcolor

\definecolor{cvprblue}{rgb}{0.21,0.49,0.74}
\usepackage[pagebackref,breaklinks,colorlinks,allcolors=cvprblue]{hyperref}

\def\paperID{15906} % Paper ID here
\def\confName{CVPR}
\def\confYear{2026}

\newcommand{\sysnamenospace}{\text{VideoScience}}

\usepackage{minted}
\usepackage{listings}
\usepackage[most]{tcolorbox}
\title{
Benchmarking Scientific Understanding and Reasoning for Video Generation using \sysnamenospace-Bench
}

\author{
\textbf{Lanxiang Hu} \quad
\textbf{Abhilash Shankarampeta} \quad
\textbf{Yixin Huang} \quad
\textbf{Zilin Dai} \quad
\textbf{Haoyang Yu} \\
\textbf{Yujie Zhao} \quad
\textbf{Haoqiang Kang} \quad
\textbf{Daniel Zhao} \quad
\textbf{Tajana Rosing} \quad
\textbf{Hao Zhang} \\
University of California, San Diego
}

\begin{document}

\twocolumn[{%
\renewcommand\twocolumn[1][]{#1}%
\maketitle

\vspace{-2.5em}

    \begin{figure}[H]
        \centering
    
        % ================= Row 1: Teaser =================
        \begin{minipage}{0.97\textwidth}
            \small
            
            % First scenario

            % left panel
            \noindent
            \begin{minipage}[t]{0.25\textwidth}
              \vspace{0pt}
              \raggedright
              \textbf{Physical Commonsense:}\\[0.2em]
              {\color{red!70!black}\textit{An inflated balloon is brought close to the flame of a lit candle.}}
            \end{minipage}%
            \hfill
            % right panel
            \begin{minipage}[t]{0.72\textwidth}
              \vspace{0pt}
              \centering
              \includegraphics[height=1.3cm]{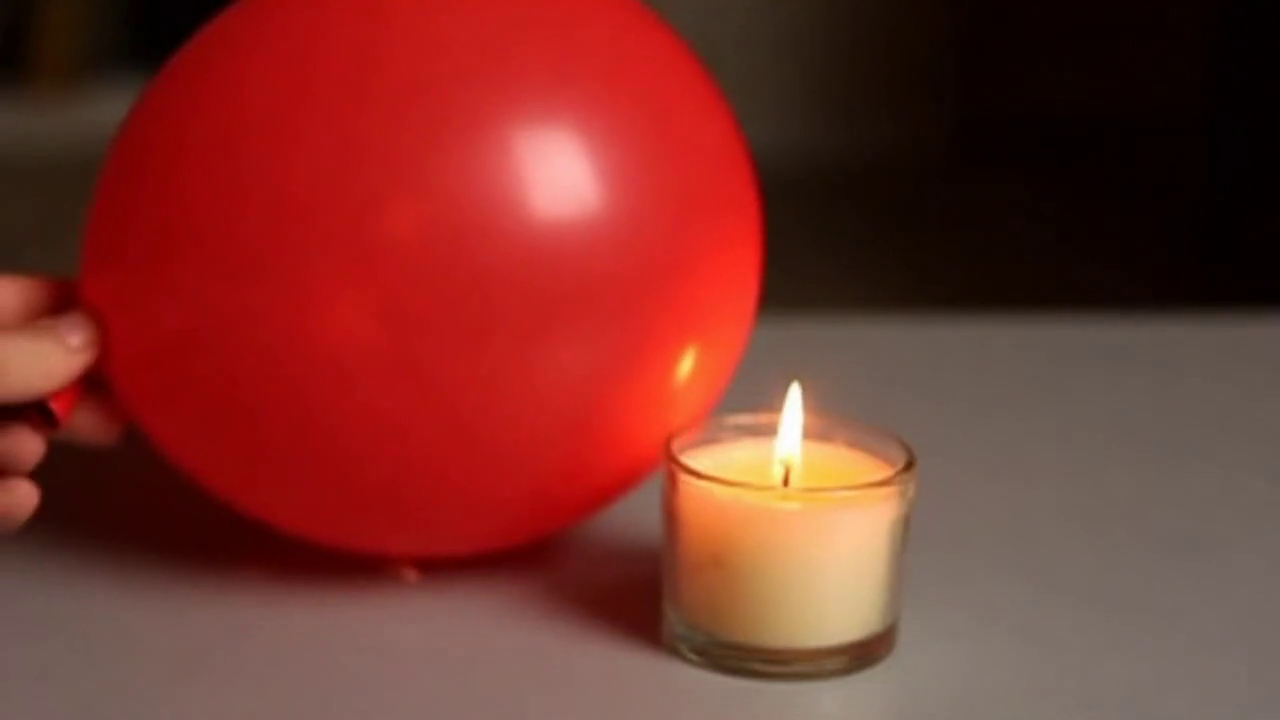}\hspace{1pt}%
              \includegraphics[height=1.3cm]{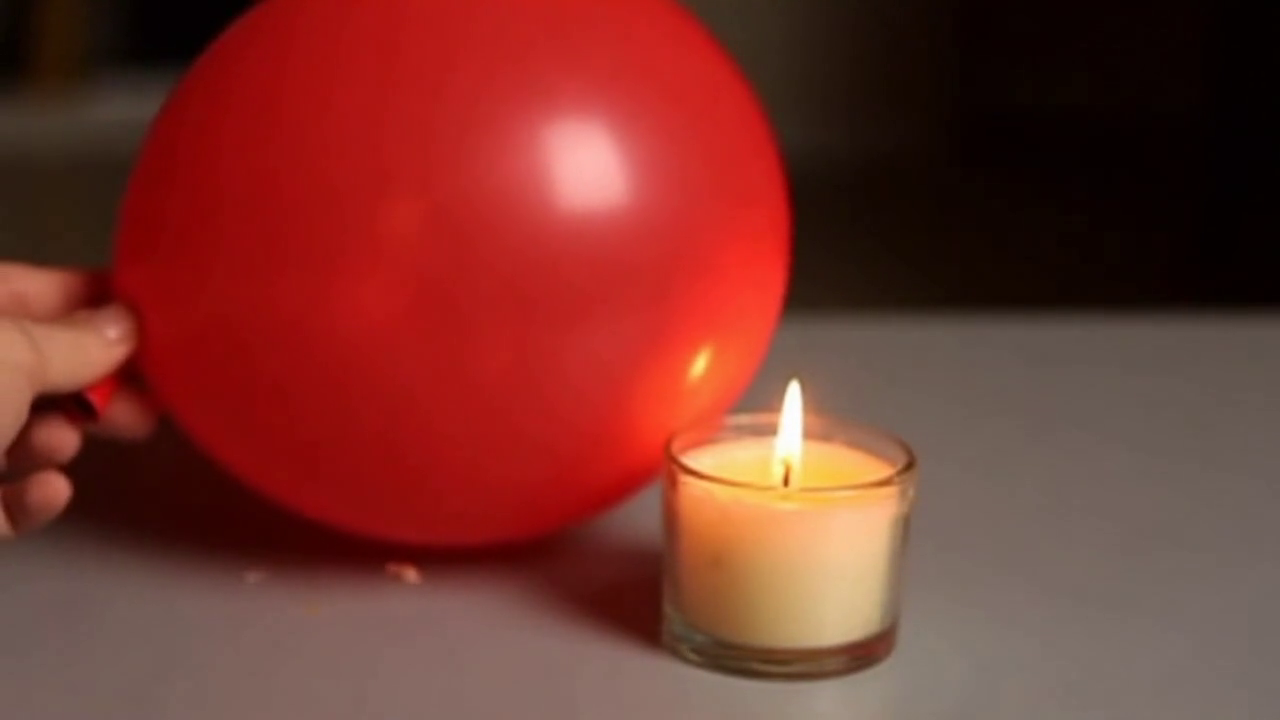}\hspace{1pt}%
              \includegraphics[height=1.3cm]{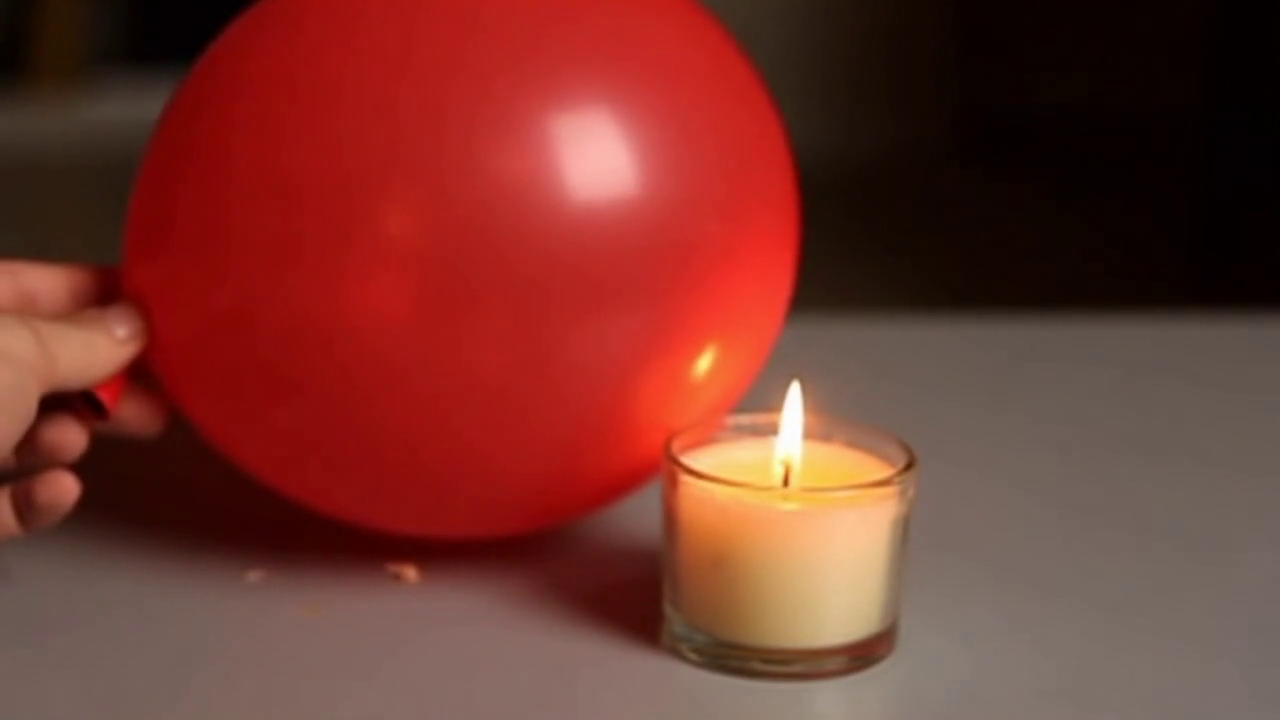}\hspace{1pt}%
              \includegraphics[height=1.3cm]{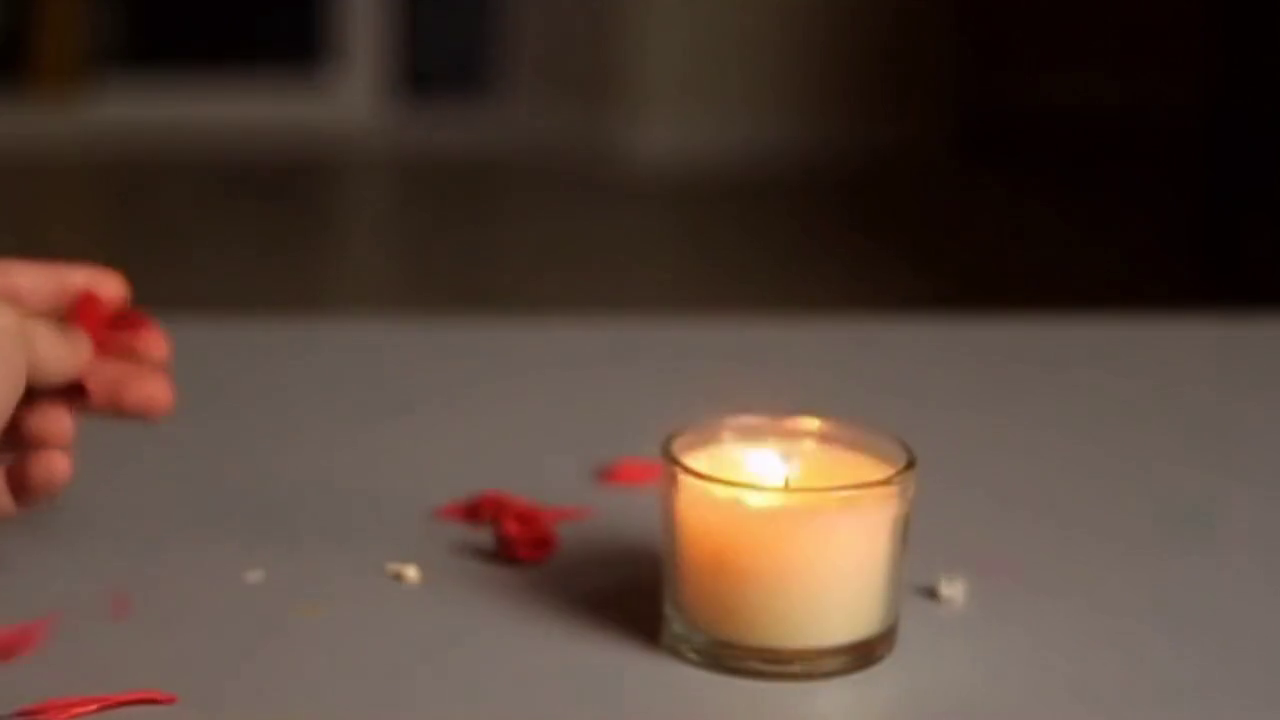}\hspace{1pt}%
              \includegraphics[height=1.3cm]{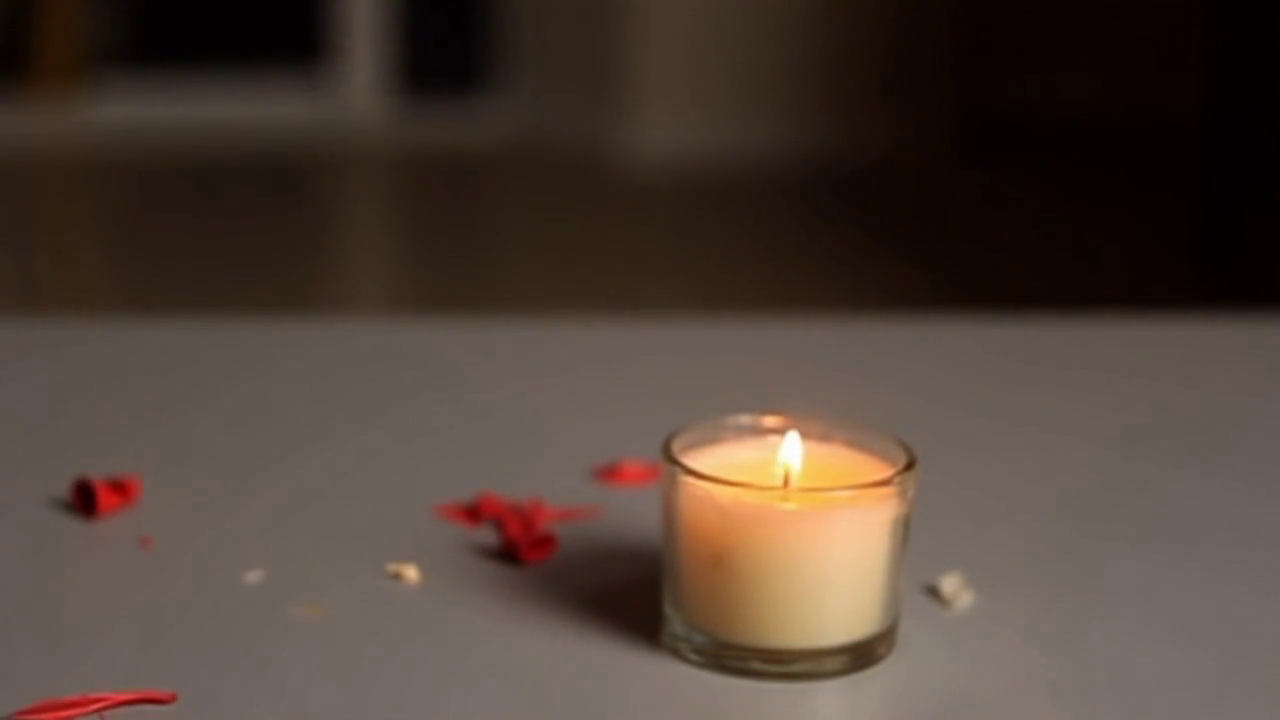}%
            \end{minipage}
            
            \vspace{0.5em}
            
% A small metal ball is dropped through a clear plastic tube, accelerating smoothly and shooting out the bottom almost immediately

% A visually identical neodymium magnet is dropped through a vertical copper pipe, sinking in eerie slow motion as invisible eddy currents in the metal dramatically brake its fall.

            % Second scenario
            \begin{minipage}[t]{0.25\textwidth}
              \vspace{0pt}
              \raggedright
              {\color{red!70!black}\textit{A transparent tube carrying a steady stream of water pours out of the tube.}}
            \end{minipage}%
            \hfill
            % right panel
            \begin{minipage}[t]{0.72\textwidth}
              \vspace{0pt}
              \centering
              \includegraphics[height=1.3cm]{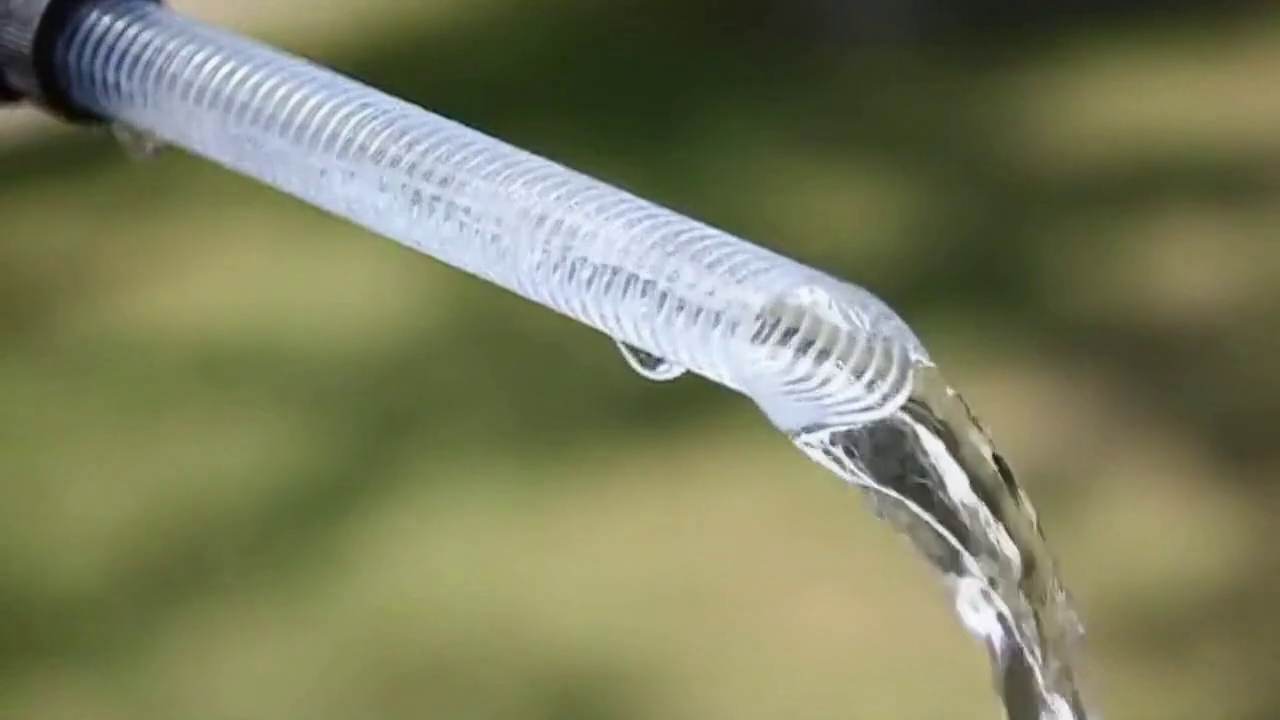}\hspace{1pt}%
              \includegraphics[height=1.3cm]{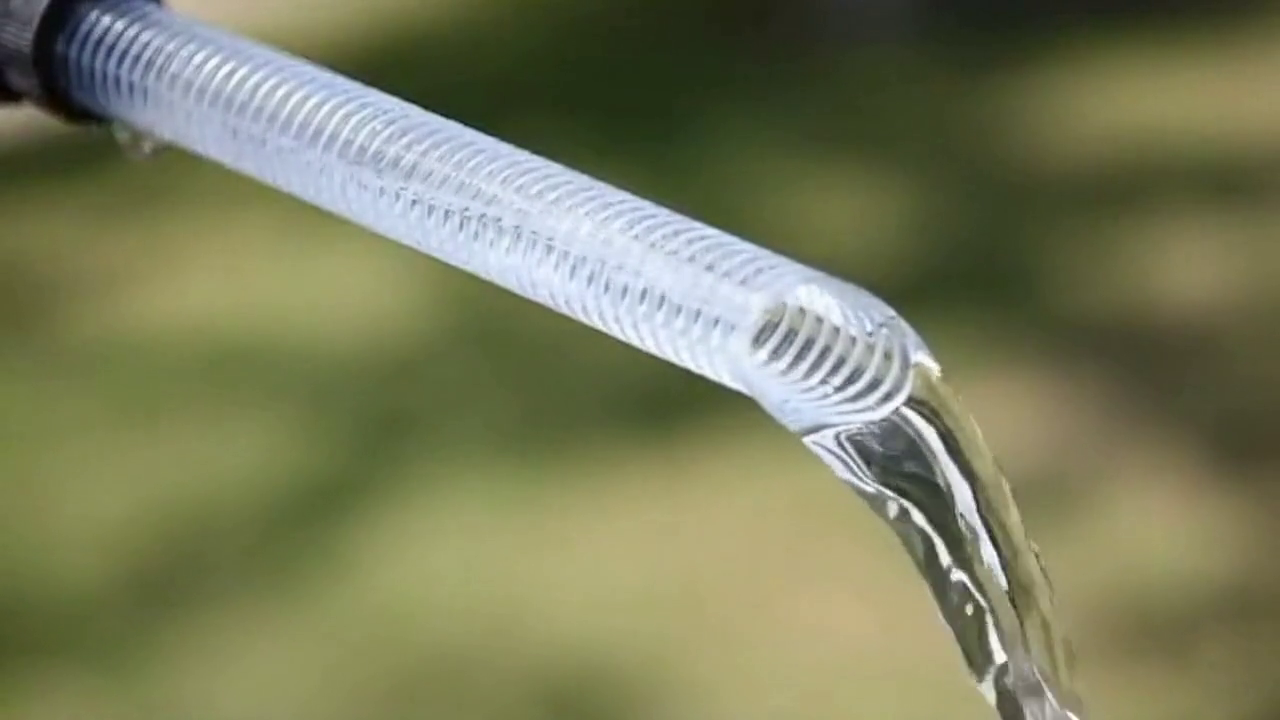}\hspace{1pt}%
              \includegraphics[height=1.3cm]{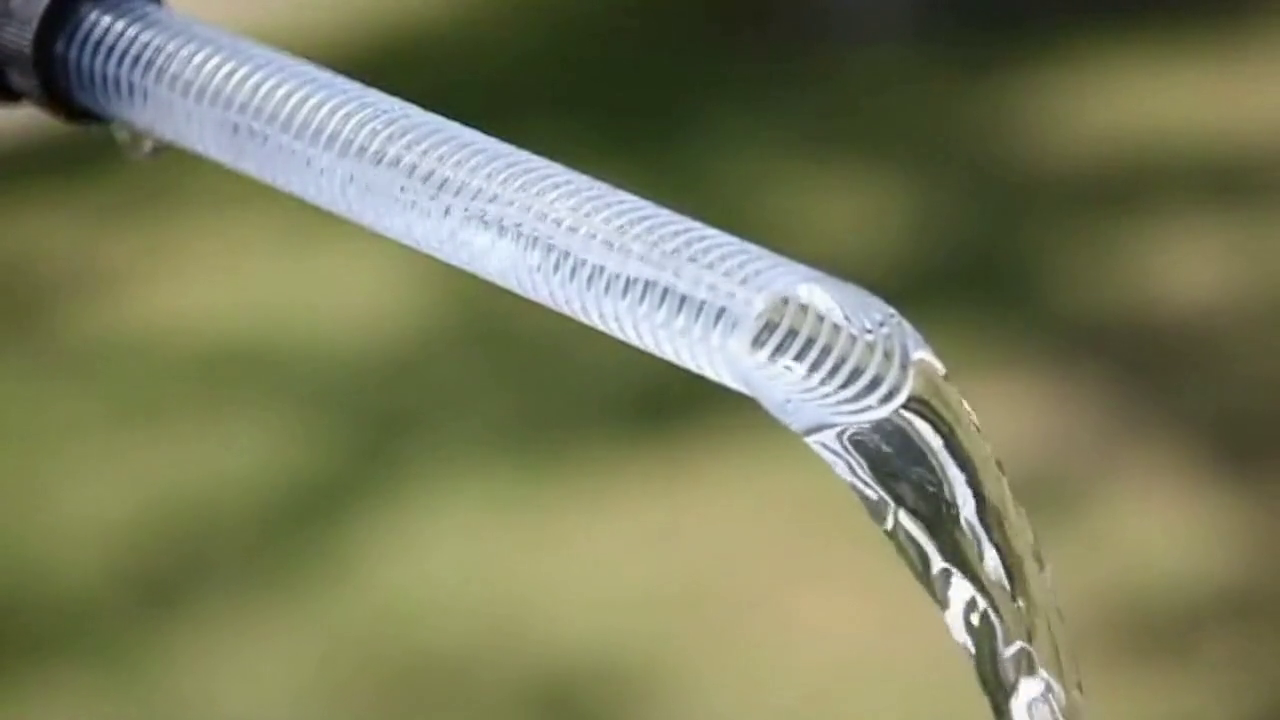}\hspace{1pt}%
              \includegraphics[height=1.3cm]{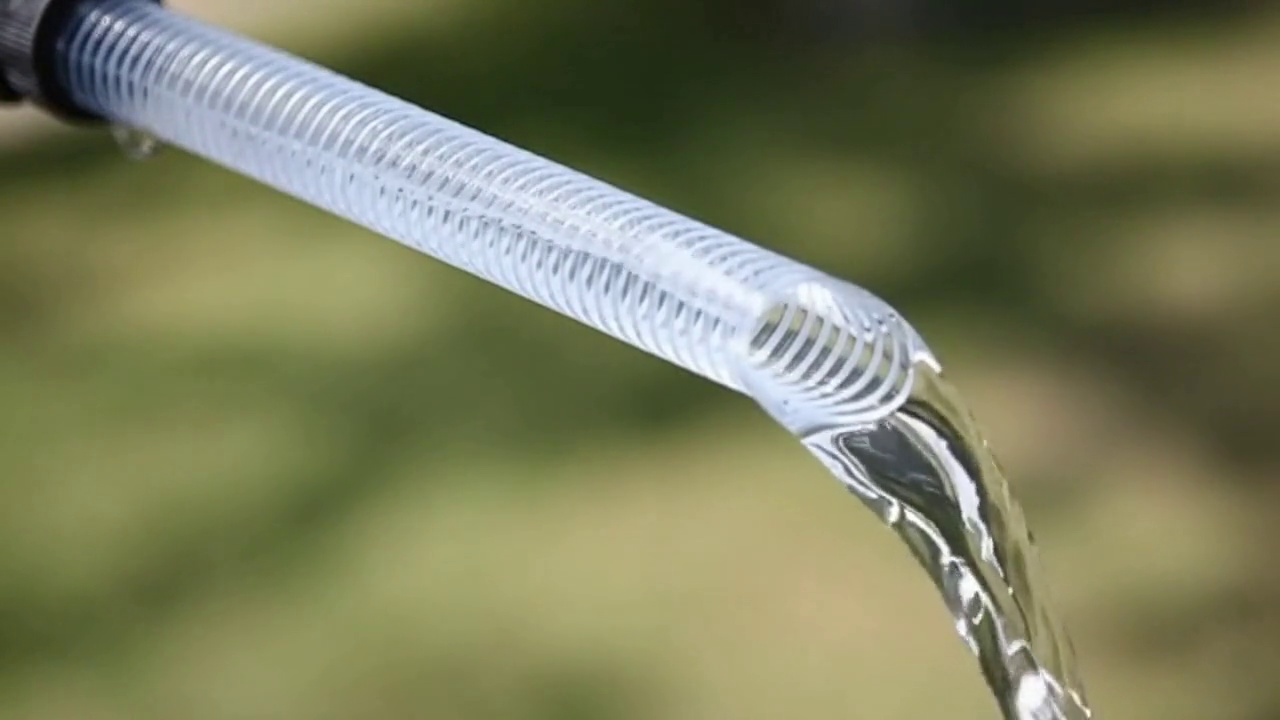}\hspace{1pt}%
              \includegraphics[height=1.3cm]{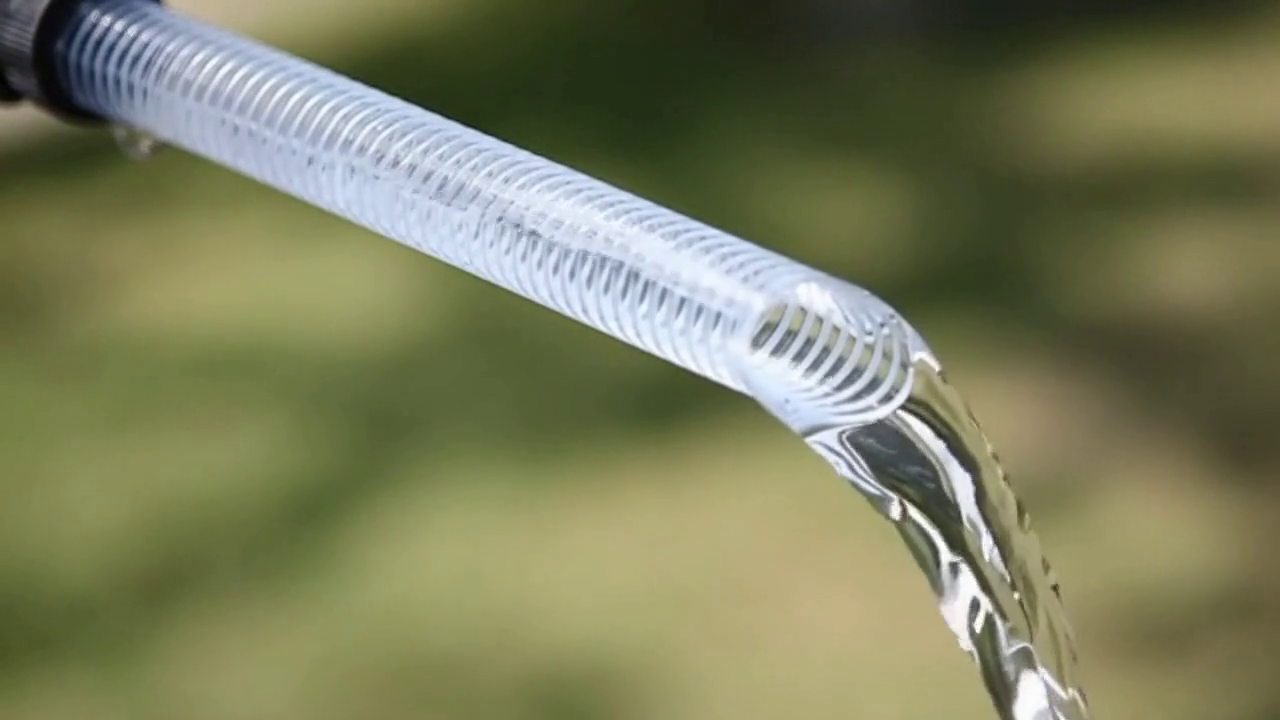}%
            \end{minipage}
            
            \vspace{0.5em}
            
            % Third scenario
            \noindent
            \begin{minipage}[t]{0.25\textwidth}
              \vspace{0pt}
              \raggedright
              \textbf{Scientific Reasoning:}\\[0.2em]
              {\color{green!60!black}\textit{An inflated balloon \underline{\smash{filled with a}} \underline{\smash{small amount of water}} is brought close to the flame of a lit candle.}}
            \end{minipage}%
            \hfill
            \begin{minipage}[t]{0.72\textwidth}
              \vspace{0pt}
              \centering
              \includegraphics[height=1.3cm]{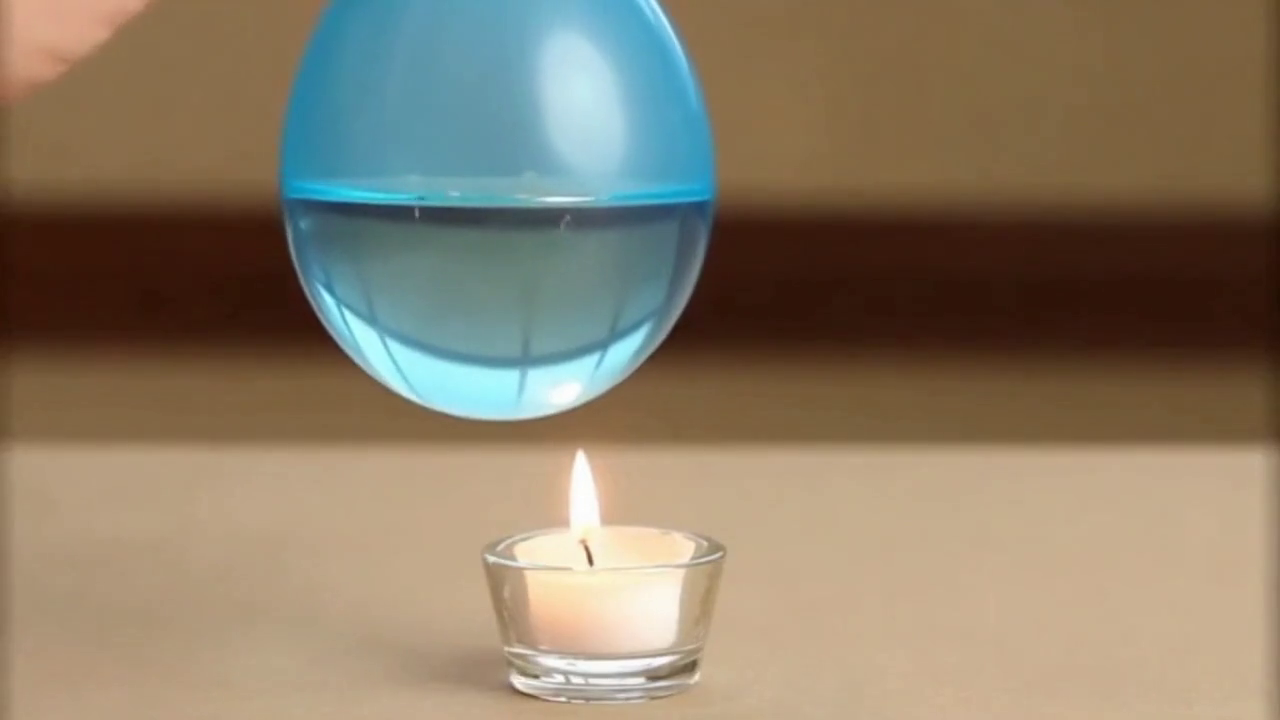}\hspace{1pt}%
              \includegraphics[height=1.3cm]{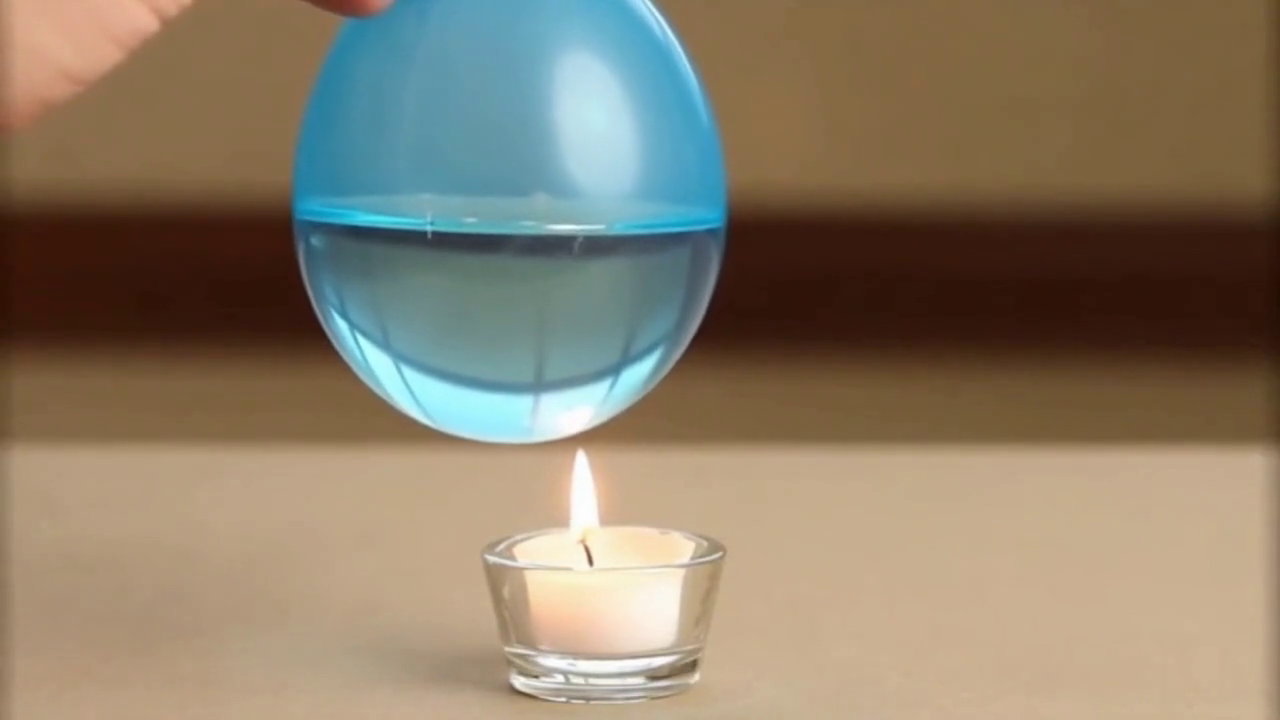}\hspace{1pt}%
              \includegraphics[height=1.3cm]{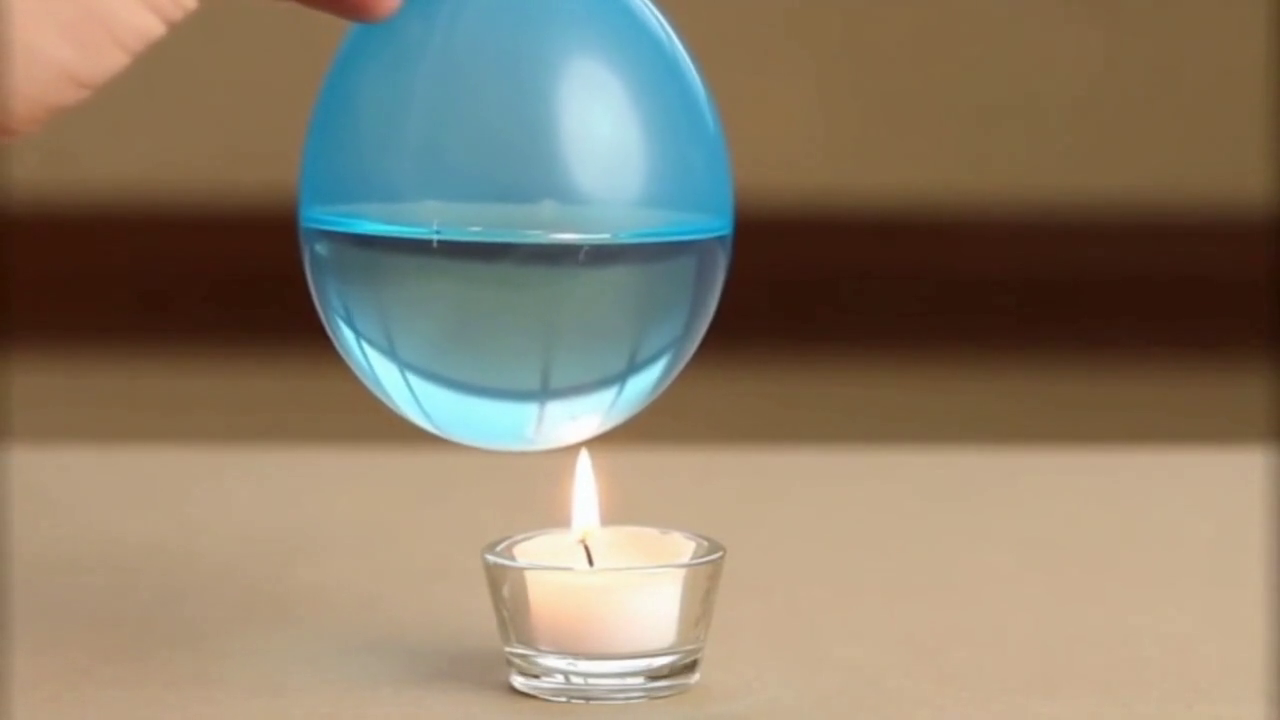}\hspace{1pt}%
              \includegraphics[height=1.3cm]{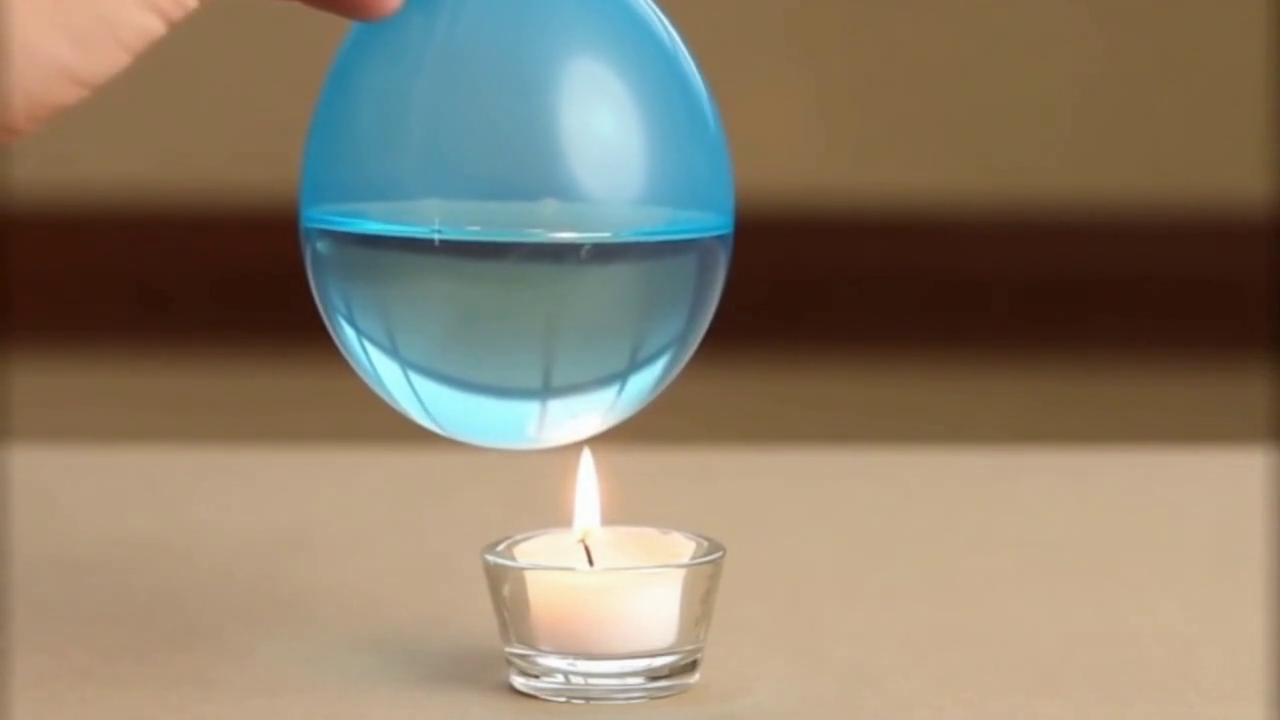}\hspace{1pt}%
              \includegraphics[height=1.3cm]{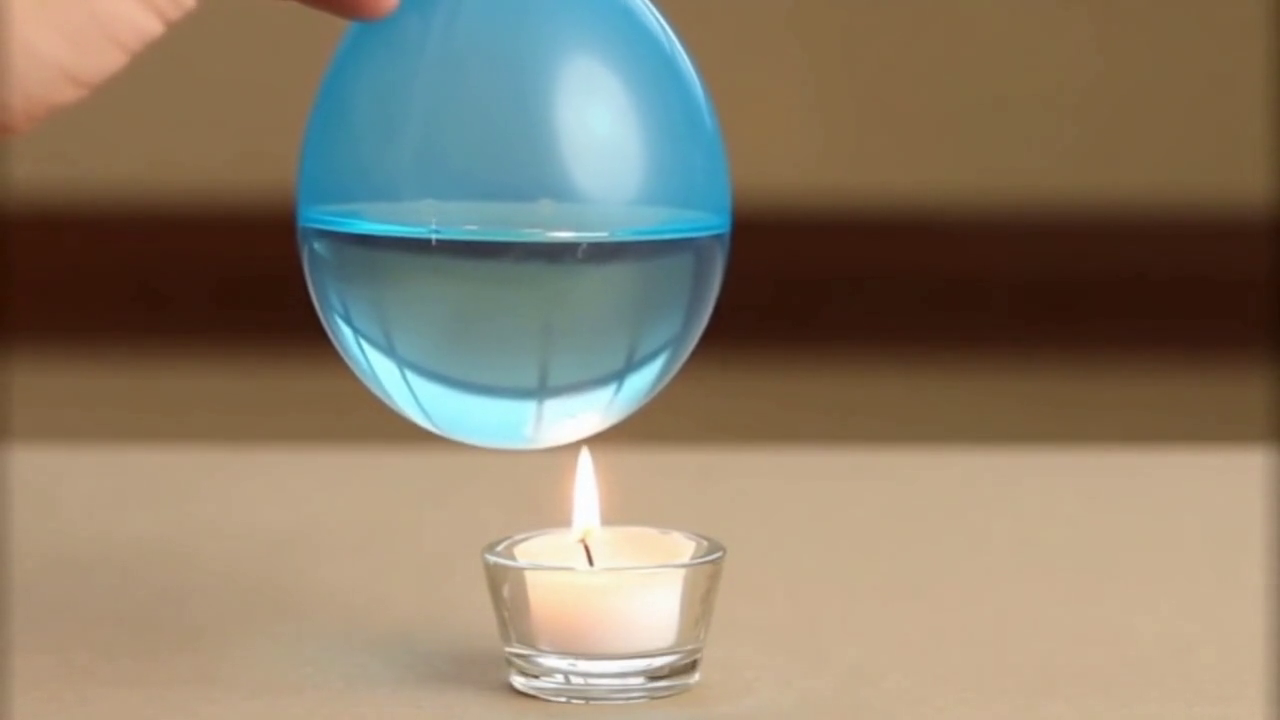}%
            \end{minipage}

            \vspace{0.5em}
            
            % Fourth scenario
            \begin{minipage}[t]{0.25\textwidth}
              \vspace{0pt}
              \raggedright
              {\color{green!60!black}\textit{A transparent tube carrying a steady stream of water is taped to the front of \underline{\smash{a speaker playing}} \underline{\smash{a low-frequency sound}}.}}
            \end{minipage}%
            \hfill
            % right panel
            \begin{minipage}[t]{0.72\textwidth}
              \vspace{0pt}
              \centering
              \includegraphics[height=1.3cm]{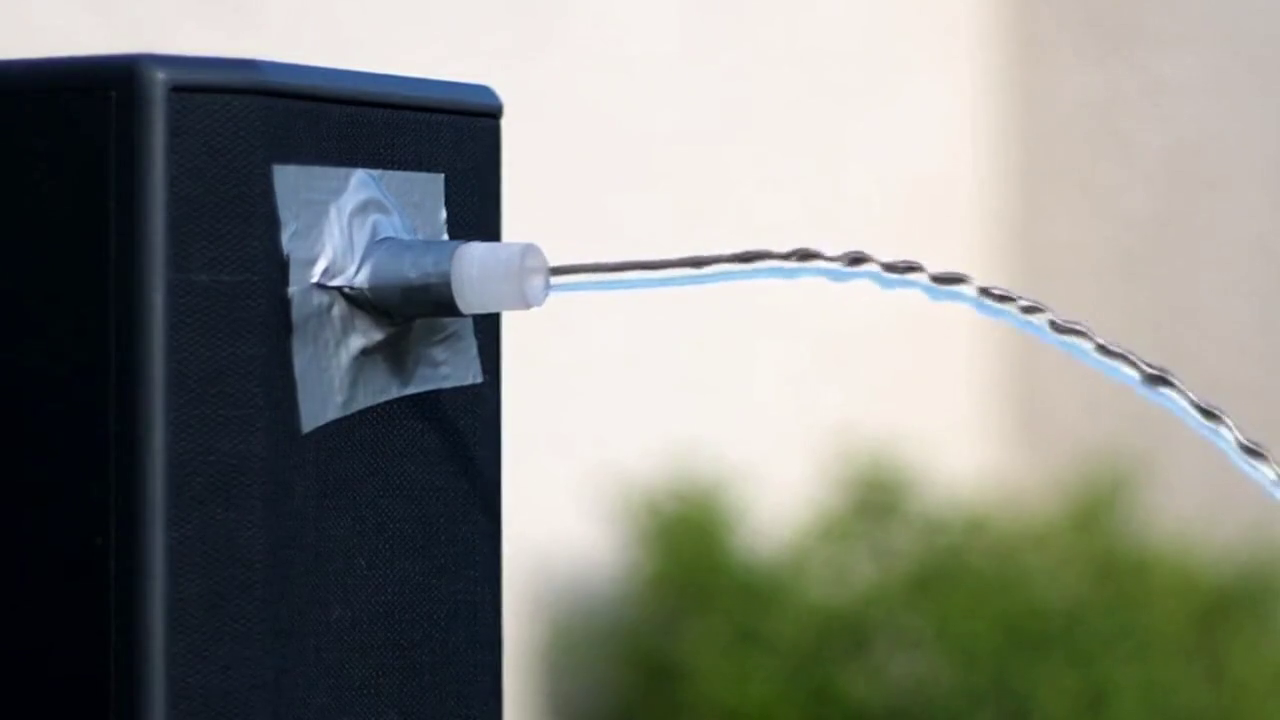}\hspace{1pt}%
              \includegraphics[height=1.3cm]{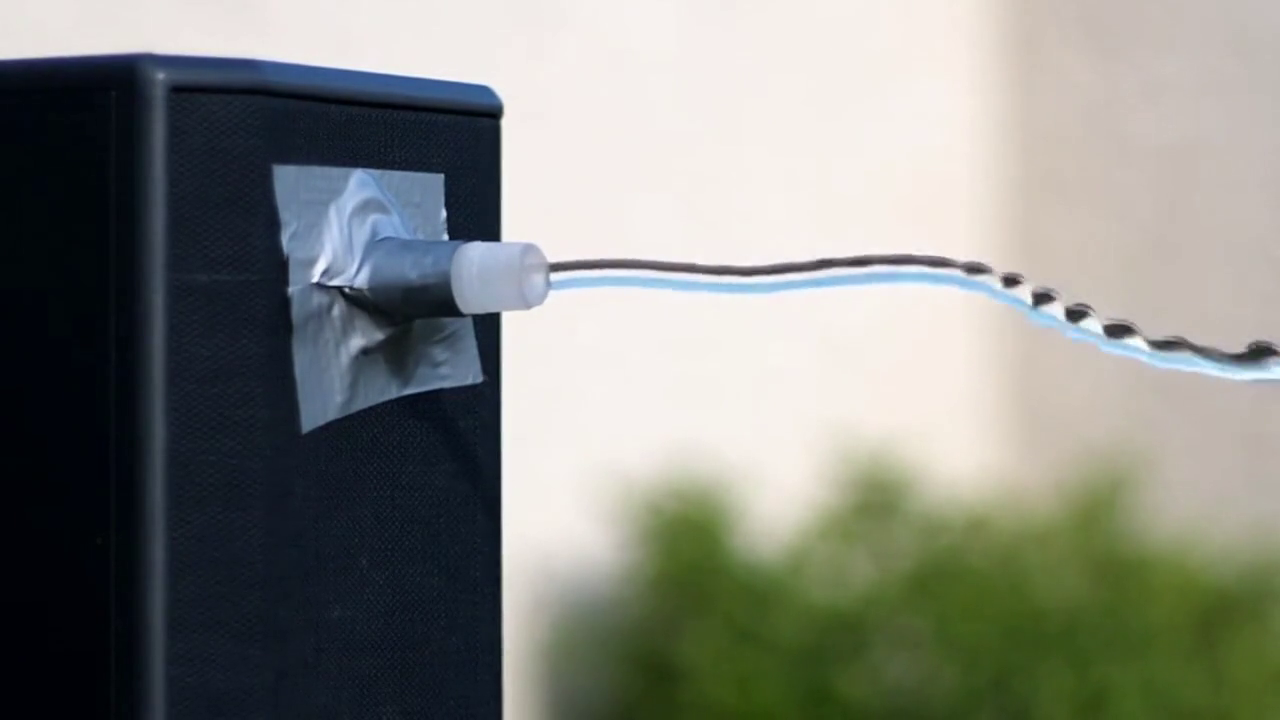}\hspace{1pt}%
              \includegraphics[height=1.3cm]{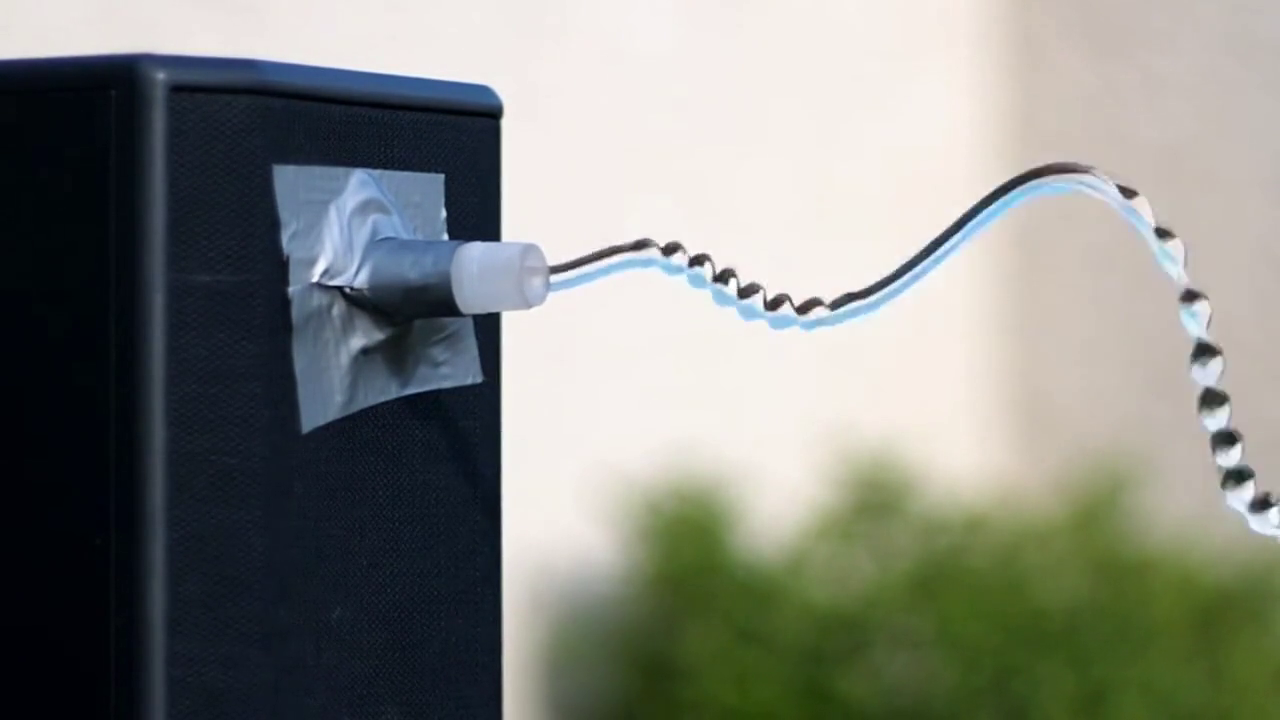}\hspace{1pt}%
              \includegraphics[height=1.3cm]{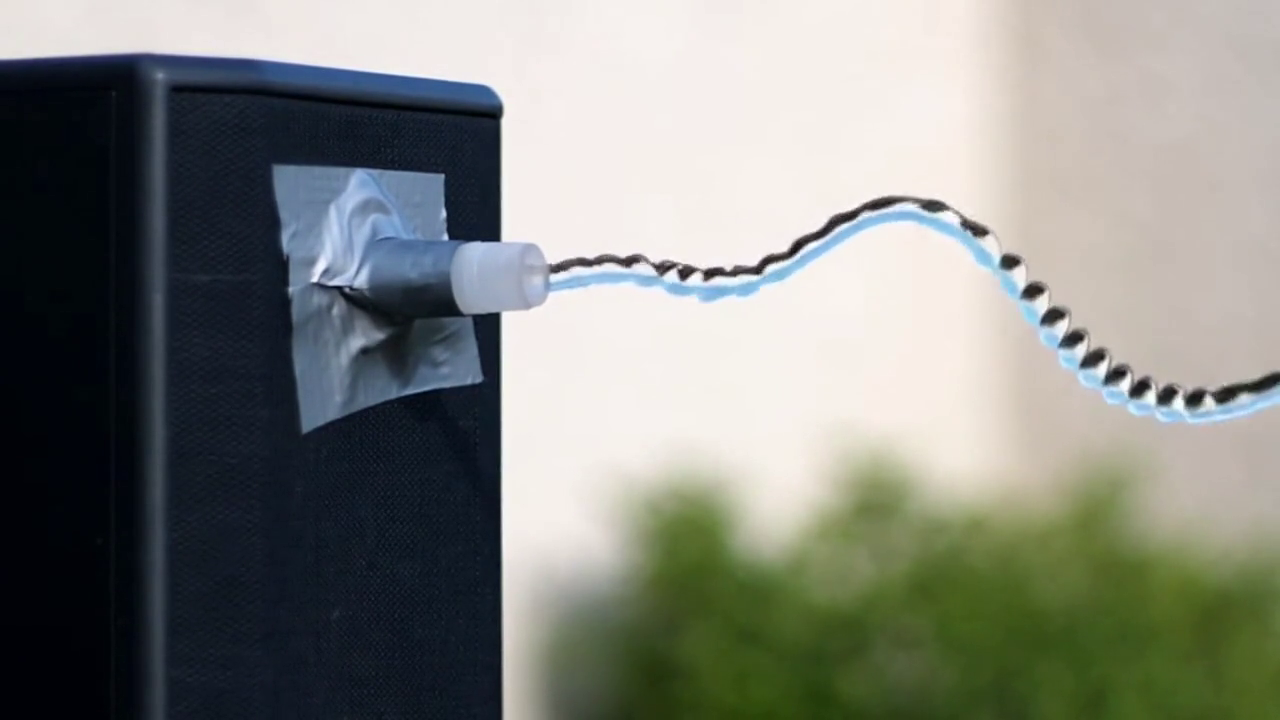}\hspace{1pt}%
              \includegraphics[height=1.3cm]{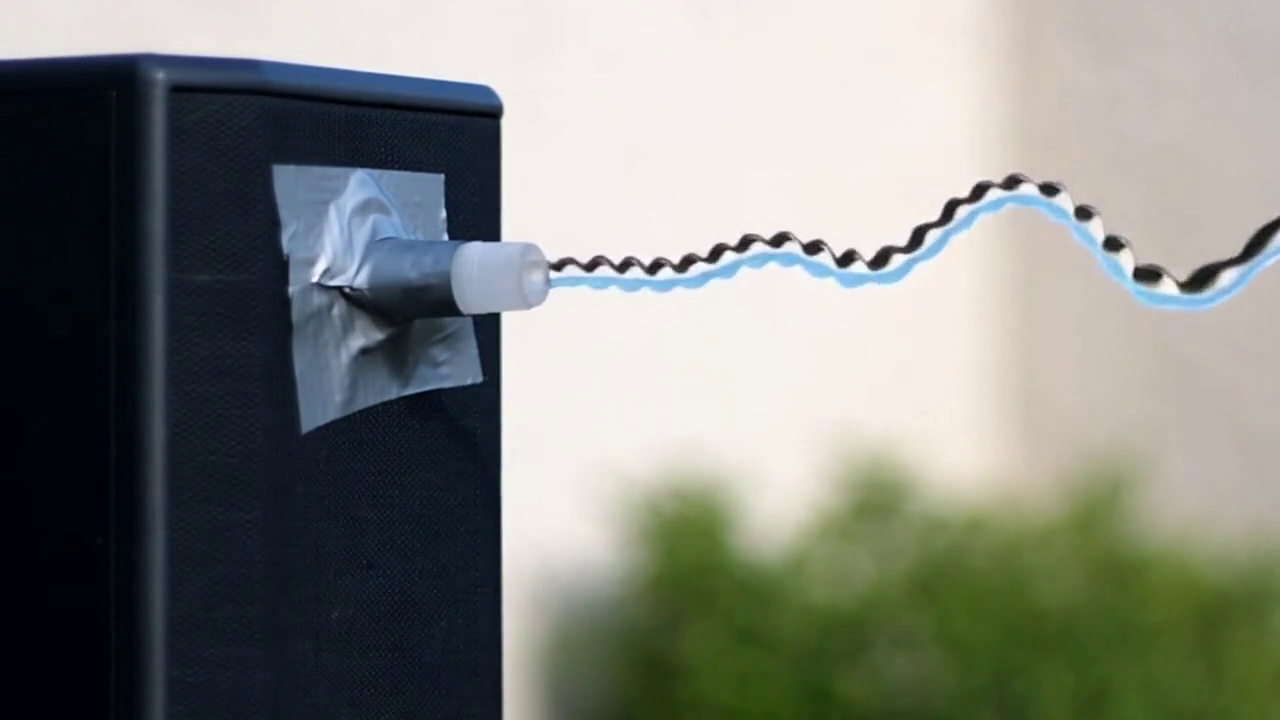}%
            \end{minipage}

        \end{minipage}
    
        %\vspace{1.5em}

        \begin{minipage}{0.99\textwidth}
        % ================= Row 2: Figure 3 (left) & Figure 6 (right) =================
        \begin{minipage}{0.51\textwidth}
            \centering
            \includegraphics[width=\linewidth]{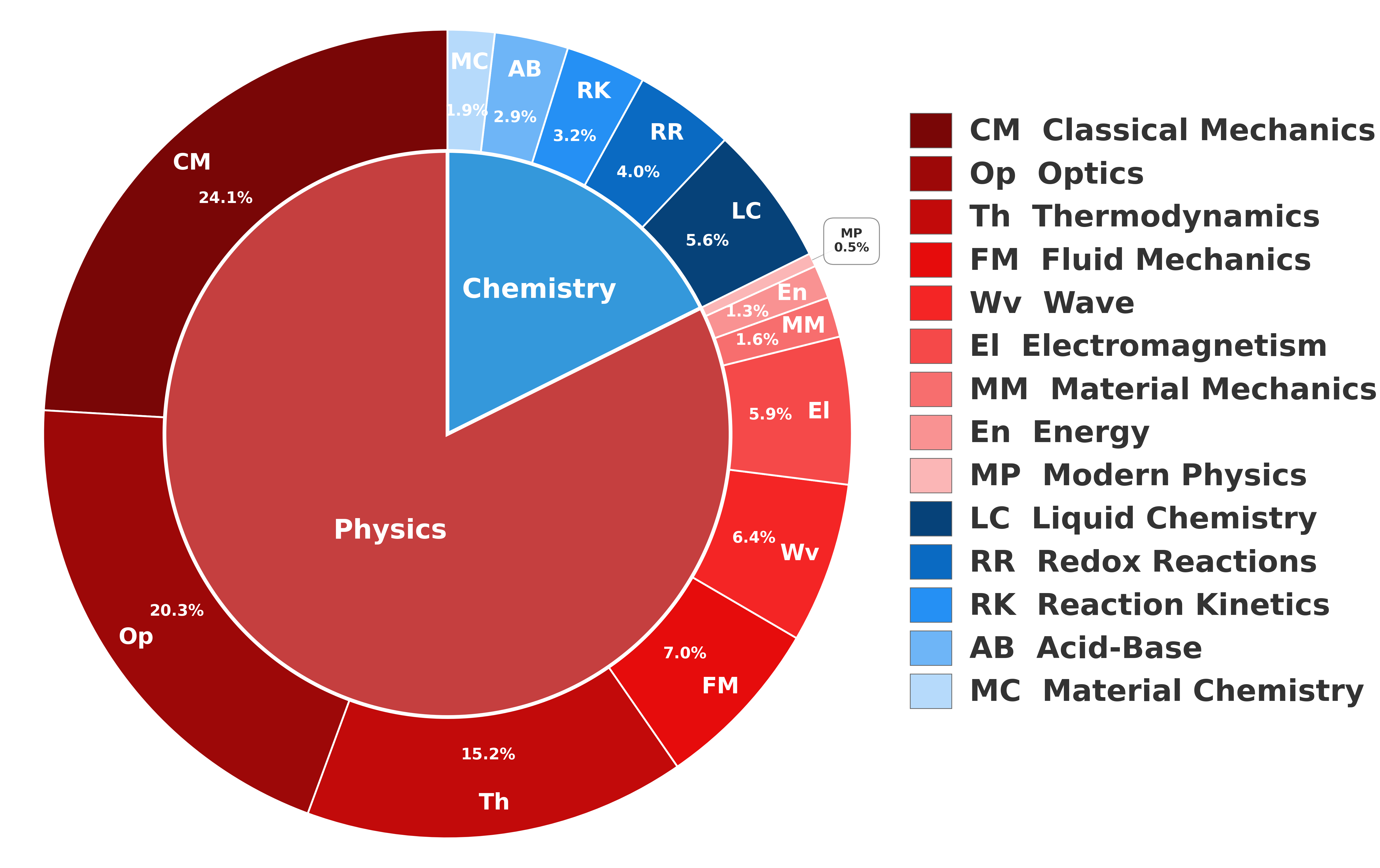}
            %\\[-0.2em]
            % {\small\textbf{(a) Subcategory frequency of questions}}
        \end{minipage}
        \hspace{-1.4em}
        \begin{minipage}{0.56\textwidth}
            \centering
            \includegraphics[width=\linewidth]{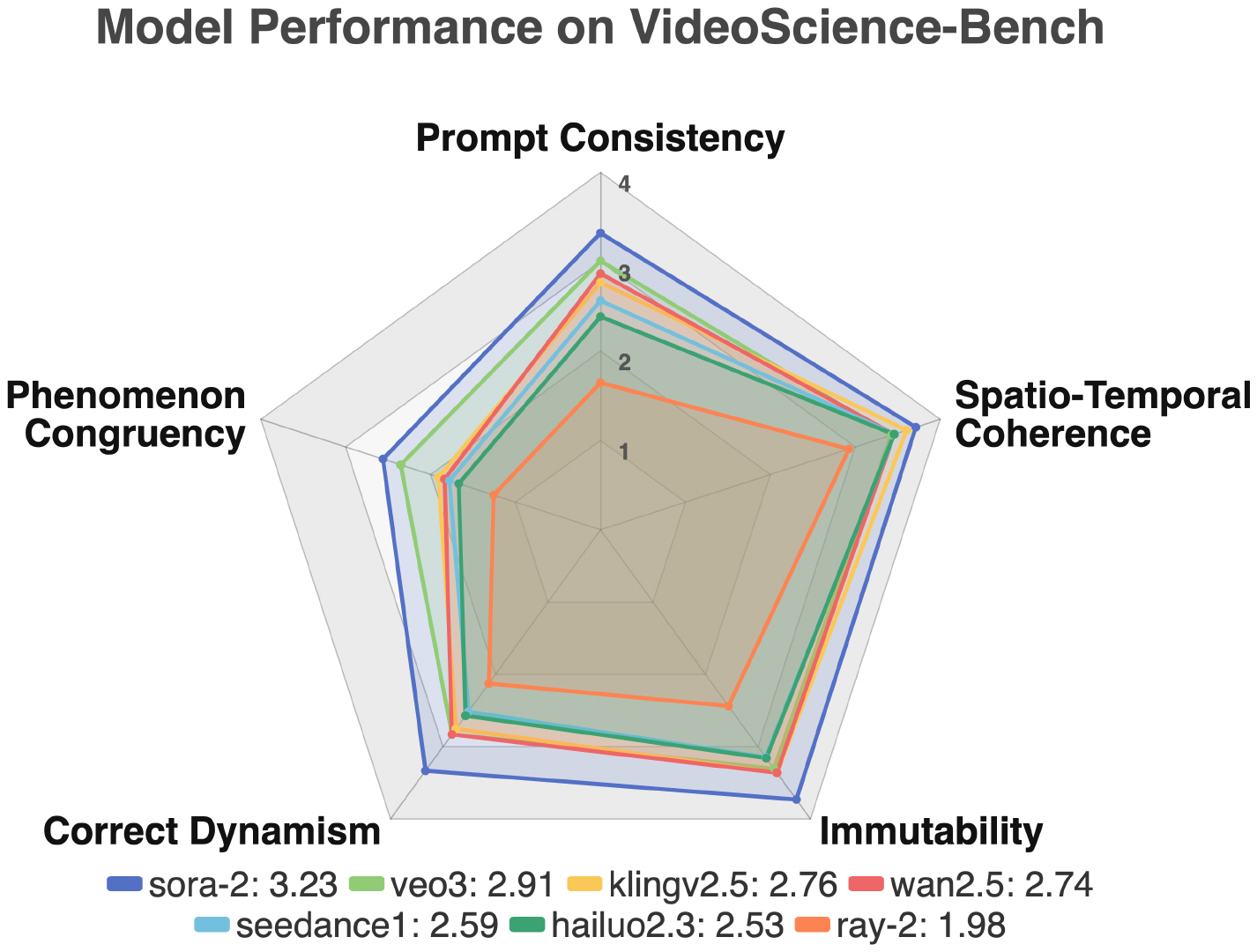}
            %\\[-0.2em]
            % {\small\textbf{(b) Expert model performance on \sysnamenospace-Bench}}
        \end{minipage}
        \end{minipage}
        % \caption{\textbf{Overview of \sysnamenospace-Bench.}
        % Top: Comparison of physical-commonsense versus scientific-reasoning generations using Sora-2. The scientific reasoning scenario hinges on recognizing that water’s high specific heat capacity lets it act as a heat sink, thereby preventing immediate balloon rupture.
        % Bottom-left: Subcategory frequency of questions From \sysnamenospace-Bench. Note that we are using two-letter code in the pie chart that maps the subcategory provided in the legend.
        % Bottom-right: Expert annotated model performance of seven video models on \sysnamenospace-Bench. The scores serve as the ground truth for our quantitative correlational analysis.}
        % \label{fig:first_page_overview}
    \end{figure}

\vspace{-2.5em}
\captionof{figure}{\textbf{\sysnamenospace-Bench Overview.}
        \textbf{Top}: Comparison of physical-commonsense versus scientific-reasoning generations using Sora-2. The first scientific reasoning scenario hinges on recognizing that water’s high specific heat capacity lets it act as a heat sink, thereby preventing immediate balloon rupture. The second example requires reasoning over physical vibration, wave and material properties.
        \textbf{Bottom-left}: Subcategory frequency of questions From \sysnamenospace-Bench. Note that we are using two-letter code in the pie chart that maps the subcategory provided in the legend.
        \textbf{Bottom-right}: Expert annotated model performance of seven video models on \sysnamenospace-Bench. The scores serve as the ground truth for our quantitative correlational analysis.}
        \label{fig:first_page_overview}
\vspace{0.5em}
}]

\begin{abstract}
% v0
%The next frontier for video generation lies in developing models capable of zero-shot reasoning, where understanding real-world scientific laws is essential to correct physical outcome modeling under diverse conditions. However, existing video benchmarks are physical commonsense-based, offering limited insight into video models' scientific reasoning capability. We introduce \sysnamenospace, a benchmark designed to evaluate undergraduate-level scientific understanding in video models. Each prompt encodes a composite scientific scenario that requires understanding and reasoning across multiple scientific concepts to generate the correct phenomenon. The benchmark comprises 200 carefully curated prompts spanning 14 topics and 103 concepts in physics and chemistry. We conduct expert-annotated evaluations across seven state-of-the-art video models in T2V and I2V settings along four dimensions: Immutability, Correct Dynamism, Spatio-Temporal Continuity, and Phenomenon Congruency. Using a VLM-as-a-Judge 
%with CV-augmented reasoning 
%to assess video generations, we observe strong correlation with human assessments. To the best of our knowledge, \sysnamenospace\ is the first benchmark to evaluate video models not only as generators but also as reasoners, requiring their generations to demonstrate scientific understanding consistent with expected physical and chemical phenomena.

%v1

The next frontier for video generation lies in developing models capable of zero-shot reasoning, where understanding real-world scientific laws is crucial for accurate physical outcome modeling under diverse conditions. However, existing video benchmarks are physical commonsense-based, offering limited insight into video models' scientific reasoning capability. We introduce \sysnamenospace-Bench, a benchmark designed to evaluate undergraduate-level scientific understanding in video models. Each prompt encodes a composite scientific scenario that requires understanding and reasoning across multiple scientific concepts to generate the correct phenomenon. The benchmark comprises 200 carefully curated prompts spanning 14 topics and 103 concepts in physics and chemistry. We conduct expert-annotated evaluations across seven state-of-the-art video models in T2V and I2V settings along four dimensions: Immutability, Correct Dynamism, Spatio-Temporal Continuity, and Phenomenon Congruency. Using a VLM-as-a-Judge to assess video generations, we observe strong correlation with human assessments. To the best of our knowledge, \sysnamenospace-Bench is the first benchmark to evaluate video models not only as generators but also as reasoners, requiring their generations to demonstrate scientific understanding consistent with expected physical and chemical phenomena. Our data and evaluation code are available at: \href{https://github.com/hao-ai-lab/VideoScience}{github.com/hao-ai-lab/VideoScience}.

\end{abstract}    
\section{Introduction}
\label{sec:intro}
% \input{figure_text/first_page_figures}

% video models are getting prevalent

% video model benchmarks are lacking, especially for zero-shot reasoning --> be a strong world model requires real-world scientific understanding and reasoning capability. current video model benchmarks are limited to physical common sense

% introduce our benchmark

% introduce our evaluation method

% summary

%\paragraph{Context.}
%Video generation models are rapidly maturing, with recent systems demonstrating improved visual fidelity, longer-range temporal coherence, and stronger prompt adherence. \todo{add sota model citations}

%\lx{I think we should add a teaser figure here. Could be a fun qualitative example.}

The launches of Sora-2~\citep{openai2025sora2} and Veo-3.1~\citep{deepmind2025veo3} mark a transformation in the video-generation landscape: video generation models~\citep{blattmann2023stable_video_diffusion, lu2023vdtgeneralpurposevideodiffusion} now deliver cinematic-level visual fidelity and exhibit unprecedented world-modeling capability. As video models evolve, the community’s notion of what constitutes “good” video generation is shifting beyond distributional quality metrics such as Fréchet Video Distance (FVD)~\citep{unterthiner2018fvd} and Fréchet Video Motion Distance (FVMD)~\citep{liu2024fvmd} toward evaluations that probe a model's world-modeling capability~\citep{huang2024vbench, he2024videoscore, he2025videoscore2, deng2025scivideobench} and assess whether the generated videos follow commonsense physical regularities~\citep{li2025worldmodelbench, lee2024vhelm, lin2024genaibench, bansal2024videophy, bansal2025videophy2, meng2025phygenbench}. Recent models have even begun to exhibit zero-shot reasoning abilities, tackling tasks involving scientific, mathematical, and spatial reasoning~\citep{wiedemer2025videomodelszeroshotlearners, tong2025thinkingvideovideogeneration}.

%Scientifically trained reasoners are able to synthesize multiple principles, mentally simulate experiments, and anticipate the resulting phenomena. 
Human reasoners trained in science model the world using a rich set of scientific principles and anticipate the resulting phenomena, like whether citric acid removes iron stains and what will happen if an LED is powered with or without a resistor.
%These developments raise a fundamental question: \textit{beyond visual fidelity and physical common sense, can video models demonstrate scientific reasoning ability when modeling the real world?} 
This motivates our central question: \textit{beyond visual fidelity and physical commonsense, can video models correctly portray scientific phenomena and demonstrate scientific reasoning about the real world?}
World modeling grounded on scientific principles plays a key role in advancing discovery by allowing models to predict outcomes of real-world processes.
%based on learned causal and scientific relationships. 
Unlike commonsense-based world modeling, true scientific world modeling requires reasoning across \textit{multiple interacting concepts}, for example, Figure~\ref{fig:first_page_overview} shows that integrating specific‐heat and heat‐transfer principles predicts water will absorb the heat and prevent immediate balloon rupture.
%\mk{[Teaser Fig], xxx}
%\yujie{Add an interest example to be more comprehensive.}

% \input{figure_text/teaser_examples}

However, existing video model benchmarks mostly focus on physical commonsense, restricting evaluations to everyday scenarios that only require basic understanding of physical laws~\citep{li2025worldmodelbench, bansal2025videophy2, meng2025phygenbench, bansal2024videophy, sun2024t2vcompbench}. We introduce \textit{\sysnamenospace-Bench}, a benchmark designed to assess the scientific understanding and reasoning of video models. Each instance involves multiple scientific concepts and requires undergraduate-level knowledge, spanning 103 concepts across physics and chemistry. We evaluate whether video models can serve as zero-shot scientific reasoners, capable of understanding complex experimental setups and generating expected phenomena. The benchmark measures performance along five key dimensions: \emph{Prompt Consistency}, \emph{Expected Phenomenon}, \emph{Dynamism}, \emph{Immutability}, and \emph{Coherence}. 

We also introduce \textit{\sysnamenospace-Judge}, a checklist-based, 
%and key-frame-based 
automated VLM-as-a-Judge evaluation framework ~\citep{liu_2024_evalcrafter_cvpr, han_2025_videobench_cvpr}. \sysnamenospace-Judge first generates a prompt-specific checklist to guide the evaluation, then selects key frames to identify causally salient moments in each video. These components jointly support the VLM-based scoring process, and the resulting scores are aggregated into a single weighted metric reflecting the model’s overall reasoning quality. For validation, we validate the automated VLM judgments significantly better align with domain-expert annotations in comparison with other physical-commonsense-based and visual fidelity benchmarks.

In summary, this paper investigates whether video generation models can reason over multiple interacting scientific concepts. To this end, we make three key contributions:
\begin{enumerate}

\item We introduce \sysnamenospace-Bench, a high-quality benchmark targeting zero-shot scientific reasoning in video generation models, encompassing real-world phenomena across 103 physics and chemistry concepts.

\item We develop \sysnamenospace-Judge, a checklist-based 
%and key-frame-based 
evaluation framework that leverages VLM-as-a-Judge, integrates key-frame selection and computer-vision tools for detecting physical cues (e.g., motion, object interactions), and outputs quantitative scores of scientific validity alongside perceptual quality.

\item We conduct a comprehensive empirical study on \sysnamenospace-Bench. While current video models produce high-quality, photorealistic and temporally coherent videos, \sysnamenospace-Bench remains challenging: many systems still lack the capacity to model complex scientific phenomena and often violate basic physical laws. Among the evaluated models, Sora-2 and Veo-3 perform best and show early signs of scientific reasoning.
\end{enumerate}

% In summary, this paper evaluates whether \mk{SOTA} video models can perform physical outcome modeling~\mk{why physical outcome? how about chemistry?}at an undergraduate-level of scientific sophistication. To this end, we make three key contributions:
% \begin{enumerate}
% \item We introduce \sysnamenospace-Bench, a high-quality benchmark targeting zero-shot, advanced scientific reasoning in video generation models, covering real-world phenomena in physics and chemistry.
% \item We develop \sysnamenospace-Judge, a checklist- and key-frame-based evaluation framework that uses VLM-as-a-judge, incorporates key-frame selection, and integrates computer-vision tools to assess scientific validity alongside perceptual quality. \lx{elaborate a bit quantitatively}
% \item We present a comprehensive empirical study across both text-to-video and image-to-video (I2V) settings, revealing significant gaps between perceptual fidelity and genuine scientific understanding and reasoning for video generation for current SOTA video models.
% \end{enumerate}

\section{\sysnamenospace-Bench}
\label{sec:bench}

\begin{figure*}[t!]
    \centering
    \includegraphics[width=\linewidth]{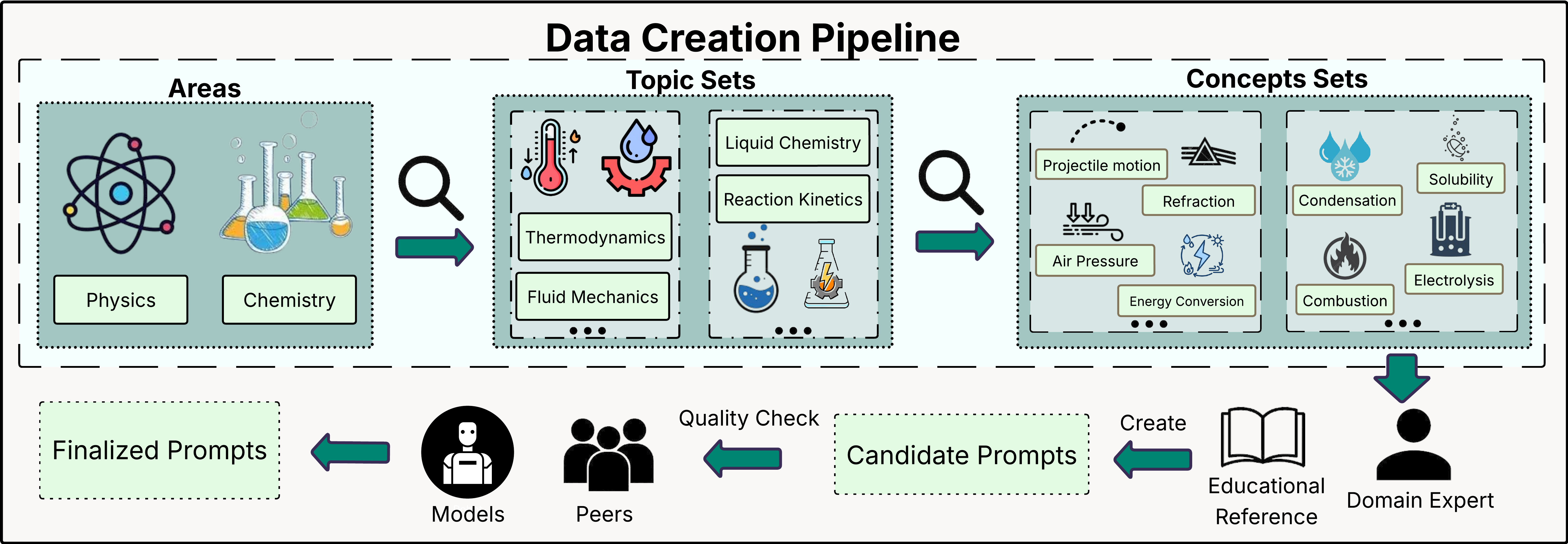}
    \caption{%
        Overview of the data creation pipeline. 
        Each researcher selects two or more scientific concepts and references relevant educational materials or videos to design a prompt. 
        Prompts undergo peer and model review, followed by model-based quality checking, before being finalized for dataset inclusion.
    }
    \vspace{-0.5em}
    \label{fig:eval_pipeline}
\end{figure*}

\sysnamenospace-Bench is designed to evaluate video models' capability of generating correct physical and chemical outcomes for complex scientific events that require an understanding of multiple concepts. 
% In this section, we elaborate on the benchmark design in Section~\ref{sec:bench-overview} and benchmark making process in Section~\ref{sec:bench-building}.

\subsection{Benchmark Overview}
\label{sec:bench-overview}

\paragraph{Categorization.}
%\todo{categorization: term $\to$ subcategory $\to$ supergroup}
% Leveraging well-established topic categorizations from undergraduate physics and chemistry education, \sysnamenospace-Bench comprises of 160 questions spanning 14 topics (9 of them related to physics and 5 related to chemistry) and 103 total concepts (79 of them related to physics and 24 related to chemistry). Example topics include physics related ones like Classical Mechanics and Optics, to chemistry related ones like Liquid Chemistry and Reaction Kinetics. The topics are carefully chosen to reflect commonly encountered course titles in undergraduate science education, while the concepts represent more fine-grained categorization corresponding to individual chapters or sections within these courses. The category-frequency distribution for all prompts is provided in Figure~\ref{fig:categorical_breakdown}.

Leveraging established topic categorizations from undergraduate physics and chemistry, \sysnamenospace-Bench comprises 160 questions covering 14 topics (9 related to physics and 5 related to chemistry) and 103 concepts (with 79 associated with physics and 24 with chemistry). Example topics include physics-related areas like Classical Mechanics and Optics, as well as chemistry-related topics like Liquid Chemistry and Reaction Kinetics. The selected topics aim to reflect commonly encountered course titles in undergraduate science education, while the concepts provide a more detailed categorization corresponding to specific chapters or sections within these courses. The category-frequency distribution for all prompts is illustrated in Figure~\ref{fig:first_page_overview}.

%\yx{double check numbers and update} 

%\paragraph{ \yx{provide justification and examples on how topics and concepts are chosen. e.g. } }

%Example categories are solubility, sound, and energy conversion. 

%\lx{json dict use for temporary reference, move to appendix later.}

% \input{figure_text/categorical_breakdown}

\paragraph{Distinction with Other Scientific Benchmarks.} 
% Existing scientific benchmarks for video models primarily test physical commonsense. For example, they include mirror reflections, bird flight, falling water, or bouncing basketballs, phenomena that can be understood with high-school level or more basic scientific knowledge~\citep{li2025worldmodelbench, bansal2025videophy2, meng2025phygenbench, bansal2024videophy}. In contrast, each problem in \sysnamenospace-Bench involves at least two interacting undergraduate-level scientific principles. For example, analyzing the bending of a laser beam traversing a sugar solution with a concentration gradient, requires integrating the Snell's law in optics with diffusion physics (Figure~\ref{fig:example175}).

Current scientific benchmarks for video models primarily assess physical common sense. These benchmarks typically include daily scenes such as mirror reflections, bird flight, falling water, and bouncing balls, which can be understood with a basic or high-school level of scientific knowledge~\citep{li2025worldmodelbench, bansal2025videophy2, meng2025phygenbench, bansal2024videophy}. In contrast, each challenge in \sysnamenospace-Bench requires the understanding of at least two undergraduate-level scientific principles. For instance, analyzing the bending of a laser beam as it passes through a sugar solution with a concentration gradient requires the understanding and application of Snell's Law in optics alongside principles from diffusion physics (see Figure~\ref{fig:example175_t2v}).

\paragraph{Data Splits.}

% \sysnamenospace-Bench consists of two parts: The T2V split, which includes 160 question prompts; and the I2V split, which includes a set of 40 question prompts paraphrased from the T2V split along with first-frame images for the initial experimental setups. A detailed account of the curation process appears in Section ~\ref{sec:bench-building}.  We provide a few examples from each split in Appendix~\ref{appendix:test_suite_examples}. 

The \sysnamenospace-Bench consists of two parts: the T2V split, which includes 160 question prompts, and the I2V split, which features a set of 40 question prompts that have been paraphrased from the T2V split. The I2V split also includes first-frame images for the initial experimental setups. A detailed description of the curation process can be found in Section ~\ref{sec:bench-building}. We provide several examples of each split in Appendix~\ref{appendix:test_suite_examples}.

%\todo{add qualitative examples}

\paragraph{Metrics.}
\label{para:rubrics}

% \sysnamenospace-Bench evaluates generated videos along five dimensions that capture both perceptual quality and physical fidelity. The perceptual aspects are \emph{Immutability} and \emph{Spatio-temporal Coherence}; the physical aspects are assessed at two levels, \emph{Phenomenon Congruency} and \emph{Correct Dynamism}. A prerequisite for meaningful assessment is \emph{Prompt Consistency}, i.e. the experimental setup and procedure must match the prompt. These criteria align with recent practice in video-generation benchmarking, and a correlational analysis is provided in Section~\ref{sec:exp-correlational-analysis}. The rating criteria for each of the five dimensions are elaborated below:
% \begin{enumerate}
%   \item \textbf{Prompt Consistency:} the experimental setup and procedure match with description in the prompt.
%   \item \textbf{Phenomenon Congruency:} the observed outcomes align with expected scientific principles.
%   \item \textbf{Correct Dynamism:} other fundamental physical laws are complied with, including motion and interactions among objects that appear in the generation.
%   \item \textbf{Immutability:} objects remain unchanged when no transformation is expected.
%   \item \textbf{Spatio-Temporal Coherence:} transitions between frames are natural and temporally consistent.
% \end{enumerate}
% Each dimension was scored on a four-point ordinal scale: 
% 1 = absent or contradictory, 2 = weak or partially incorrect, 3 = mostly correct, and 4 = clearly correct. 

\sysnamenospace-Bench evaluates generated videos across five dimensions that reflect both perceptual quality and physical fidelity. The perceptual aspects include \emph{Immutability} and \emph{Spatio-temporal Coherence}, while the physical aspects are assessed through two levels: \emph{Phenomenon Congruency} and \emph{Correct Dynamism}. For the assessment to be meaningful, \emph{Prompt Consistency} is essential, which means that the experimental setup and procedure must align with the prompt. These evaluation criteria are consistent with current practices in video-generation benchmarking, and a correlational analysis is presented in Section~\ref{sec:exp-correlational-analysis}. Each dimension is scored on a four-point ordinal scale: 1 (absent or contradictory), 2 (weak or partially incorrect), 3 (mostly correct), and 4 (clearly correct). Below, we elaborate on the rating criteria for each of the five dimensions:
\begin{itemize}
  \item \textbf{Prompt Consistency:} The experimental setup and procedures align with the descriptions provided in the prompt. 
  \item \textbf{Phenomenon Congruency:} The observed outcomes are consistent with expected scientific principles.
  \item \textbf{Correct Dynamism:} All other fundamental physical laws are followed, including those governing motion and interactions among the objects involved.  
  \item \textbf{Immutability:} Objects remain unchanged when no transformation is anticipated.
  \item \textbf{Spatio-Temporal Coherence:} Transitions between frames are smooth, natural, and temporally consistent.  
\end{itemize}

\subsection{Benchmark Building}
\label{sec:bench-building}

%\paragraph{\todo{we curate a list of concepts, ask expert to craft, and a panel to review (1. human reader check; 2. video model quality check)} }

In \sysnamenospace-Bench, we adopt a three-stage data construction pipeline as shown in Figure~\ref{fig:eval_pipeline}.

\paragraph{Question Making.} 

Following the categorization outlined in Section~\ref{sec:bench-overview}, eight domain-specific graduate student experts selected subsets of categories and authored 20 prompts each. Each prompt was designed to require reasoning involving at least two scientific concepts, ensuring a sufficient level of difficulty. The prompts were based on public educational experiment examples~\citep{wonderofscience2025phenomena, experimentarchive2025, rsc2025practical_videos} to guarantee scientific validity and to ensure that each target phenomenon could be feasibly represented in a video of no more than 10 seconds. Among these contributors, two graduate students also curated 40 I2V questions, each paired with a reference video that depicted the target experiment as the ground truth. The reference videos were sourced from public datasets and online resources~\citep{xu2025expvid, zou2025physlab}, with a key frame extracted from the initial segment used as the conditional input image to illustrate the complete experimental setup.

\paragraph{Panel Review and Quality Check.}
After drafting the full test suite, all participating graduate students convened as a review panel. The panel cross-checked each expert’s set of 20 prompts to verify specificity, procedural accuracy, and clarity of the expected outcomes. As a final quality assurance step, we employed capable video generation models, Sora-2 and Wan-2.5-T2V-Preview~\citep{wan2025wan2.5preview}, to synthesize the opening seconds of each prompt, confirming that the key experimental components and setups can be captured.

\section{\sysnamenospace-Judge}

%\paragraph{\todo{simple VLM as a judge}}
%Concretely, our judge consumes key frames plus auxiliary CV signals and returns dimension-wise scores; we validate alignment on a human-annotated subset, following best practices in recent video benchmarks and LLM-as-a-judge studies.

%\paragraph{\todo{augmentation: +CV as tools, +rubrics-based, +both}}

\paragraph{VLM-as-a-Judge.}

\begin{figure}[htbp]
    \centering
    \includegraphics[width=0.9\linewidth]{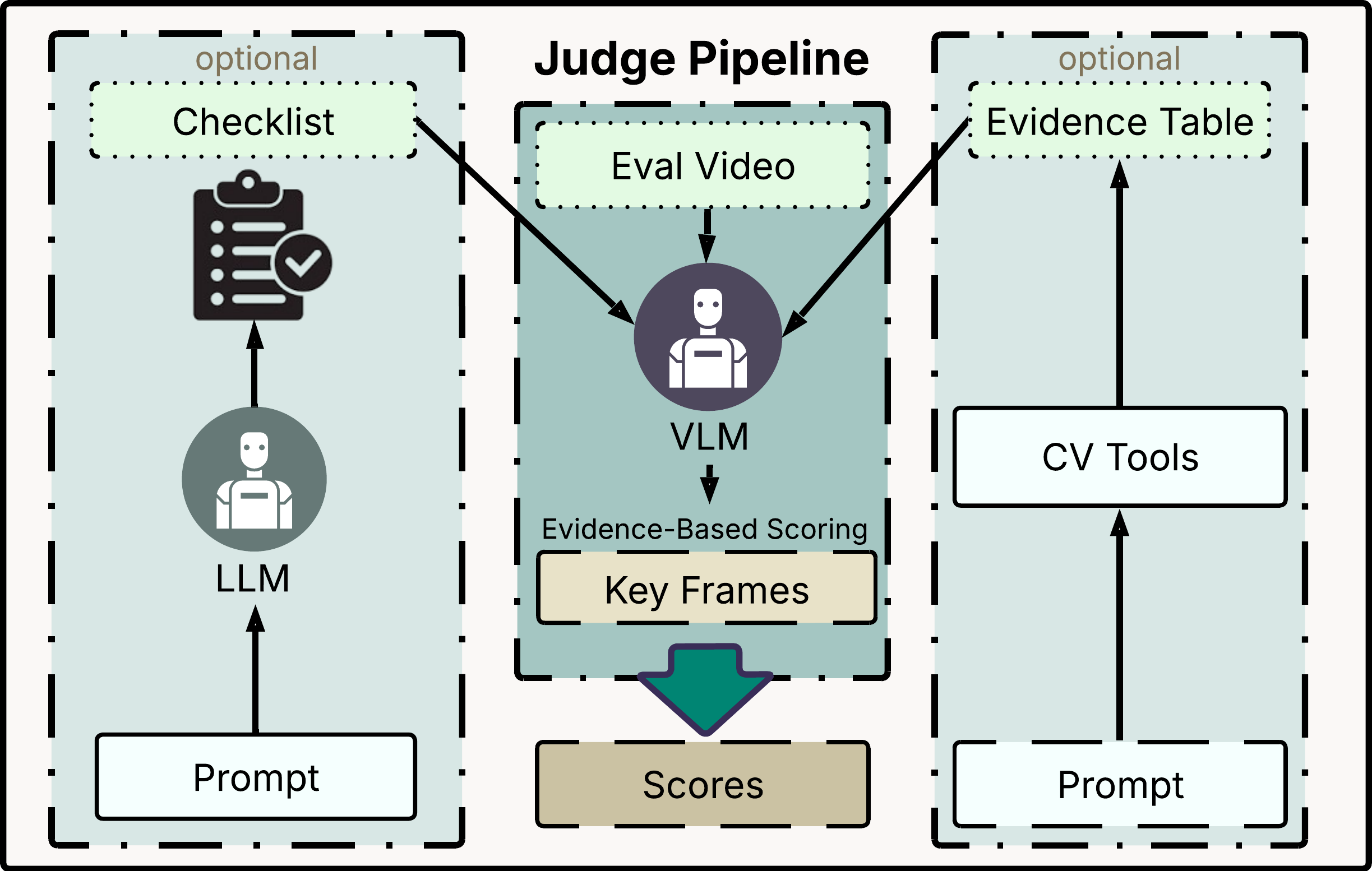}
    \caption{
        Overview of our VLM-as-a-Judge pipeline, which combines checklist-based scoring, key-frame extraction, and evidence aggregation to ensure grounded and traceable video evaluations.
    }
    \vspace{-2em}
    \label{fig:judge_pipeline}
\end{figure}

Recent studies have shown that powerful VLMs can produce human-aligned judgments when carefully prompted and calibrated~\cite{liu_2024_evalcrafter_cvpr,han_2025_videobench_cvpr}.
In \sysnamenospace-Bench, we adopt a VLM-as-a-Judge framework and examine the performance gap between reasoning-oriented VLMs and non-reasoning models such as GPT-4o. We find that directly prompting VLMs, particularly non-reasoning ones, to evaluate video generations often results in under-justified assessments and false-positive ratings (see Appendix~\ref{appendix:judge_model_choice}). This observation motivates the design of a quantitative auto-eval harness featuring a checklist-based grading scheme, where each rating must be supported by salient video frames and an evidence table aggregated from computer-vision analysis of the videos as shown in Figure~\ref{fig:judge_pipeline}.
%our VLM-as-a-Judge pipeline integrates rubric-guided scoring, key-frame extraction, and structured evidence collection to ensure grounded and traceable judgments.
The full prompt template used in our VLM-as-a-Judge framework is provided in Appendix~\ref{appendix:prompt_judge}.

\paragraph{Checklist Generation and Key Frame Extraction.} To improve interpretability and reduce false positives, 
%we develop a context engineering technique to enhance VLM-based grading. Specifically, 
we incorporate prompt-specific checklists generated by an LLM agent, and the fraction of checklist items satisfied within each evaluation category is used to derive an ordinal rating on a four-point scale. Each checklist is scored in a reductive, itemized manner, where deductions are only made when supported by concrete evidence. In \sysnamenospace-Judge, we leverage the ground-truth reference phenomenon and reference video (in the I2V split) to guide the VLM in identifying key frames where checklist violations occur, thereby providing evidence-grounded deductions. The prompt template used for checklist generation is provided in Appendix~\ref{appendix:prompt_checklist}.

\paragraph{CV-Augmented Reasoning.}
%We add deterministic CV modules that also produce quantitative evidence for the judge. In practice, lightweight scripts perform keyframe selection, detect objects and attributes, track entities, estimate motion with optical flow, and align text–video content. When a reference video is available, we also compute simple similarity scores (LPIPS~\citep{zhang2018perceptual} using VGG~\citep{simonyan2015very}). Each module emits a compact JSON record (\texttt{frame\_second}, \texttt{type}, \texttt{region}, \texttt{score}, \texttt{attributes}). An aggregator maps these records to checklist items, so GPT-5-Pro receives (i) the checklist and (ii) an evidence table. The model then converts the fraction of satisfied items per category into a four-point rating. 

%We add deterministic CV modules that produce quantitative evidence for the judge. Lightweight scripts perform keyframe selection, object detection, multi-object tracking, optical-flow motion analysis, and text–video alignment. When a reference video is available, we compute LPIPS similarity (VGG backbone)~\citep{zhang2018perceptual,simonyan2015very}. Each module outputs a compact JSON record (\texttt{frame\_second}, \texttt{type}, \texttt{region}, \texttt{score}, \texttt{attributes}). An aggregator links these records to checklist items, so GPT-5-Pro receives (i) the checklist and (ii) an evidence table. The judge then maps the fraction of satisfied items per category to a four-point rating.

\sysnamenospace-Judge further incorporates deterministic CV modules that produce quantitative evidence for the judge. The toolbox includes: Grounding DINO for open-vocabulary-conditioned detection to verify entity presence and attributes in experiment setups~\citep{liu2023groundingdino}, ByteTrack for identity association and to quantify spatio-temporal coherence~\citep{zhang2022bytetrack}, RAFT optical flow to capture and quantify motion direction and magnitude~\citep{teed2020raft}, as well as CLIP4Clip for text–video alignment~\citep{luo2022clip4clip}. When a reference video is available, we additionally report LPIPS similarity with a VGG backbone \citep{zhang2018perceptual,simonyan2015very}. Each module outputs compact JSON records (\texttt{frame\_second}, \texttt{type}, \texttt{region}, \texttt{score}, \texttt{attributes}). Depending on the evaluation setting, the prompt aggregator optionally attaches a checklist, an evidence table, both, or neither to the prompt, and instructs the model to convert the fraction of satisfied items per category into a four-point rating.

\begin{figure*}[t]
\centering
\setlength{\tabcolsep}{3pt}
\renewcommand{\arraystretch}{1.15}

\begin{tabular}{@{}p{0.15\textwidth}p{0.82\textwidth}@{}}
\toprule
\multicolumn{2}{c}{\textbf{Aluminum-Iodine Reaction}}\\
\midrule
\textbf{Prompt} & 
A few \textbf{\underline{\smash{drops of water}}} are added to a small pile of \textbf{\underline{\smash{powdered aluminum}}} and \textbf{\underline{\smash{iodine crystals}}} in a shallow dish.\\[0.2em]

\textbf{Expected} & 
No reaction occurs before water is added. After water is added, the mixture spontaneously \textbf{\underline{\smash{ignites}}} with a \textbf{\textcolor{violet}{\underline{violet--blue flash}}} and smoke.\\
\midrule

\multicolumn{2}{c}{\cellcolor{green!10}\textbf{Sora-2:} {\color{green!60!black} Successfully Generated Expected Phenomenon}}\\[0.1em]

\multicolumn{2}{c}{
  \includegraphics[height=1.2cm]{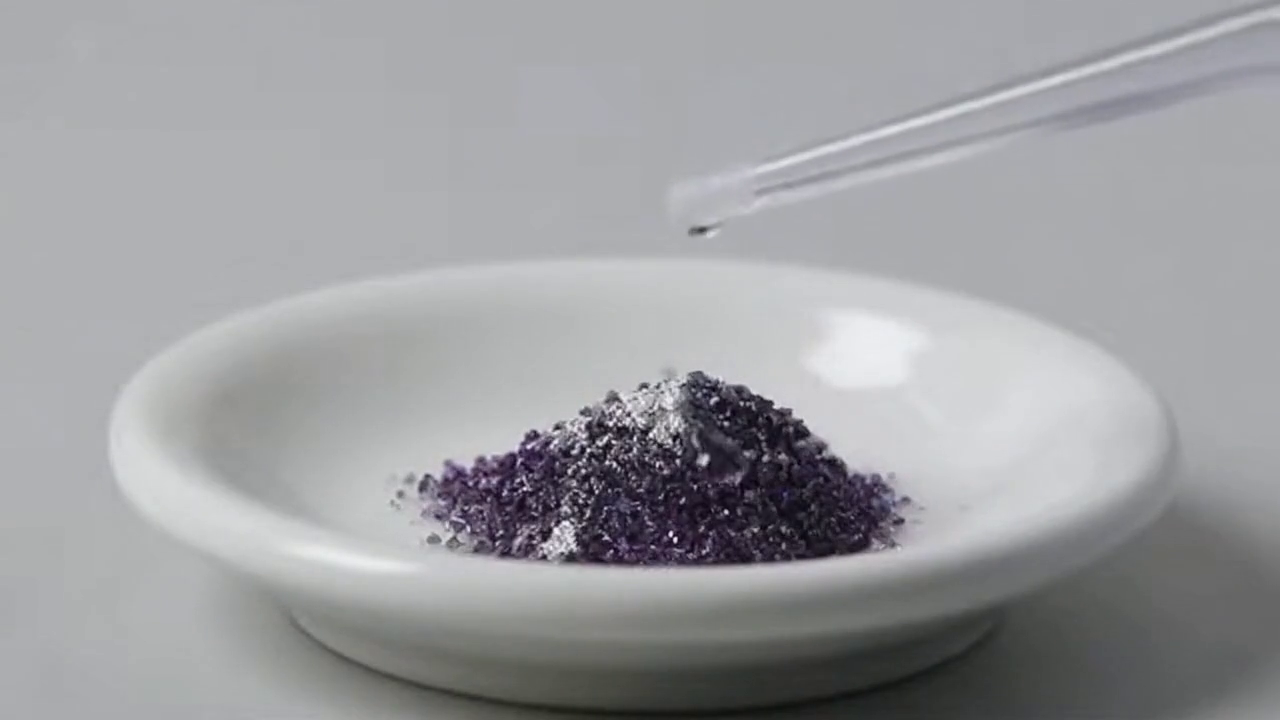}\hspace{1pt}%
  \includegraphics[height=1.2cm]{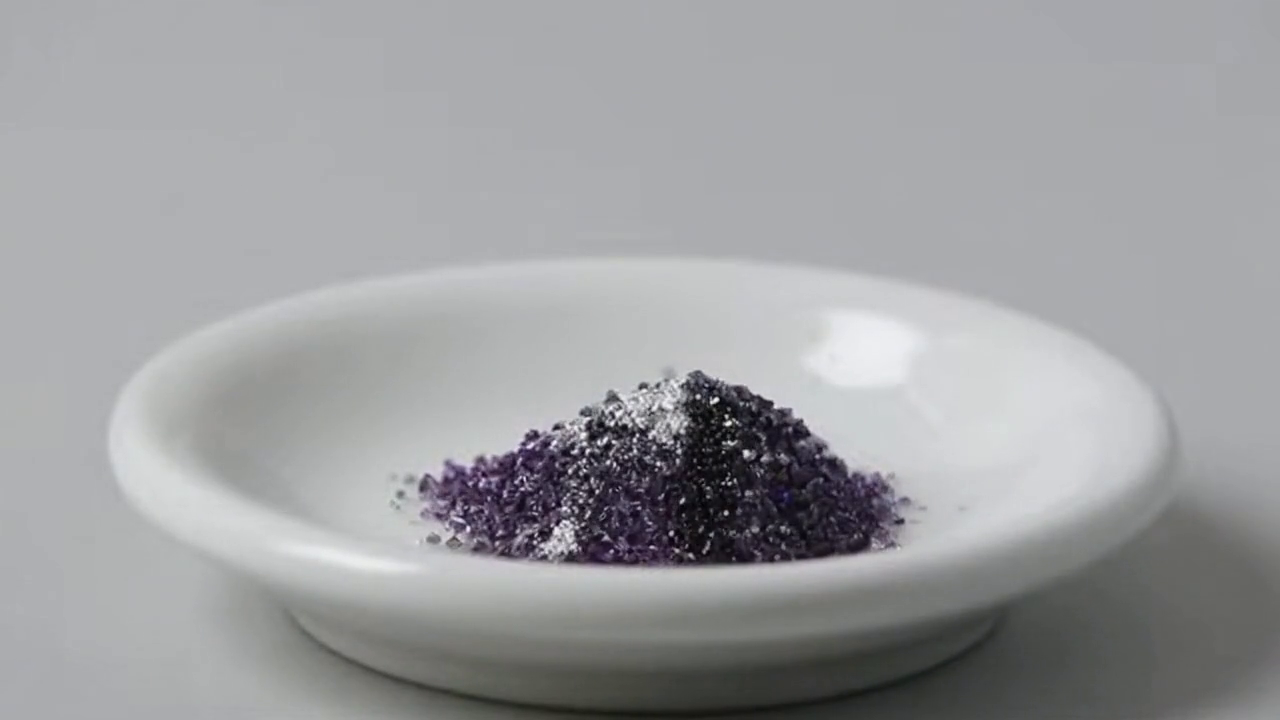}\hspace{1pt}%
  \includegraphics[height=1.2cm]{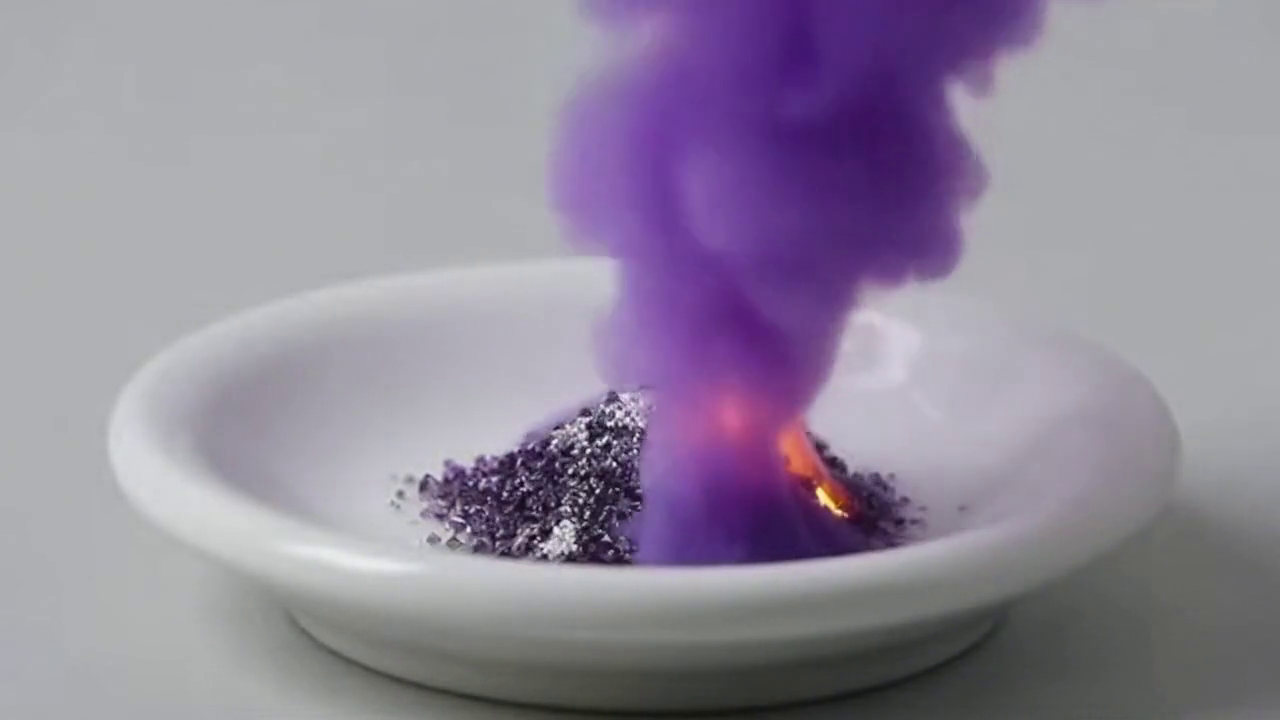}\hspace{1pt}%
  \includegraphics[height=1.2cm]{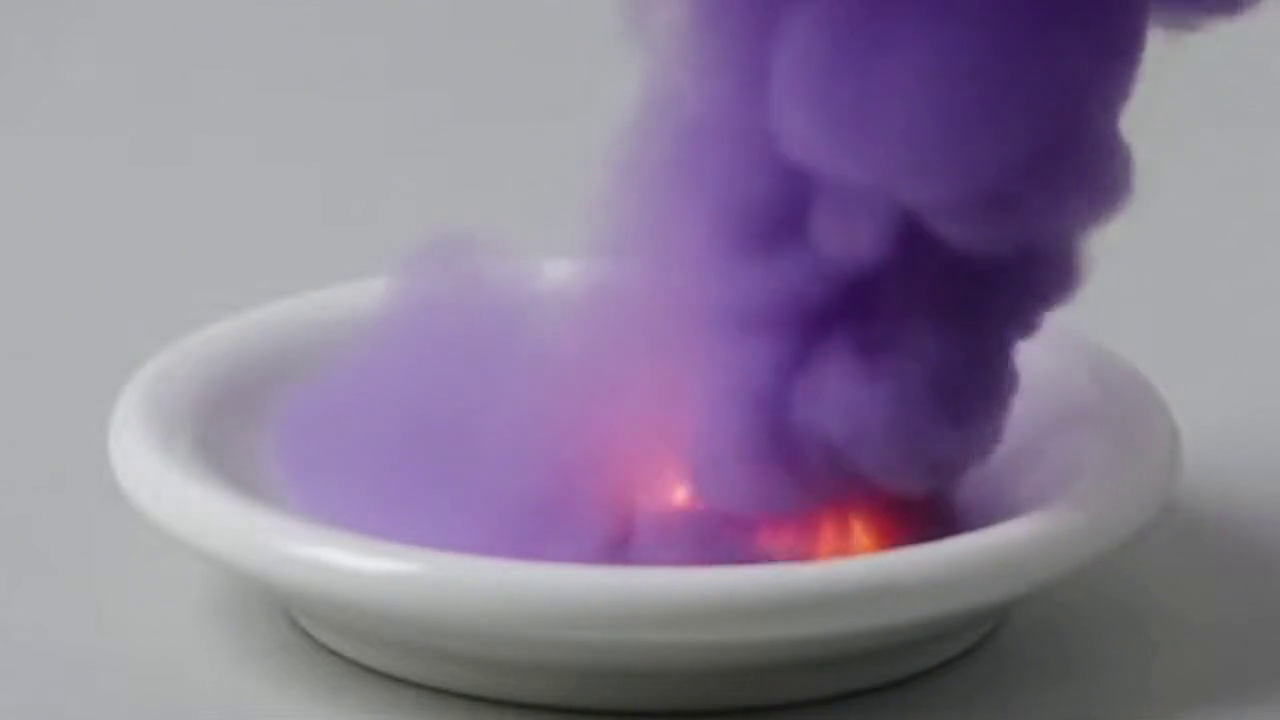}\hspace{1pt}%
  \includegraphics[height=1.2cm]{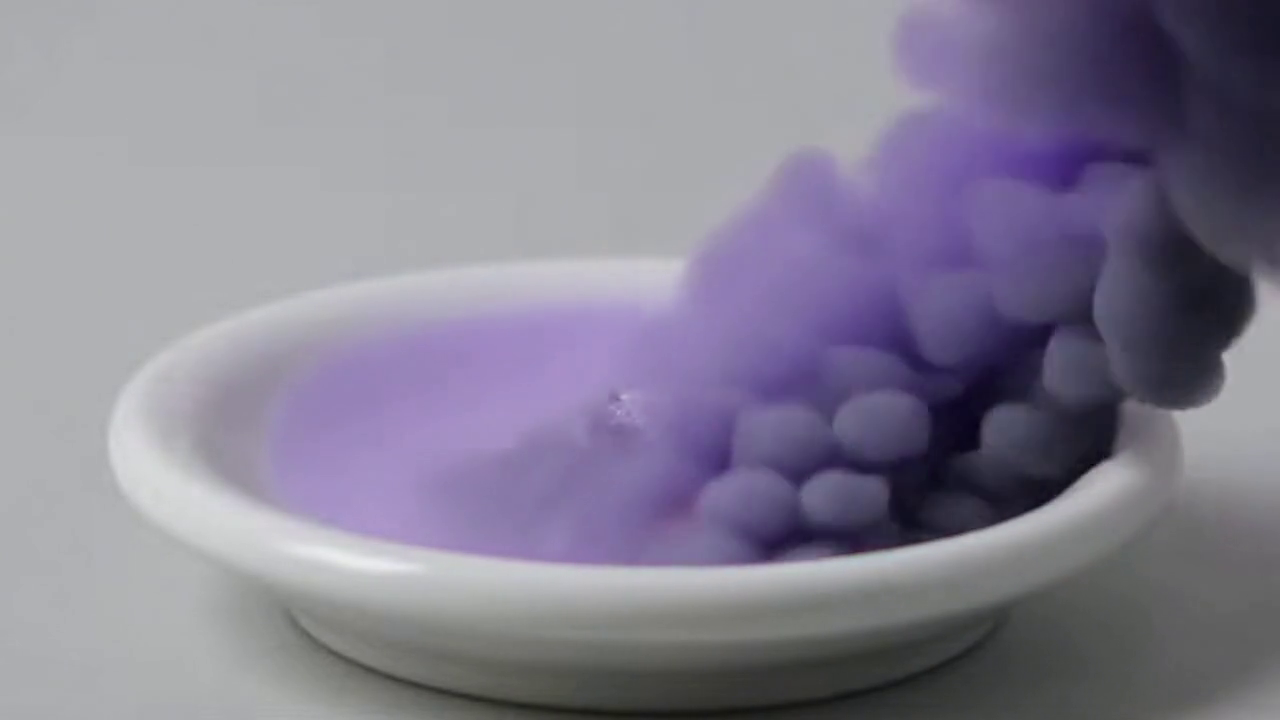}\hspace{1pt}%
  \includegraphics[height=1.2cm]{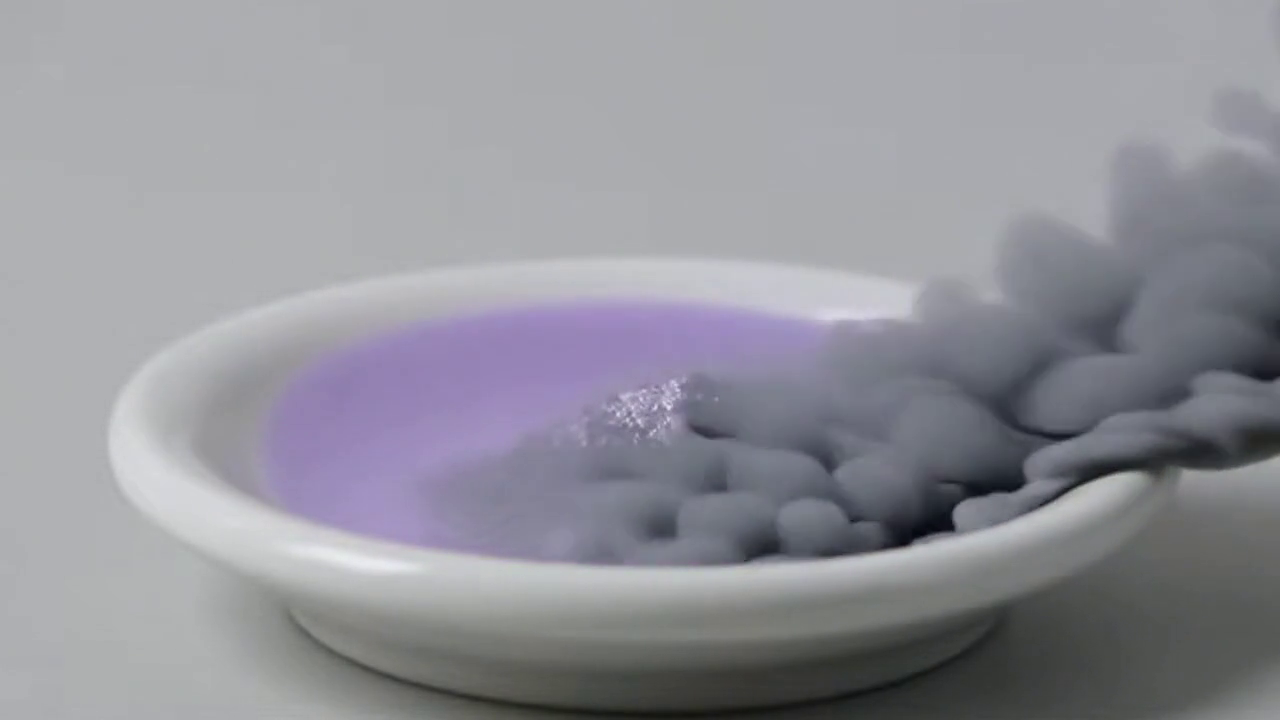}\hspace{1pt}%
  \includegraphics[height=1.2cm]{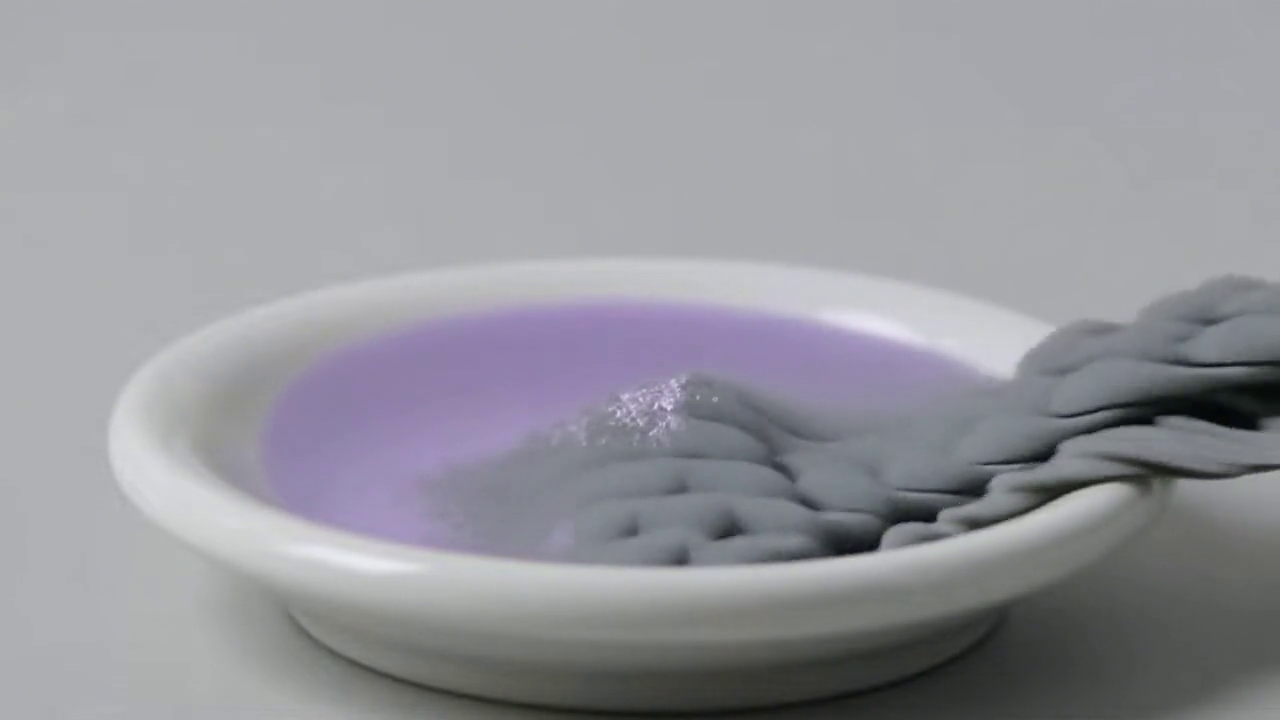}%
}\\[0.1em]

\multicolumn{2}{c}{
  \textbf{PCS:} 4/4 \quad 
  \textbf{PCG:} 4/4 \quad 
  \textbf{CDN:} 4/4 \quad 
  \textbf{IMB:} 4/4 \quad 
  \textbf{STC:} 4/4
}\\
\midrule

\multicolumn{2}{c}{\cellcolor{red!10}\textbf{Hailuo-2.3:} {\color{red!70!black}Failed to Generate Expected Phenomenon}}\\[0.1em]

\multicolumn{2}{c}{
  \includegraphics[height=1.2cm]{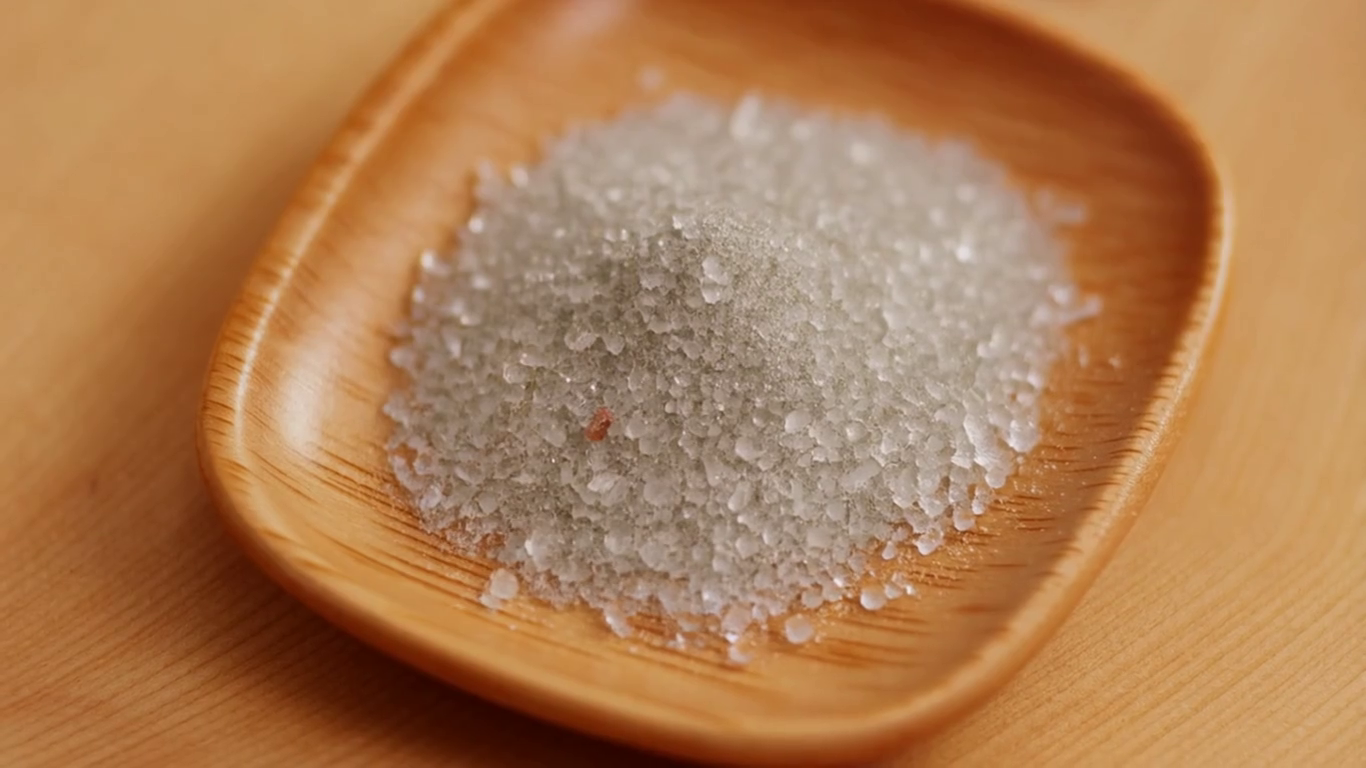}\hspace{1pt}%
  \includegraphics[height=1.2cm]{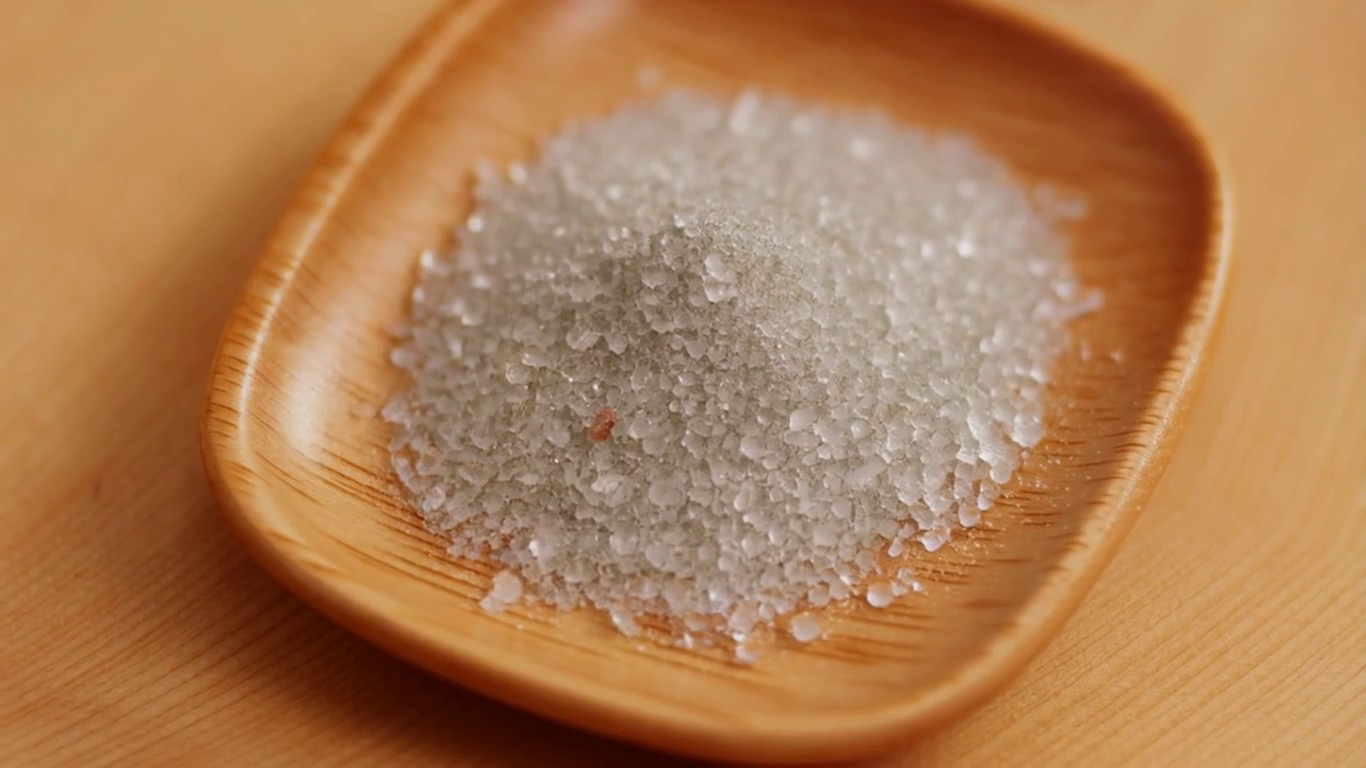}\hspace{1pt}%
  \includegraphics[height=1.2cm]{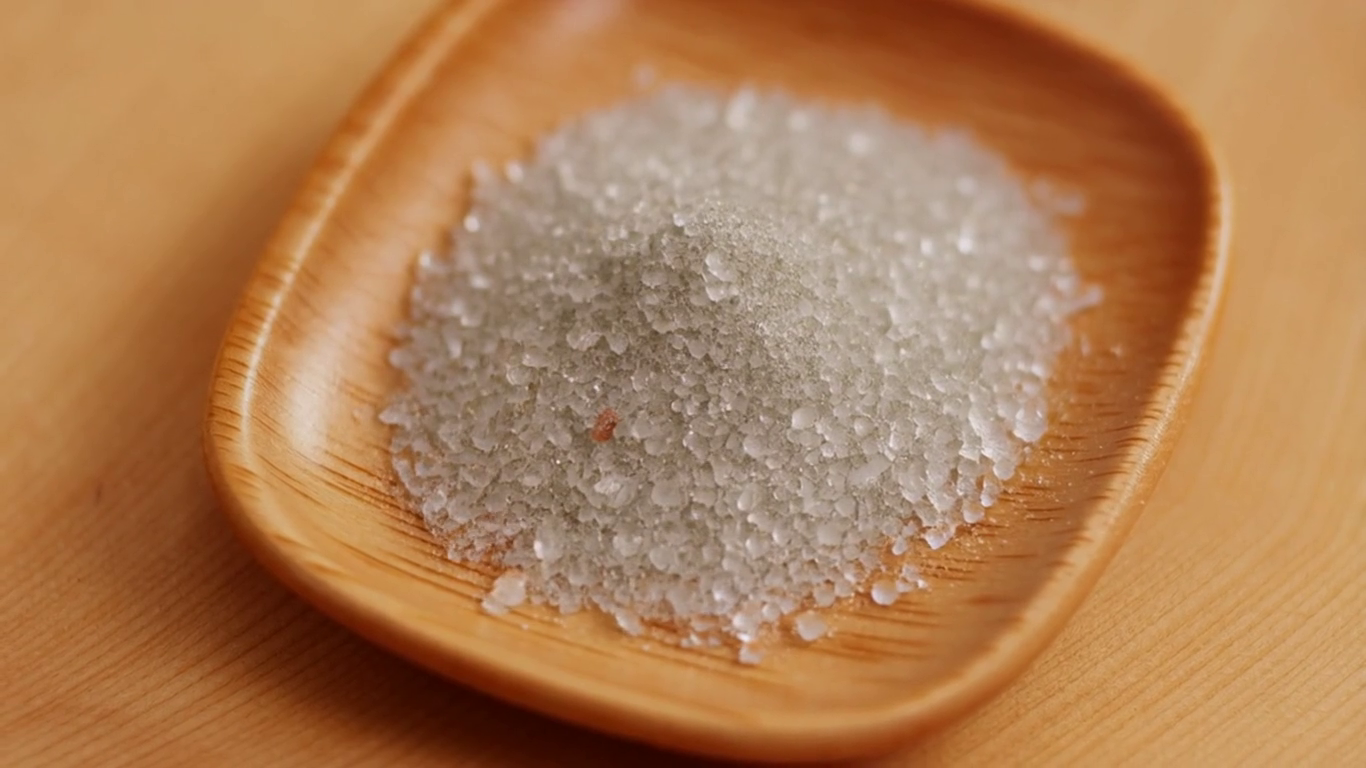}\hspace{1pt}%
  \includegraphics[height=1.2cm]{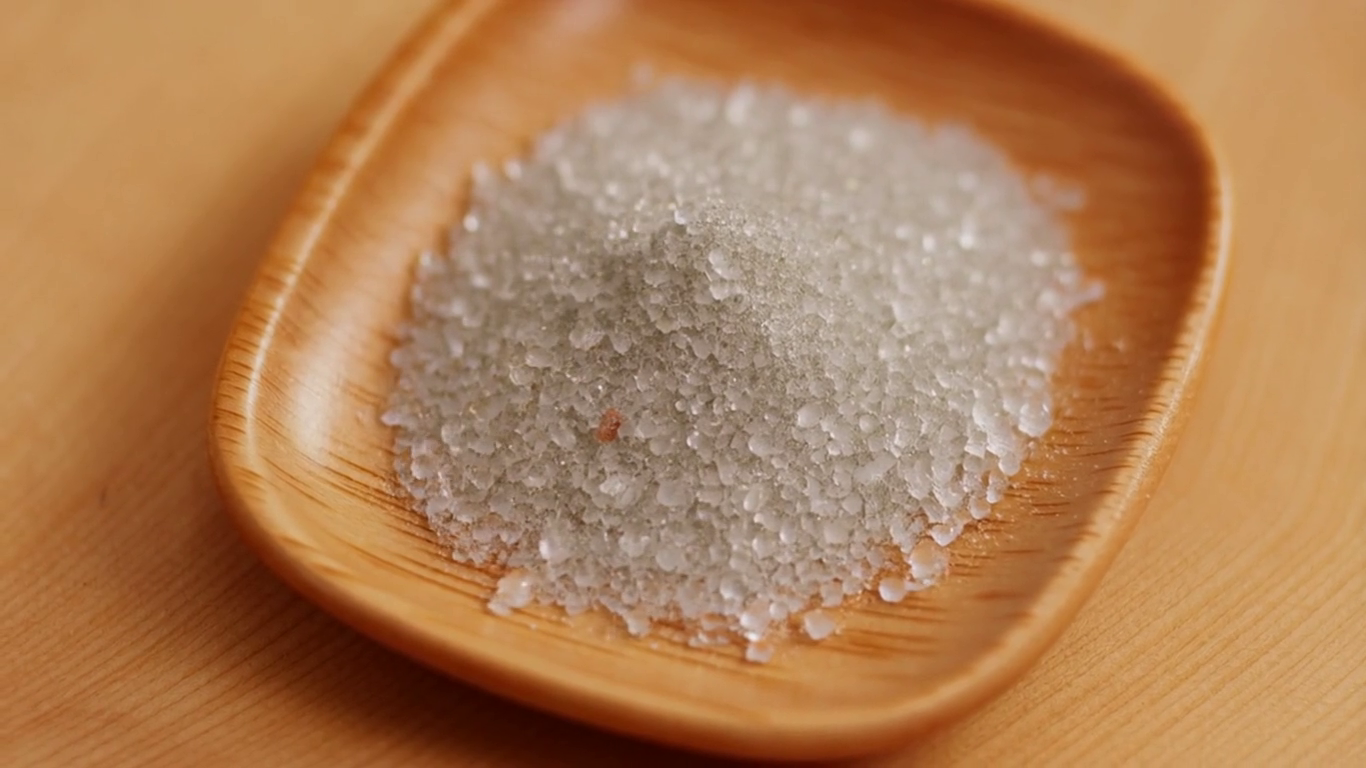}\hspace{1pt}%
  \includegraphics[height=1.2cm]{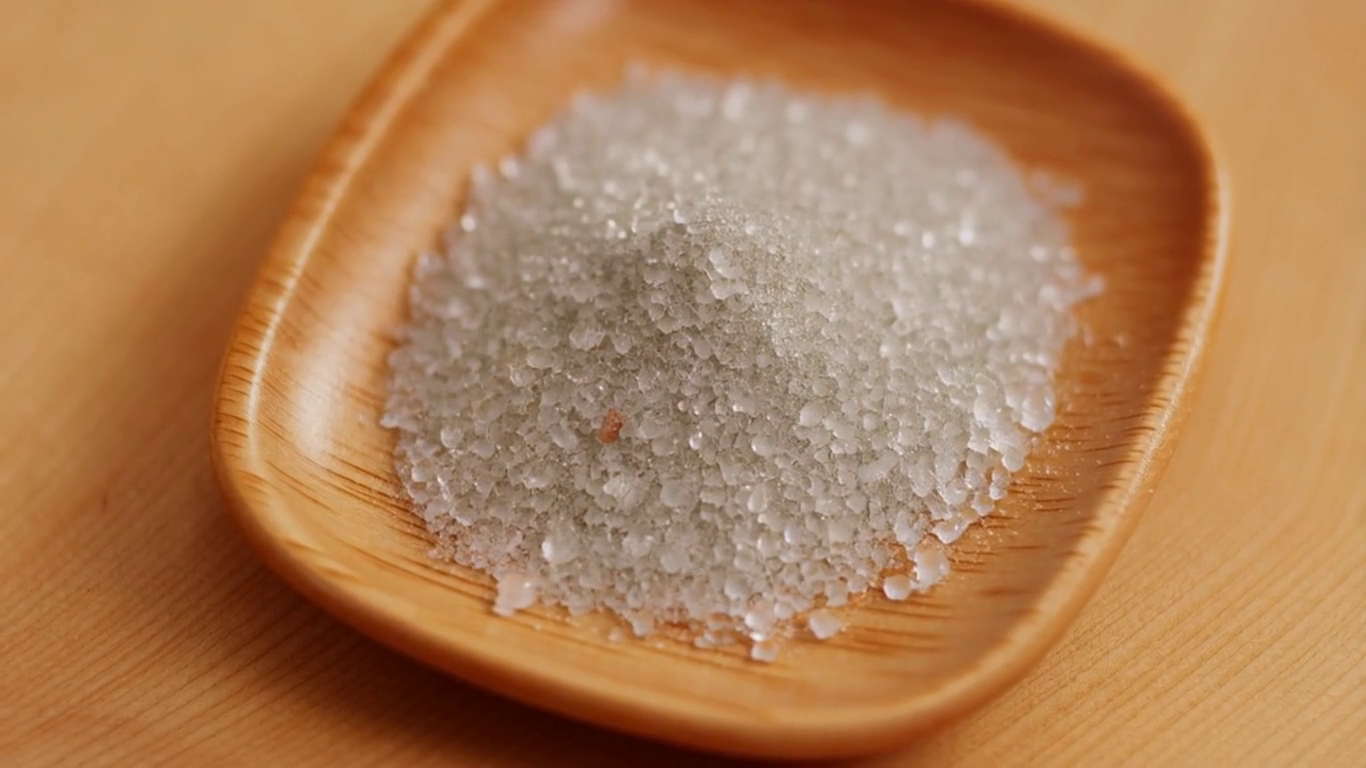}\hspace{1pt}%
  \includegraphics[height=1.2cm]{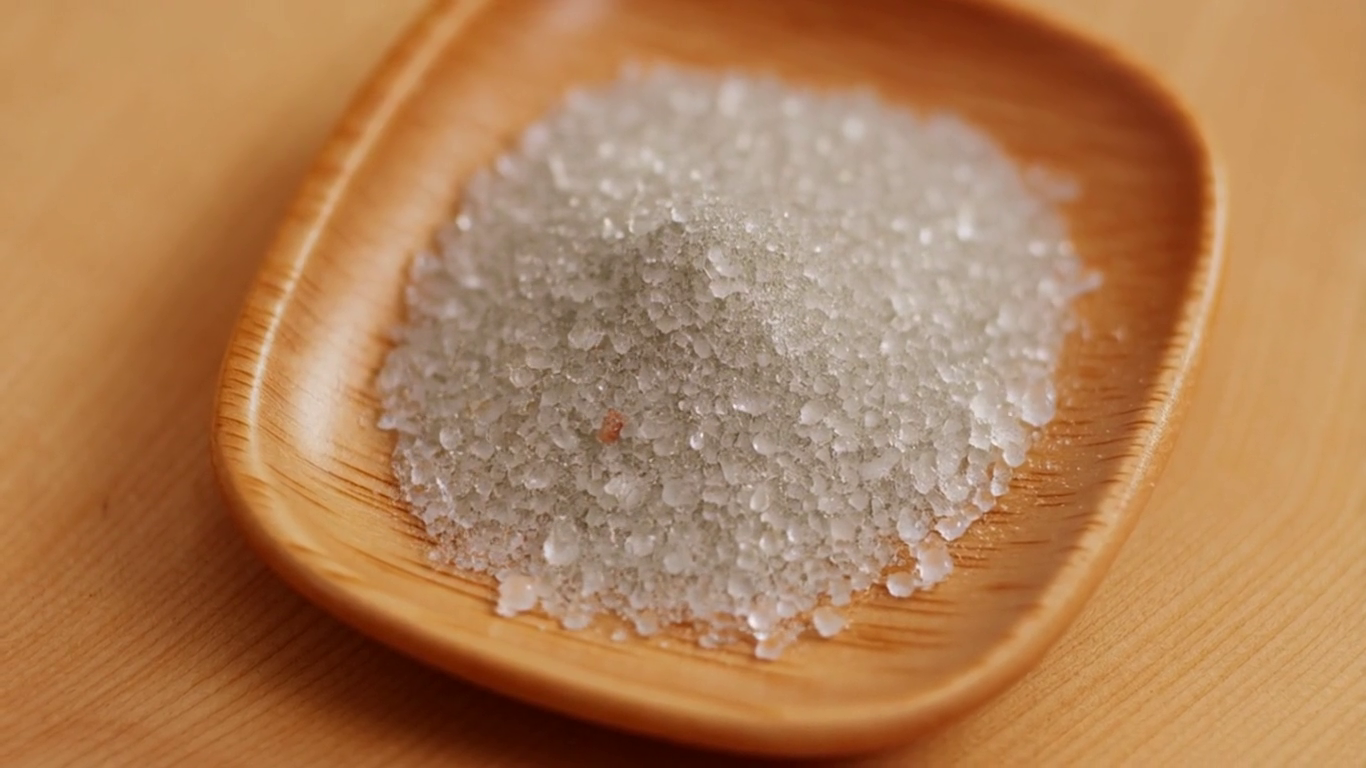}\hspace{1pt}%
  \includegraphics[height=1.2cm]{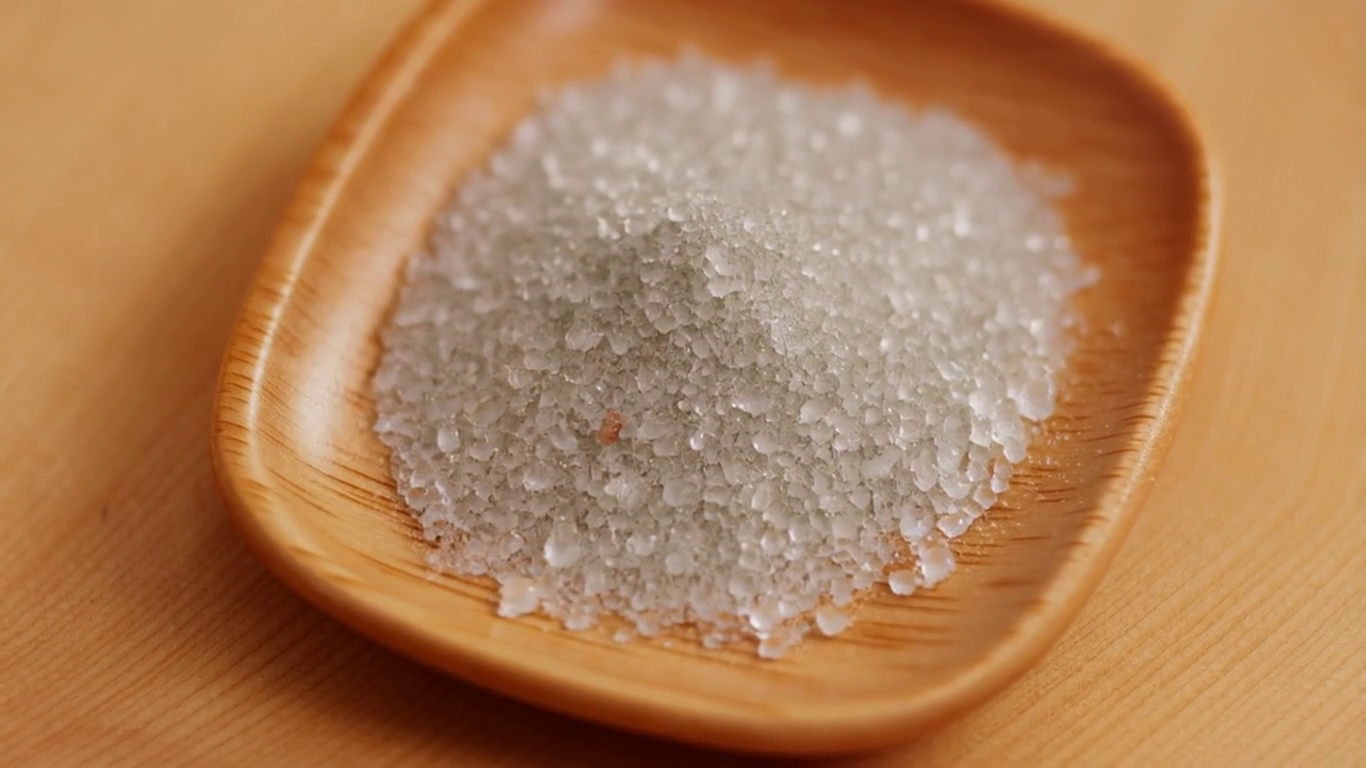}\hspace{1pt}%
}\\[0.1em]

\multicolumn{2}{c}{
  \textbf{PCS:} 2/4 \quad 
  \textbf{PCG:} 1/4 \quad 
  \textbf{CDN:} 4/4 \quad 
  \textbf{IMB:} 4/4 \quad 
  \textbf{STC:} 4/4
}\\

\midrule
\midrule

\multicolumn{2}{c}{\textbf{Rotating Cups with Balls}}\\
\midrule
\textbf{Prompt} & 
Two \textbf{\underline{\smash{plastic cups}}} are joined mouth-to-mouth, with a \textbf{\underline{wooden stick}} fixed along the outside connecting them. One \textbf{\underline{small ball}} is placed inside each cup, and the stick is \textbf{\underline{\smash{spun rapidly}}} around its center.\\[0.2em]

\textbf{Expected} & 
As the system spins, both \textbf{\underline{balls move outward}} and \textbf{\underline{\smash{press against the sides}}} of their cups. They \textbf{\underline{remain}} in that outward position while the rotation continues.\\
\midrule

\multicolumn{2}{c}{\cellcolor{red!10}\textbf{Sora-2:} {\color{red!70!black}Failed to Generate Expected Phenomenon}}\\[0.1em]

\multicolumn{2}{c}{
  \includegraphics[height=1.2cm]{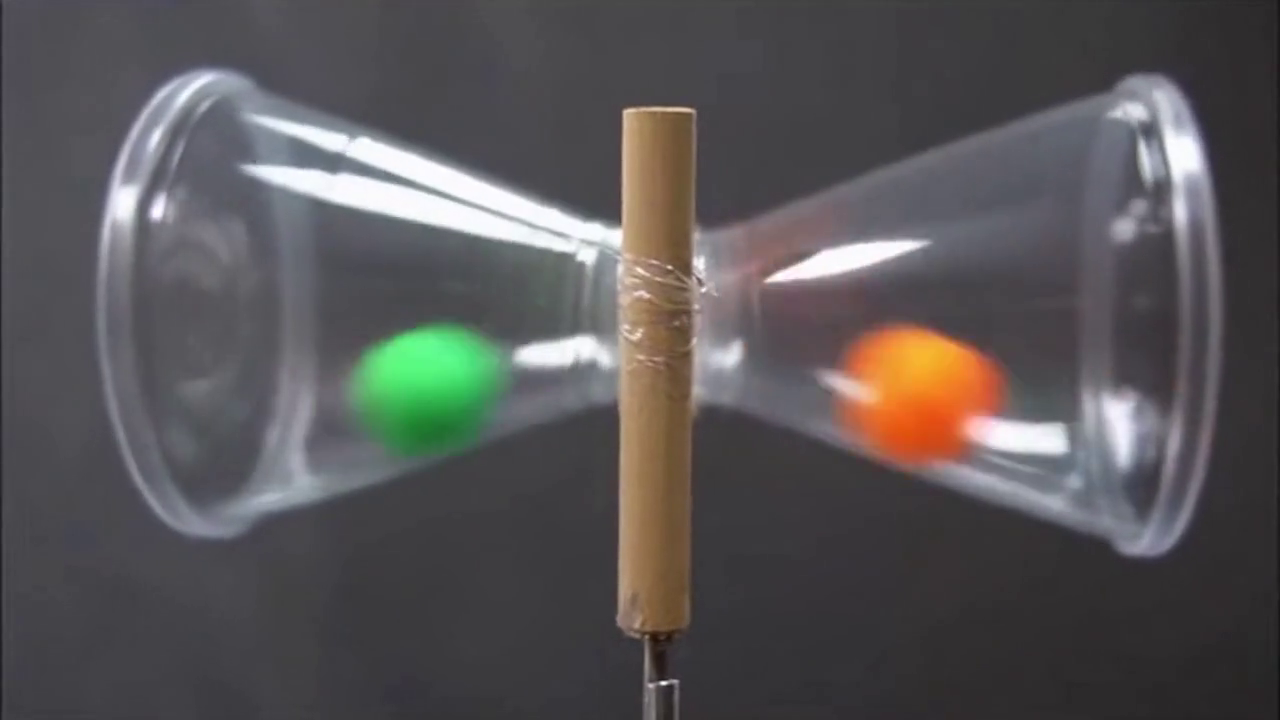}\hspace{1pt}%
  \includegraphics[height=1.2cm]{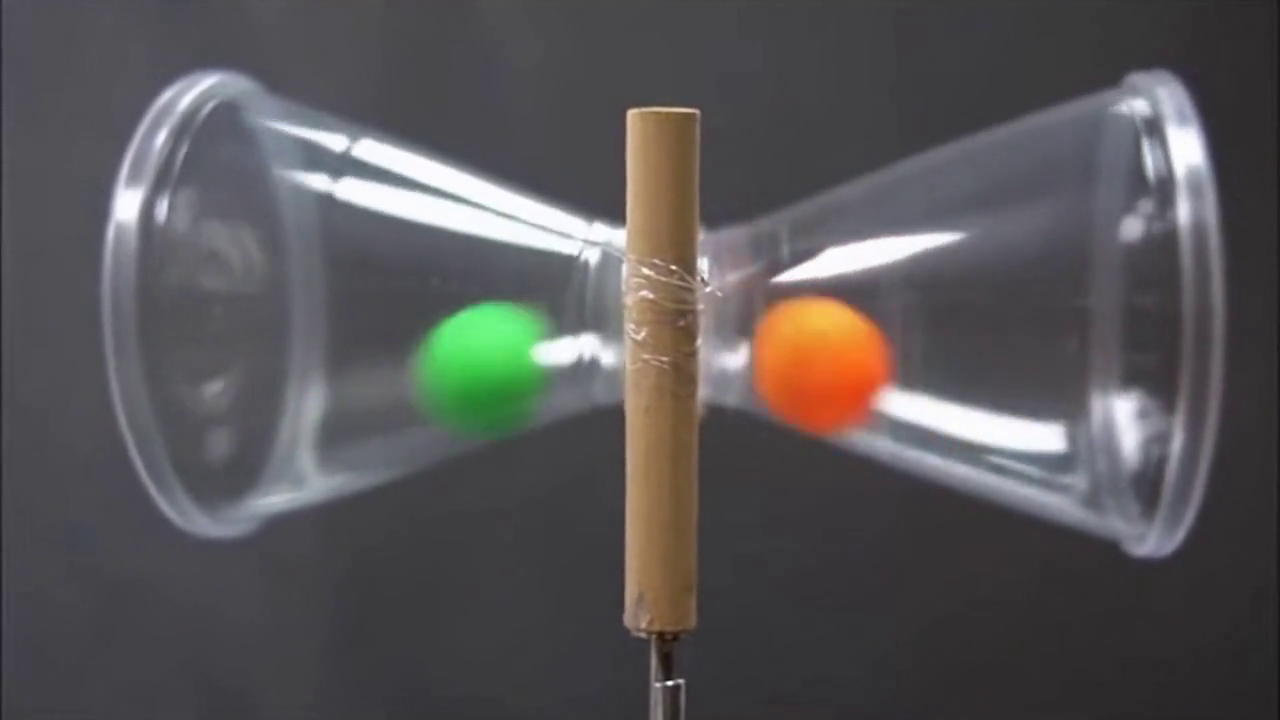}\hspace{1pt}%
  \includegraphics[height=1.2cm]{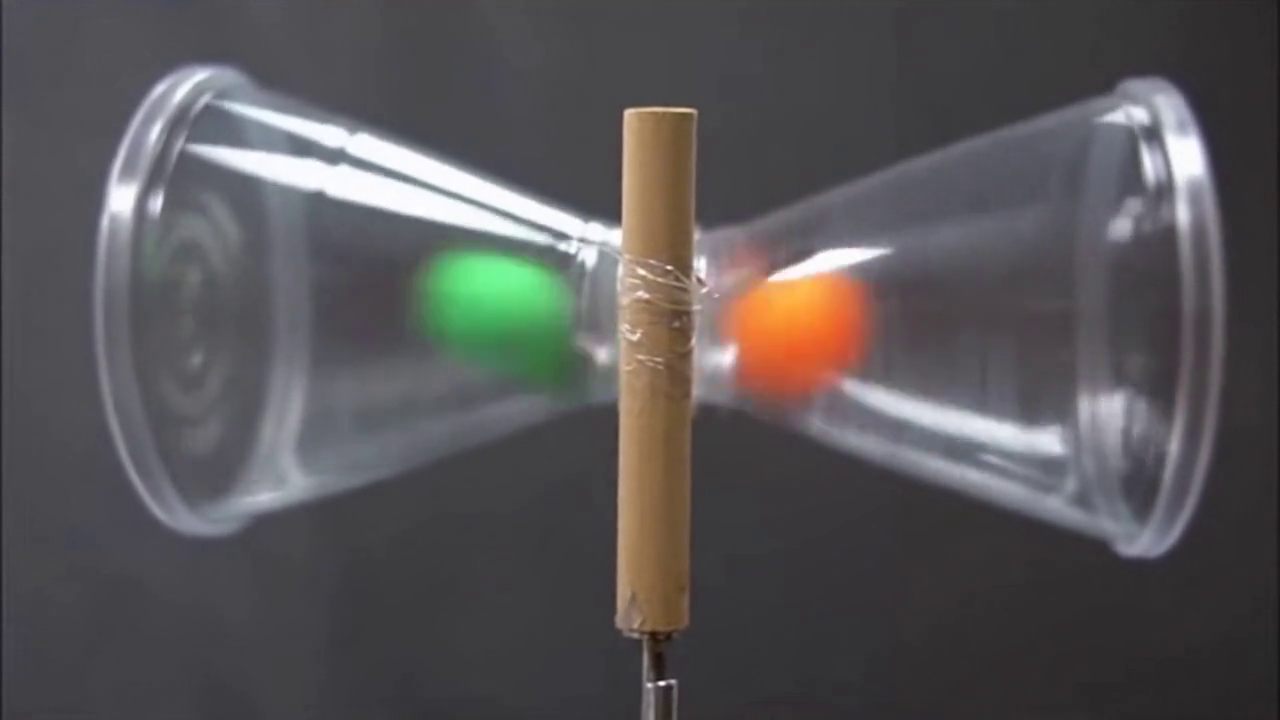}\hspace{1pt}%
  \includegraphics[height=1.2cm]{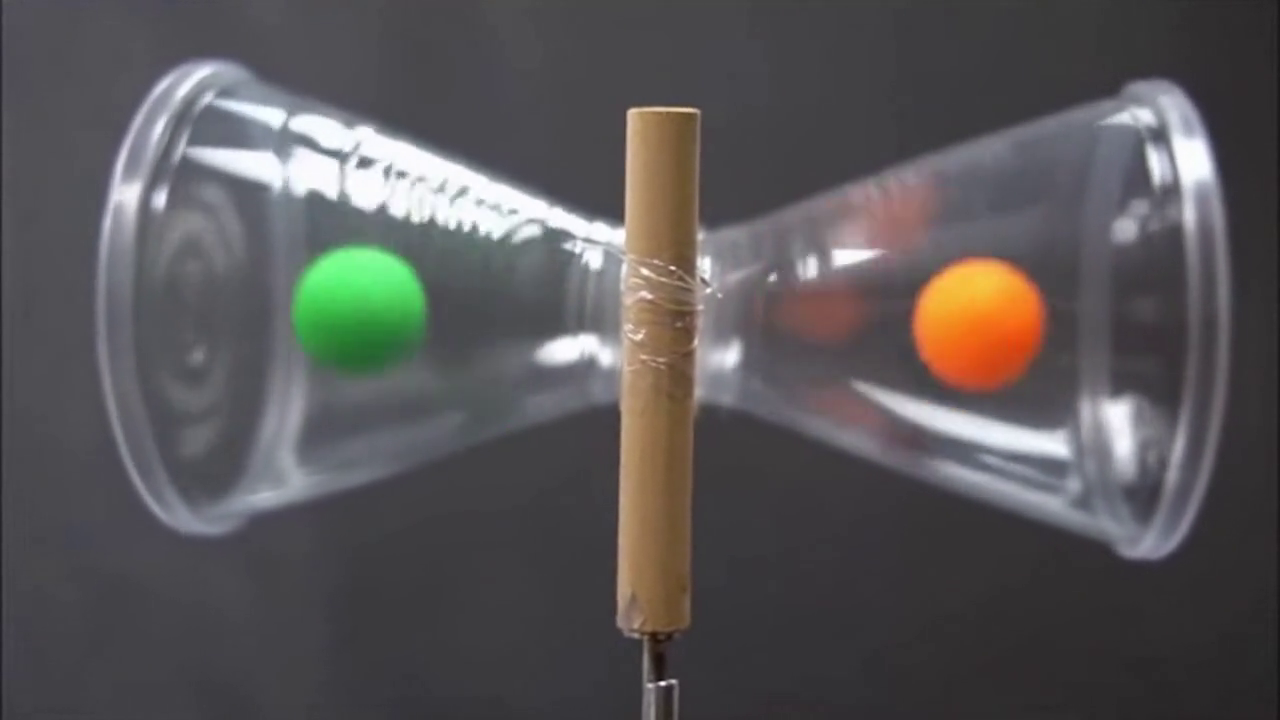}\hspace{1pt}%
  \includegraphics[height=1.2cm]{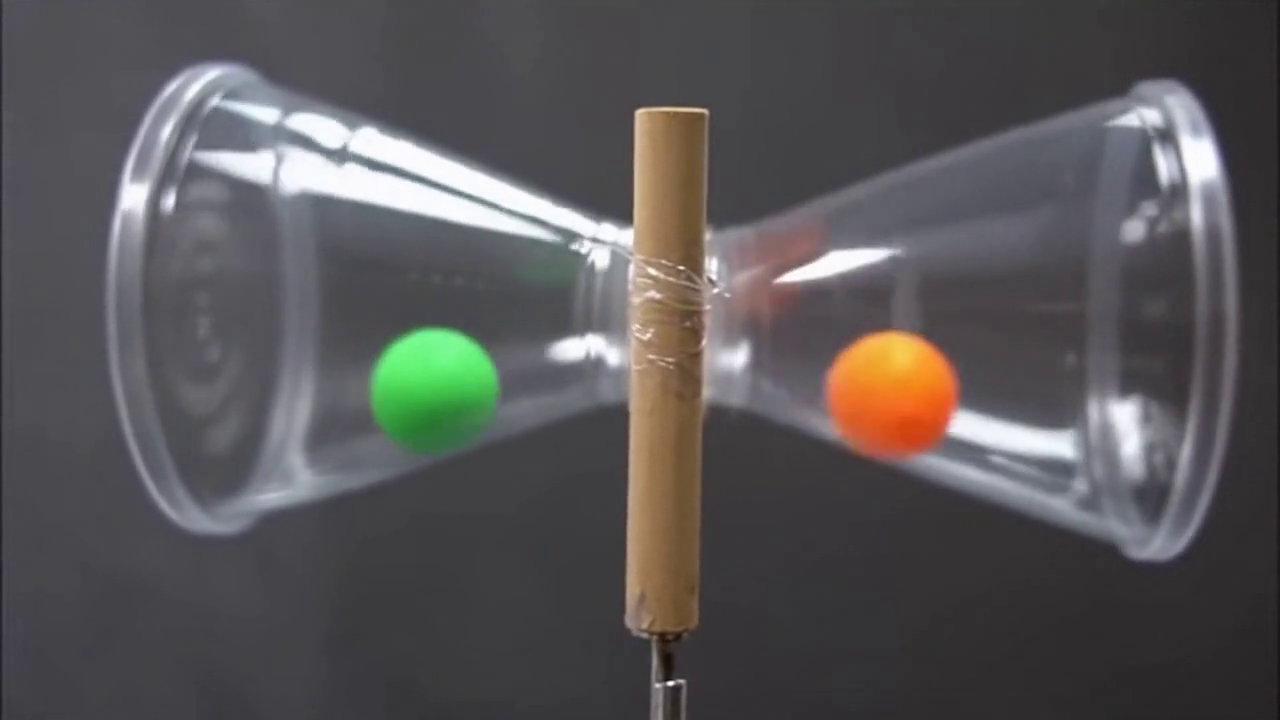}\hspace{1pt}%
  \includegraphics[height=1.2cm]{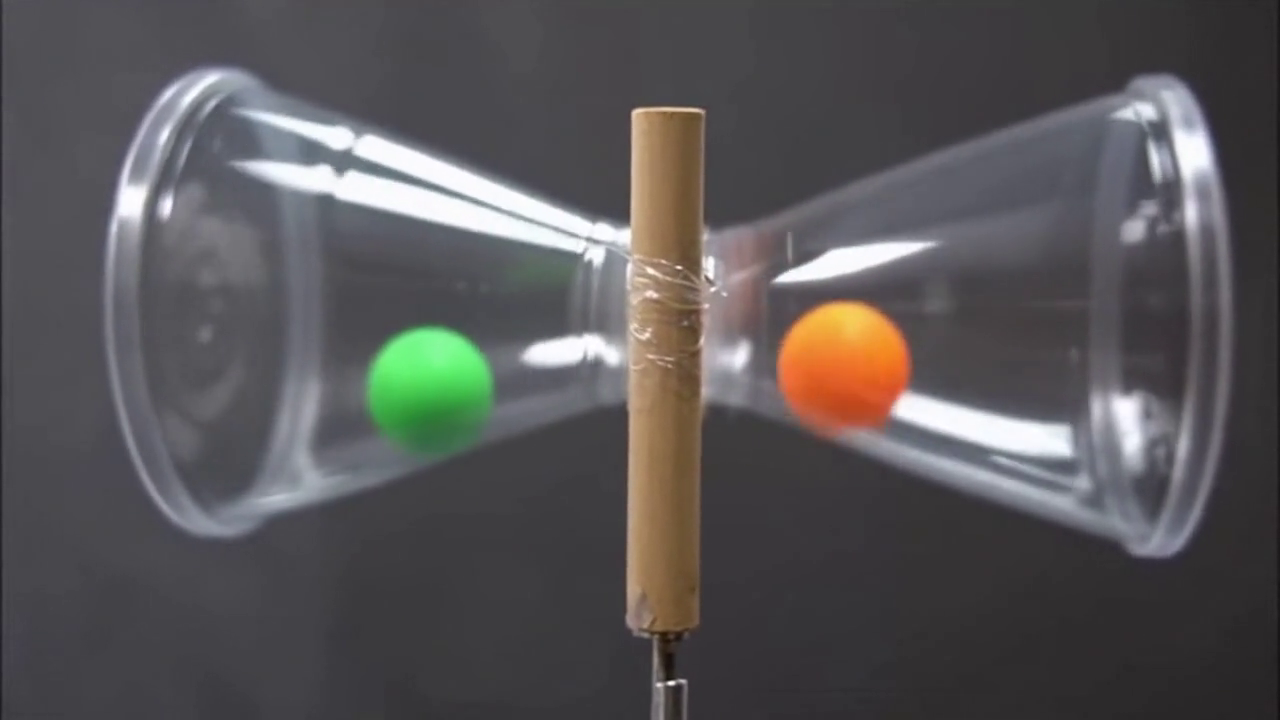}\hspace{1pt}%
  \includegraphics[height=1.2cm]{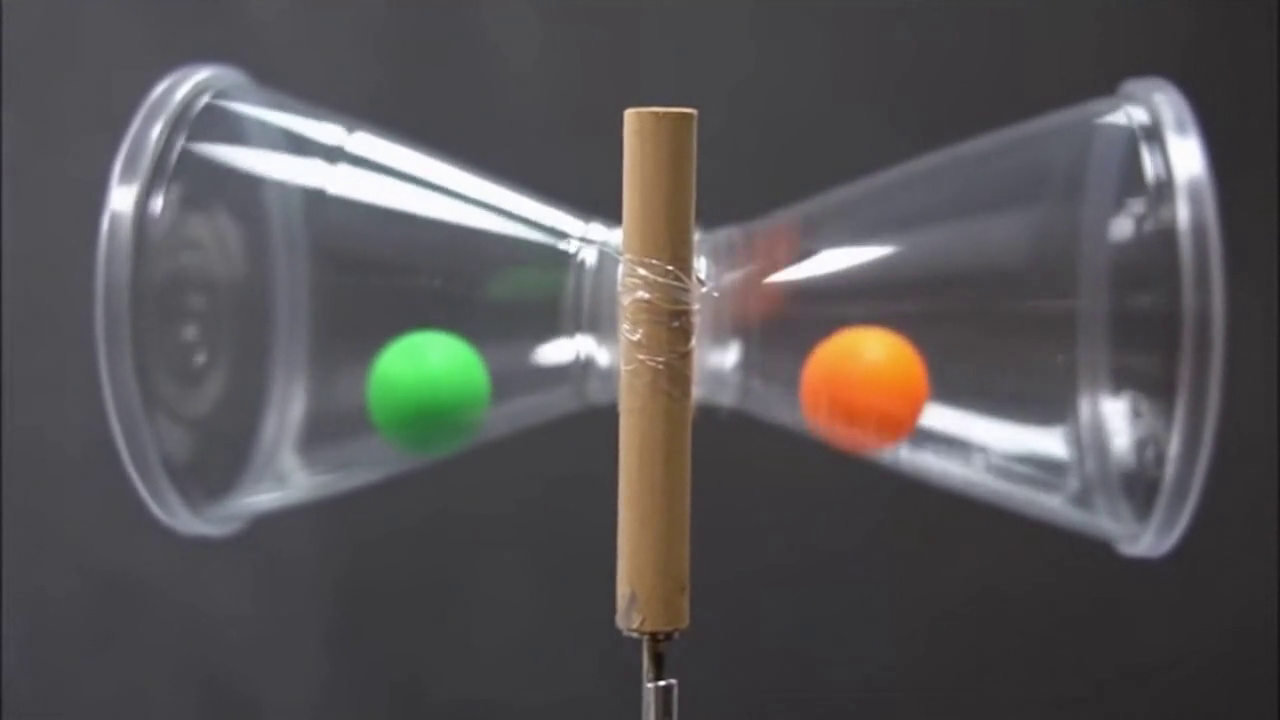}
}\\[0.1em]

\multicolumn{2}{c}{
  \textbf{PCS:} 2/4 \quad 
  \textbf{PCG:} 1/4 \quad 
  \textbf{CDN:} 3/4 \quad 
  \textbf{IMB:} 4/4 \quad 
  \textbf{STC:} 4/4
}\\
\midrule

\multicolumn{2}{c}{\cellcolor{red!10}\textbf{Veo-3:} {\color{red!70!black}Failed to Generate Correct Experiment Setup}}\\[0.1em]

\multicolumn{2}{c}{
  \includegraphics[height=1.2cm]{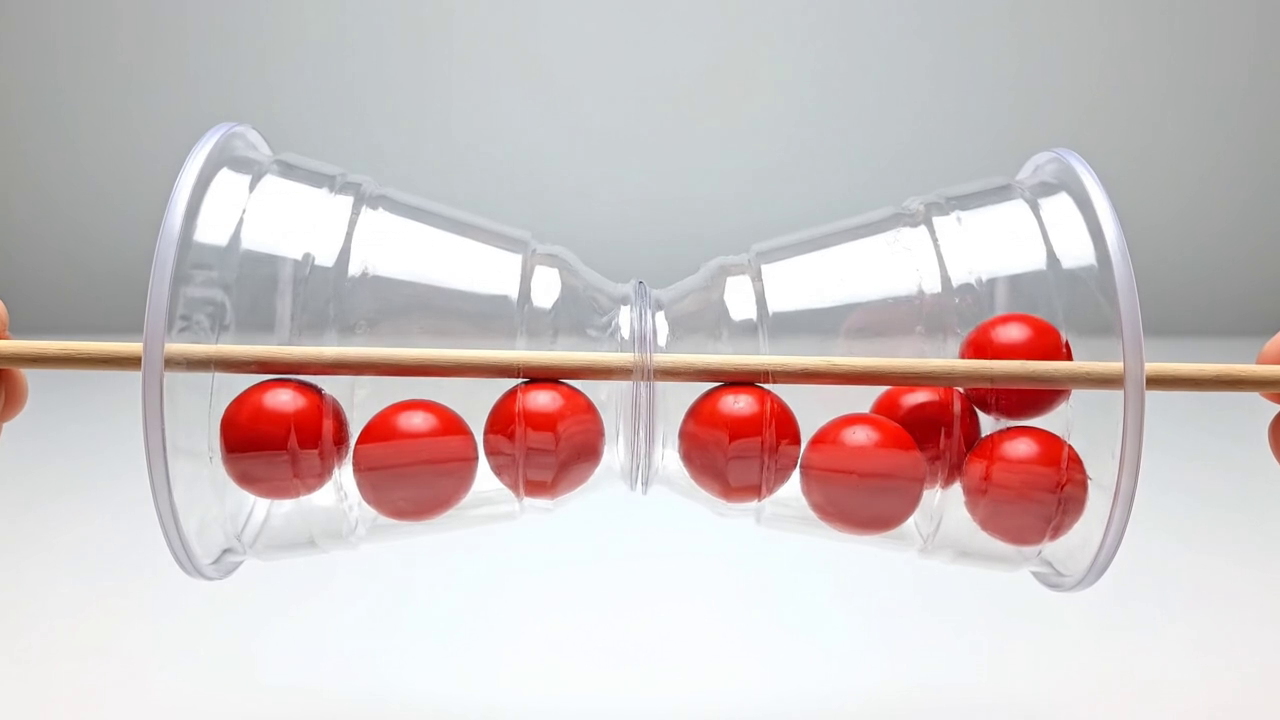}\hspace{1pt}%
  \includegraphics[height=1.2cm]{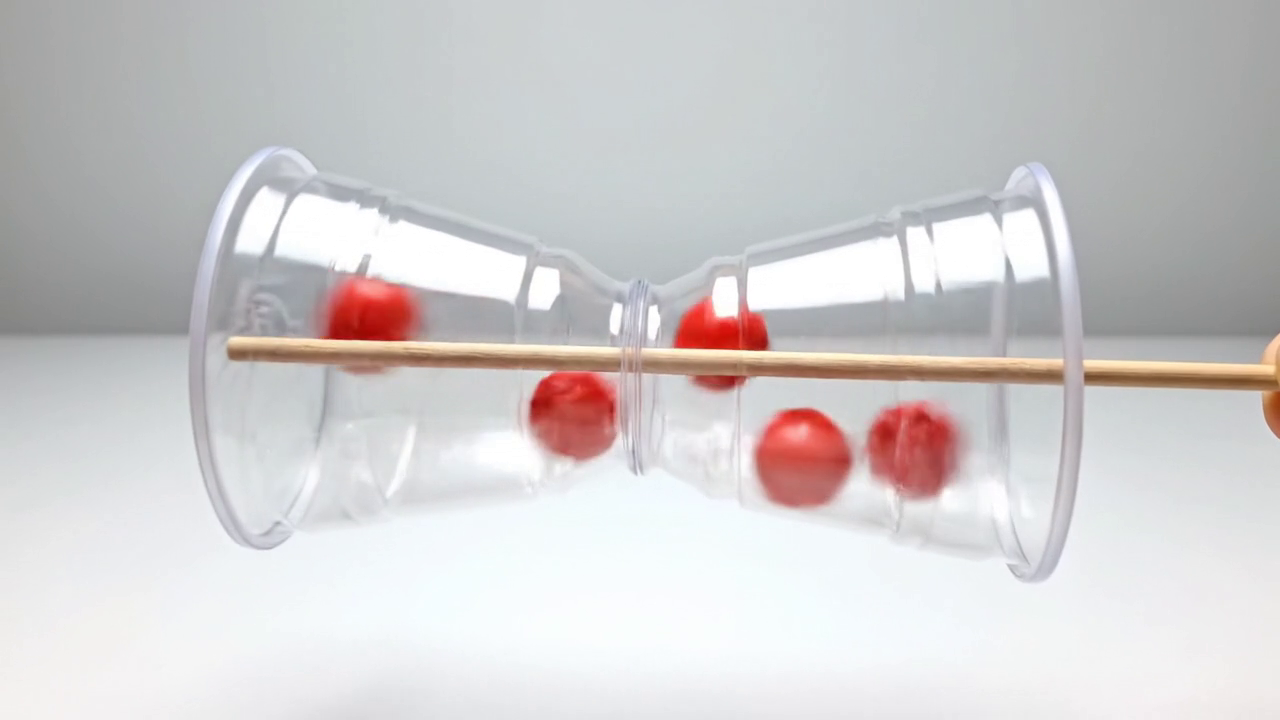}\hspace{1pt}%
  \includegraphics[height=1.2cm]{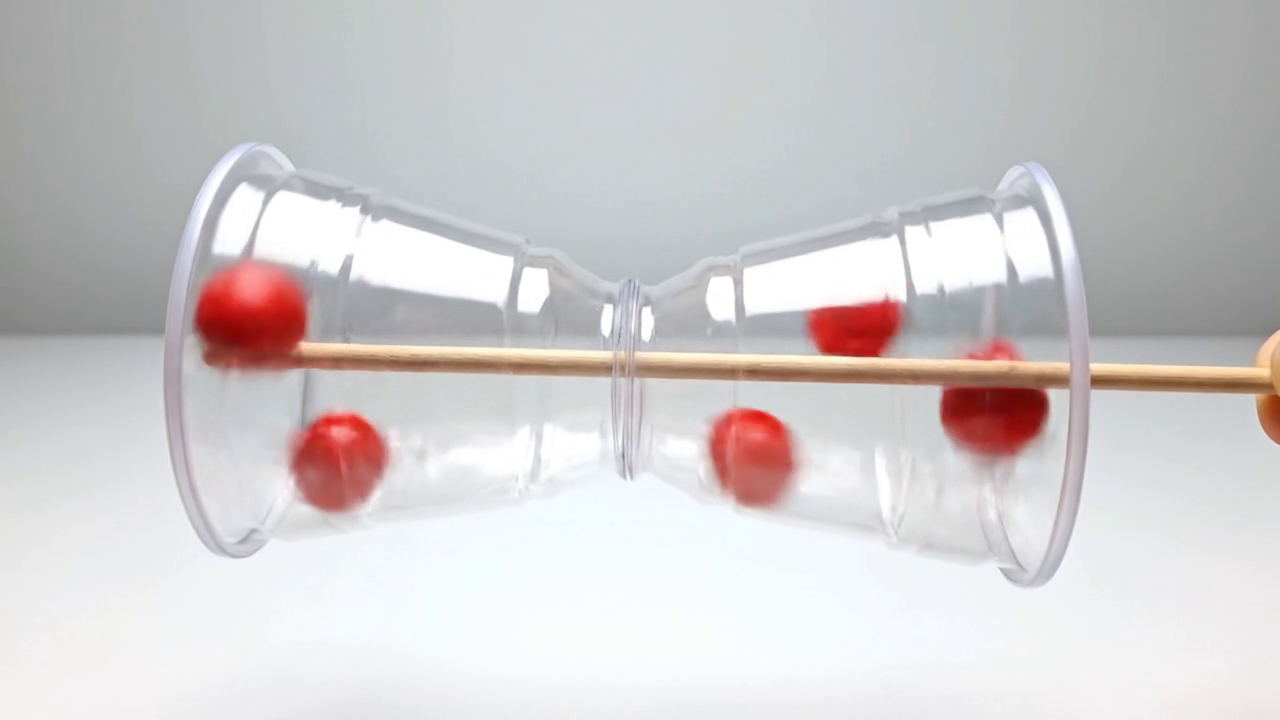}\hspace{1pt}%
  \includegraphics[height=1.2cm]{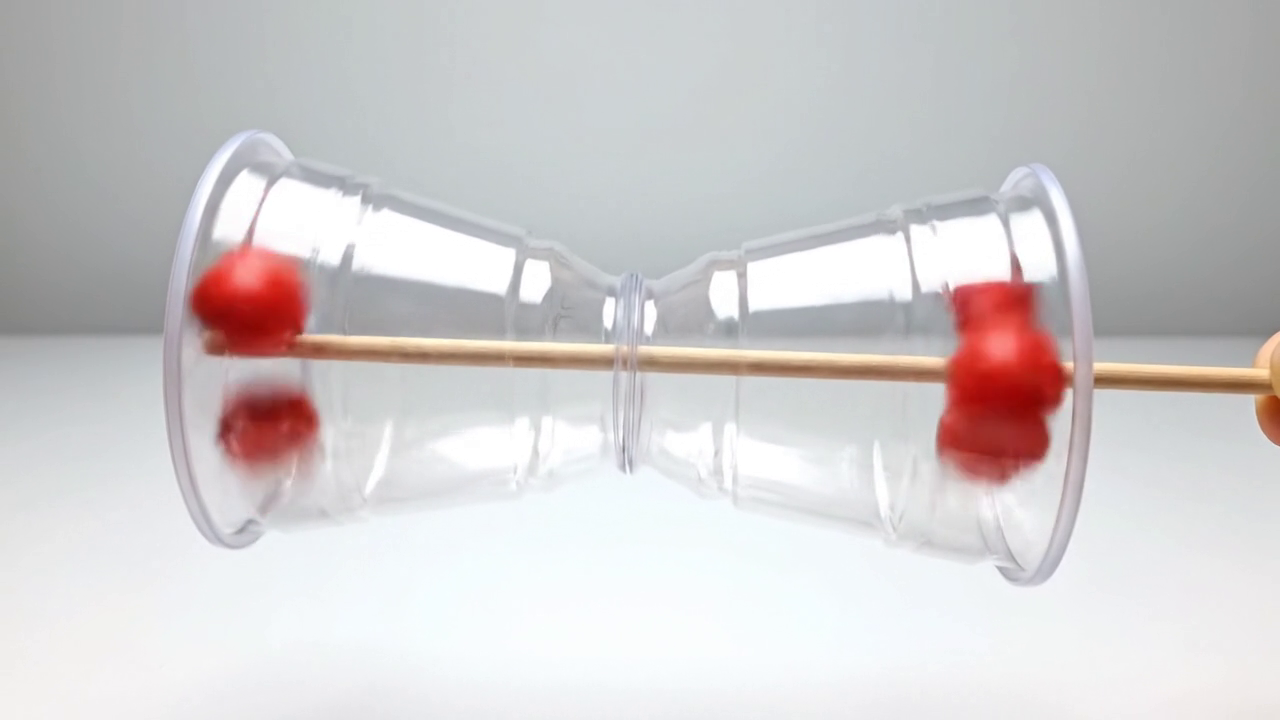}\hspace{1pt}%
  \includegraphics[height=1.2cm]{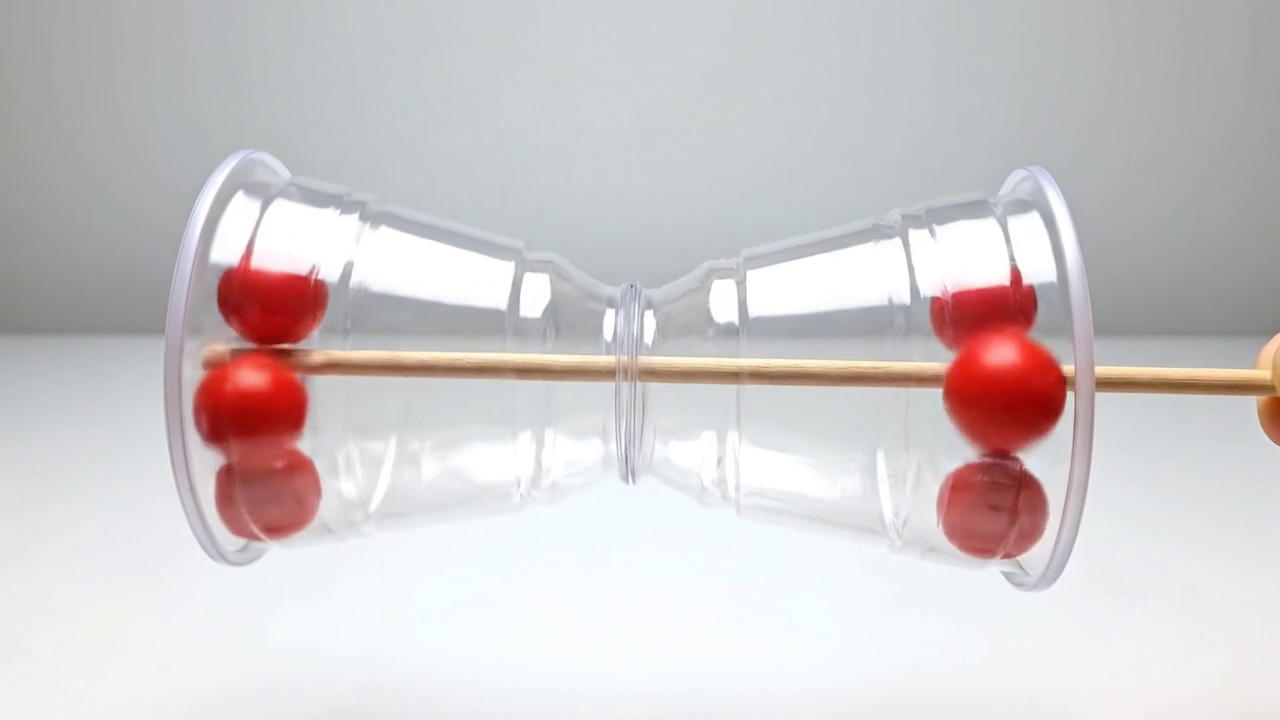}\hspace{1pt}%
  \includegraphics[height=1.2cm]{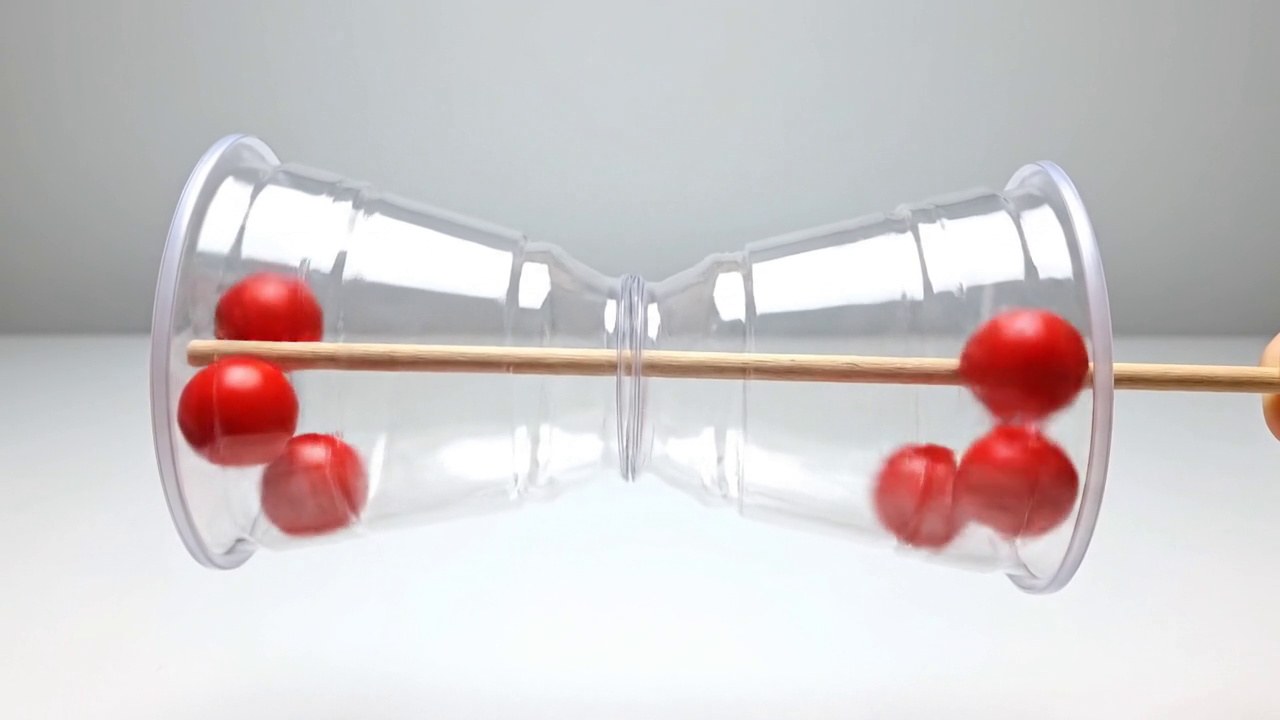}\hspace{1pt}%
  \includegraphics[height=1.2cm]{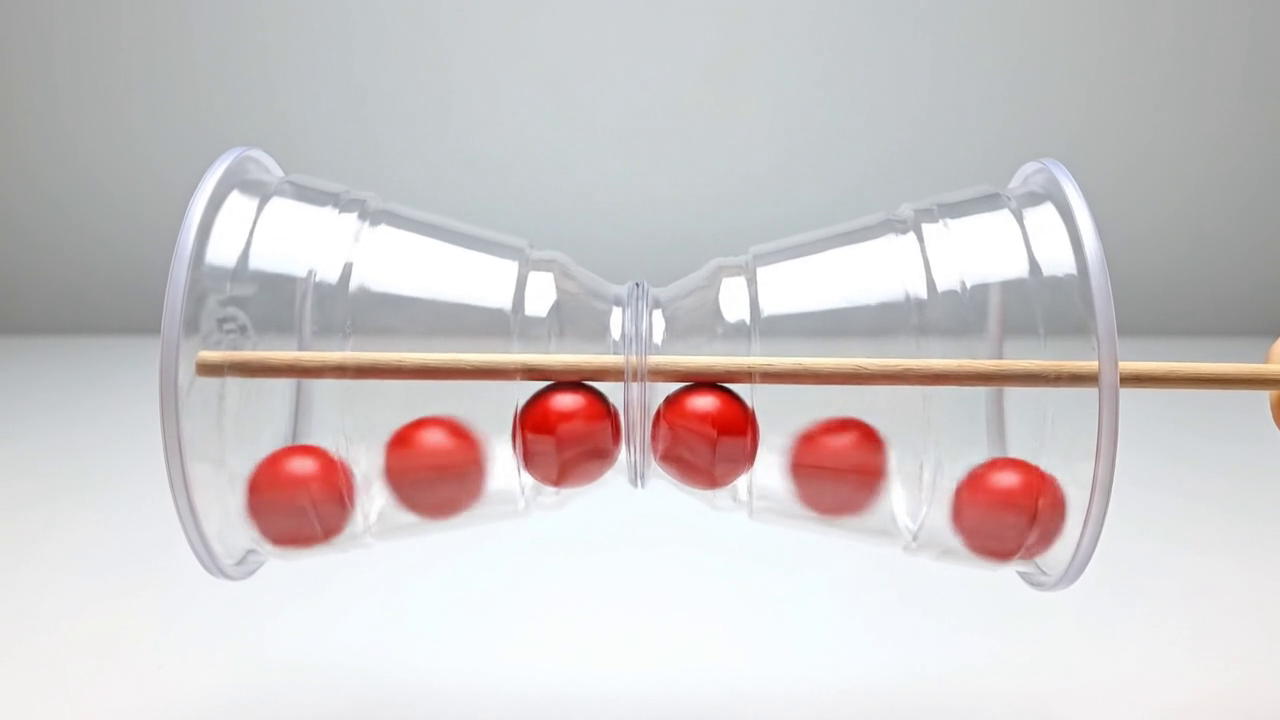}
}\\[0.1em]

\multicolumn{2}{c}{
  \textbf{PCS:} 1/4 \quad 
  \textbf{PCG:} 2/4 \quad 
  \textbf{CDN:} 1/4 \quad 
  \textbf{IMB:} 2/4 \quad 
  \textbf{STC:} 2/4
}\\

% \multicolumn{2}{c}{\cellcolor{red!10}\textbf{Hailuo-2.3:} {\color{red!70!black}Failed to Generate Expected Phenomenon}}\\[0.1em]

% \multicolumn{2}{c}{
%   \includegraphics[height=1.2cm]{figures/ex143_hailuo_01.png}\hspace{1pt}%
%   \includegraphics[height=1.2cm]{figures/ex143_hailuo_02.png}\hspace{1pt}%
%   \includegraphics[height=1.2cm]{figures/ex143_hailuo_03.png}\hspace{1pt}%
%   \includegraphics[height=1.2cm]{figures/ex143_hailuo_04.png}\hspace{1pt}%
%   \includegraphics[height=1.2cm]{figures/ex143_hailuo_05.png}\hspace{1pt}%
%   \includegraphics[height=1.2cm]{figures/ex143_hailuo_06.png}\hspace{1pt}%
%   \includegraphics[height=1.2cm]{figures/ex143_hailuo_07.png}
% }\\[0.1em]

% \multicolumn{2}{c}{
%   \textbf{PCS:} 3/4 \quad 
%   \textbf{PCG:} 1/4 \quad 
%   \textbf{CDN:} 3/4 \quad 
%   \textbf{IMB:} 4/4 \quad 
%   \textbf{STC:} 3/4
% }\\
\bottomrule
\end{tabular}

\caption{Comparison of video generation models on \sysnamenospace. \textbf{Top:} Aluminum-iodine reaction testing chemical dynamics. Sora-2 correctly depicts the expected ignition phenomenon, while Hailuo-2.3 fails to generate the reaction. \textbf{Bottom:} Rotating cups with balls testing centrifugal force understanding. Both Sora-2 and Veo-3, the two highest-ranked models in our evaluation, fail to conduct correct experimental procedure and simulate the expected phenomenon accurately. Human Annotation Rating: Prompt Consistency (PCS), Phenomenon Congruency (PCG), Correct Dynamism (CDN), Immutability (IMB), and Spatio-Temporal Coherence (STC).}
\label{fig:qualitative_analysis_examples}
\end{figure*}
\section{Experiments}
\label{sec:exp}

\subsection{Model Evaluation.}

%\paragraph{ \lx{discuss human annotation results, and auto eval results.} }

\paragraph{Experiment Settings.} We evaluate seven state-of-the-art video models on \sysnamenospace-Bench: Sora-2~\citep{openai2025sora2}, Veo-3~\citep{deepmind2025veo3}, Seedance 1.0 Pro~\citep{seedance2025,seedance1pro2025}, Kling-v2.5-Turbo-Pro~\citep{kling_25_turbo_pro_2025}, Hailuo 2.3~\citep{minimax_hailuo_2_3_2025}, Ray2~\citep{luma_ray2_2025}, and Wan-2.5-T2V-Preview~\citep{wan2025wan2.5preview}. We generate three videos for each model in all of our experiments. Only one video is generated from Veo-3 due to the prohibitively high API cost.

\paragraph{Human Annotation Results.}

% \input{figure_text/model_wise_radar}

% We ask domain experts to rate all video outputs based on the five-dimensional rubrics (defined in section~\ref{para:rubrics}).

% Based on these annotations, we computed both per-criterion and aggregate model scores. This human-grounded evaluation serves as the reference signal for analyzing video models' scientific-reasoning performance and the effectiveness of \sysnamenospace-Judge by   
% %calibrating automatic VLM-as-a-Judge metrics and for 
% correlating judge predictions with expert assessments. The annotated model performance is summarized in Figure~\ref{fig:model_wise_radar}, and weighted average scores across the five dimensions are shown in the legend. We assign the weights: \emph{Phenomenon Congruency} (0.3), \emph{Prompt Consistency} (0.2), \emph{Spatio-temporal Coherence} (0.2), \emph{Correct Dynamism} (0.15), and \emph{Immutability} (0.15), to reflect the relative importance of accurately capturing the expected phenomenon, experimental setup, and procedural integrity. 
We asked domain experts to rate video outputs of all the models based on the five-dimensional rubrics (defined in section~\ref{para:rubrics}). Using the human annotations, we calculated both per-criterion and aggregate model scores. This human-grounded evaluation serves as a reference for analyzing the scientific reasoning performance of video models and the effectiveness of \sysnamenospace-Judge. It allows us to calibrate automatic VLM-as-a-Judge metrics and correlate judge predictions with expert assessments. 

The performance of the annotated models is summarized in Figure~\ref{fig:first_page_overview}, and the weighted average scores across the five dimensions are presented in the legend. The weights assigned are as follows: \emph{Phenomenon Congruency} (0.3), \emph{Prompt Consistency} (0.2), \emph{Spatio-temporal Coherence} (0.2), \emph{Correct Dynamism} (0.15), and \emph{Immutability} (0.15). These weights reflect the relative importance of accurately capturing the expected phenomenon, ensuring proper experimental setup, and maintaining procedural integrity.

\begin{table}[htbp]
\centering
\small
\resizebox{\columnwidth}{!}{
\begin{tabular}{lccccc}
\toprule
\textbf{Model} &
\textbf{PCS} &
\textbf{PCG} &
\textbf{CDN} &
\textbf{IMB} &
\textbf{STC} \\
\midrule
 Sora-2    & $\textbf{3.32} $ & $\textbf{2.56}$ & $\textbf{3.33}$ & $\textbf{3.73}$ & $\textbf{3.71}$ \\
 Veo-3       & $3.01$ & $2.35$ & $2.83$ & $3.30$ & $3.42$ \\
Kling-v2.5  & $2.77$ & $1.91$ & $2.75$ & $3.36$ & $3.60$ \\
Wan-2.5    & $2.87$ & $1.84$ & $2.83$ & $3.36$ & $3.46$ \\
Seedance-1.0-Pro & $2.56$ & $1.78$ & $2.52$ & $3.15$ & $3.46$ \\
Hailuo-2.3  & $2.39$ & $1.67$ & $2.57$ & $3.16$ & $3.46$ \\
Ray2      & $1.65$ & $1.26$ & $2.13$ & $2.44$ & $2.92$ \\
\bottomrule
\end{tabular}
}
\caption{Raw human annotation scores (mean) per model and score dimension (1--4 Likert scale). Prompt Consistency (PCS), Phenomenon Congruency (PCG), Dynamism (CDN), Immutability (IMB), and Coherence (STC).}
\vspace{-0.5cm}
\label{tab:raw_human_annotation_scores}
\end{table}

The results from human annotations are summarized in Table~\ref{tab:raw_human_annotation_scores} (raw scores) and Table~\ref{tab:normalized_annotation_scores} (min-max normalized scores). Table~\ref{tab:raw_human_annotation_scores} show that while most models perform well in maintaining \emph{Spatio-temporal Coherence} and \emph{Immutability}, qualities commonly emphasized in existing video benchmarks~\citep{huang2024vbench, huang2024vbenchpp, han_2025_videobench_cvpr}, many still struggle to generate physically correct phenomena (\emph{Phenomenon Congruency}) or adhere to other fundamental physical laws (\emph{Correct Dynamism}). On \emph{Phenomenon Congruency}, even the strongest closed-source systems, such as Sora-2 and Veo-3, achieve only approximately 64\% and 58.7\% on the Likert scale, respectively, while the strongest open-source model, Wan-2.5, reaches only around 46\%. Achieving high \emph{Prompt Consistency} also remains challenging: even the best-performing models, such as Sora-2 and Veo-3, with average scores above 3, oftentimes fail to produce accurate experimental setups and procedures, frequently fail to produce accurate experimental setups and procedures, particularly in more complex scenarios involving interactions among multiple key objects.

%Together, these components establish a consistent, interpretable, and physically grounded benchmark for evaluating scientific reasoning in video generation models.

\paragraph{\sysnamenospace-Judge Results.}
%We test different settings of \sysnamenospace-Judge's on the same set of video generations and the results are reported in Table~\ref{tab:main_results}. In comparison with other baselines summarized in the table, . For each evaluation reported in the table, their ranking correlations with human annotations are measured in Kendall’s $\tau$ and Spearman’s $\rho$. In comparison with other baselines summarized in the table, we can see rankings reported from \sysnamenospace-Judge have closest alignment with human annotations on our test suite.

We evaluate different configurations of \sysnamenospace-Judge on the same set of video generations, with results summarized in Table~\ref{tab:auto_eval_results}. For each method including the baselines, we compute ranking correlations with human annotations using Kendall’s~$\tau$~\citep{kendall1938tau} and Spearman’s~$\rho$~\citep{spearman1904rho}. Compared to other baselines in the table, \sysnamenospace-Judge achieves the highest alignment with human evaluations, demonstrating its effectiveness in capturing human-judged quality across our test suite.

\begin{table*}[ht]
    \centering
    \resizebox{\linewidth}{!}{
    \begin{tabular}{cccc|ccccc}
    \toprule
        Model & VSci-Judge    & VSci-Judge (CL)  & VSci-Judge (CL+CV)& PhyGenEval & VideoScore2 & T2VBench  & LMArena-T2V  \\
        \midrule
        Sora-2               & \textbf{0.76}   &  \textbf{0.75}      & \textbf{0.76}  & \underline{0.43} & 0.72 & \underline{0.79} & \underline{1319} \\ 
        Veo-3                & \underline{0.69}  & \underline{0.65}  &  \underline{0.65}   & \textbf{0.47} & \underline{0.78} & \textbf{0.80} & \textbf{1361} \\ 
        Kling-v2.5-Turbo-Pro & 0.63          &  0.59     & 0.59   & 0.29 & 0.77 & 0.78 & 1226 \\
        Wan-2.5-T2V-Preview  & 0.66          &   0.60    & 0.59    & 0.41 & \textbf{0.79} & \underline{0.79} & 1128 \\
        Seedance 1.0 Pro     & 0.63        &    0.56      & 0.54   & 0.33 & 0.77 & 0.78 & 1191 \\
        Hailuo 2.3           & 0.58           &     0.52     & 0.50  & 0.29 & 0.72 & 0.77 & 1253 \\
        Ray2                 & 0.54             &     0.39    & 0.34    & 0.29 & 0.67 & 0.76  & 1062 \\
        \midrule
        $\tau$ ($\uparrow$) & 0.81 & 0.90 & 0.90  &  0.52 & 0.24 & 0.62 & 0.52 \\
        $\rho$ ($\uparrow$) & 0.89 & 0.96 & 0.96 &  0.61 & 0.29 & 0.75 & 0.71 \\
        \bottomrule
    \end{tabular} }
    \caption{
        We report our VLM-as-a-judge evaluation scores (using GPT-5 pro) for competing video models on \sysnamenospace-Bench, and their ranking correlations to human annotations. ``VSci-Judge" refers to \sysnamenospace-Judge, ``CL" to Checklist and ``CV" to CV-augmented reasoning. For each model, min-max normalization is applied to annotation scores and scores from other benchmarks before averaging across all test cases and runs (except for LMArena-T2V). For each evaluation, the highest score is \textbf{bolded} and the second is \underline{underlined}.
    }
    \label{tab:auto_eval_results}
\end{table*}

%LMARENA

%[
%  ["Veo-3",                1361],
%  ["Sora-2",               1319],
%  ["Hailuo 2.3",           1253],
%  ["Kling-v2.5-Turbo-Pro", 1226],
%  ["Seedance 1.0 Pro",     1191],
%  ["Wan-2.5-T2V-Preview",  1128],
%  ["Ray2",                 1062]
%]

\subsection{Quantitative Analysis}
\label{sec:exp-correlational-analysis}

%\subsubsection{Effectiveness of VLM-as-a-Judge}

\subsubsection{\sysnamenospace-Judge Design Choice}
\label{sec:exp-judge-design}

\paragraph{Model Choice.}
%\paragraph{ \lx{false positive mitigation $\to$ checklist, and CV-based} }
We evaluate the performance of \sysnamenospace-Judge using both non-reasoning and reasoning models as the backbone VLM judge. As shown in Table~\ref{tab:sc_judge_gaps}, non-reasoning models such as GPT-4o exhibit substantial deviations from human annotations, with a strong tendency to assign inflated scores, leading to a high false-positive rate. Replacing GPT-4o with a reasoning model like GPT-5 pro mitigates the inflation by up to 57\% on \emph{prompt consistency}, yet the discrepancy remains noticeable, resulting in ranking misalignments with human judgments, as shown in the “VSci-Judge” column from Table~\ref{tab:auto_eval_results}.

%\paragraph{ \lx{model design choice $\to$ false positive phenomenon, reference to appendix} }
\paragraph{Configuration Choice.}
Table~\ref{tab:harness_ablation} compares evaluation scores from different configurations of \sysnamenospace-Judge against human annotations. With GPT-5 pro, the score inflation is a less prononced issue, but notable gaps remain in dimensions such as \emph{Dynamism}, \emph{Coherence}, and \emph{Immutability}, where the VLM still tends to overestimate performance. Incorporating a checklist as hint and requiring the model to reason each rating using evidence from key frames leads to a noticeable reduction in overestimation across multiple dimensions. Further integrating CV tools as evidence sources, particularly ByteTrackwhich enables reliable identity association across frames and provides concrete evidence for penalizing frame jumps or unexpected object transformations. This substantially narrows the scoring gaps, by 0.07 on \emph{Coherence} and 0.09 on \emph{Immutability} respectively, compared with the vanilla \sysnamenospace-Judge without the checklist.

\begin{table}[!htbp]
\centering
\resizebox{\columnwidth}{!}{
\begin{tabular}{lccc}
\toprule
\textbf{Dimension}  & w/o CL $\Delta$ & w/ CL $\Delta$ & w/ (CL +CV) $\Delta$\\
\midrule
Prompt Consis.        & $-0.10 \text{\scriptsize$\pm 0.10$}$  &  $-0.06 \text{\scriptsize$\pm 0.08$}$ &  $-0.05 \text{\scriptsize$\pm 0.09$}$ \\
Exp. Phenomenon      & $-0.22 \text{\scriptsize$\pm 0.10$}$ & $-0.20 \text{\scriptsize$\pm 0.11$}$ & $-0.20 \text{\scriptsize$\pm 0.11$}$ \\
Dynamism                & $+0.32 \text{\scriptsize$\pm 0.13$}$ & $+0.27 \text{\scriptsize$\pm 0.10$}$ & $+0.29 \text{\scriptsize$\pm 0.13$}$ \\
Coherence                & $+0.28 \text{\scriptsize$\pm 0.08$}$ & $+0.26 \text{\scriptsize$\pm 0.09$}$ & $+0.21 \text{\scriptsize$\pm 0.09$}$ \\
Immutability             & $+0.27 \text{\scriptsize$\pm 0.12$}$ & $+0.24 \text{\scriptsize$\pm 0.14$}$ & $+0.18 \text{\scriptsize$\pm 0.17$}$ \\
\bottomrule
\end{tabular}
}
\caption{Gap to human annotations from Table~\ref{tab:normalized_annotation_scores} (mean $\Delta \pm$ std) for each score dimension and VSci-Judge variants using GPT-5 pro. \textbf{Larger positive $\Delta$ indicates stronger over-estimation relative to annotators.}}
\label{tab:harness_ablation}
\vspace{-1.5em}
\end{table}

%\subsubsection{Comparison with PhyGenBench} 

\subsubsection{Comparison with Baselines}

\paragraph{PhyGenEval.}
We adopt the PhyGenEval framework from PhyGenBench \cite{meng2025phygenbench} to assess physical commonsense in \sysnamenospace videos, employing GPT-4o as our VLM for video understanding and physical commonsense assessment, and CLIP to compute visual-semantic alignment between generated videos and reference prompts, with scores averaged across all prompts within each domain to obtain category-specific (Table~\ref{tab:phygenbench_baseline_results}) and total performance metrics (Table~\ref{tab:auto_eval_results}). However, the significant correlation gap between VSci-Judge ($\rho$ = 0.89) and PhyGenEval ($\rho$ = 0.61) reveals fundamental limitations in the applicability of PhyGenEval to complex real-world video generation tasks. The three-tier hierarchical framework (key phenomena detection, physics order verification, overall naturalness) of PhyGenEval enforces rigid binary judgments at each stage, leading to score quantization and reduced discrimination among similar models, while its rule-based approach of checking explicit violations of discrete physical laws (e.g., "Does the ball fall downward?", "Does water flow correctly?") cannot capture holistic aspects of physical realism such as consistent motion trajectories, causal relationships across time, acceleration patterns, momentum conservation, and nuances of motion smoothness that humans naturally evaluate. These limitations are particularly pronounced for \sysnamenospace-Bench, which differs fundamentally from PhyGenBench's simplified, single-principle scenarios. \sysnamenospace-Bench involves multiple scientific concepts (e.g., multi-object interactions, fluid dynamics) where outcomes arise from combinations of laws rather than single principles, featuring cascading effects where second-order dynamics matter.

\paragraph{VideoScore2.
%\todo{add analysis — correlation with RM: video-score/video-score2 scores analysis} \forrest{on it} 
}

Although VideoScore2 \cite{he2025videoscore2} represents a comprehensive multi-dimensional evaluation framework covering visual quality, text-to-video alignment and physical consistency, it exhibits limited effectiveness when evaluating complex scientific reasoning in video generation models. As shown in Table~\ref{tab:auto_eval_results},
% the normalized aggregate scores range from 0.67 to 0.79, with Wan-2.5-T2V-Preview achieving the highest and Ray2 the lowest.
VideoScore2's correlations with human annotations ($\tau$=0.24, $\rho$=0.29) are noticeably lower than the proposed VSci-Judge ($\tau$=0.81, $\rho$=0.89) and its checklist-augmented variants ($\tau$=0.90, $\rho$=0.96). This gap can be attributed to the fact that VideoScore2's physical consistency dimension was designed to capture broad common-sense violations, rather than performing deeper scientific reasoning or evaluating physical dynamics. Our dataset, by contrast, emphasizes non-trivial physical processes and scientific constraints. Consequently, explicit physics decomposition through domain-specific checklists yields substantially more reliable rankings than general-purpose rationales, validating the need for specialized evaluation frameworks when assessing nuanced physical reasoning in video generation. VideoScore2 is therefore not well suited for our benchmark, which requires strict, scientifically grounded physical evaluation, resulting in weaker alignment with human judgments on this dataset.

\paragraph{T2V-CompBench.
%\todo{add analysis — correlation with VLM: phygenBench (we need to run 4o + CLIP)}
}

% We evaluate our full set of text-to-video models under the Consistent Attribute Binding metric from T2V-CompBench, with results shown in table above. Videos are generated with three independent runs per model, and Llava-1.5-7B is used as the automatic judge. The scores are normalized at the end for correlation analysis. The evaluation reveals normalized scores ranging from 0.75 to 0.8, with Veo-3.1 achieving the highest score and Ray2 achieving the lowest. We find that the automatic metric aligns well with human annotations, achieving moderate-to-strong rank correlation. These results suggest that although current models perform somewhat similarly on controlled attribute-binding tasks, SOTA models such as Veo-3 exhibit more stable object-attribute rendering, whereas models like Ray2 and Hailuo 2.3 struggle with maintaining attribute fidelity across frames for different runs.

While T2V-CompBench \cite{sun2024t2vcompbench} provides a comprehensive evaluation of compositional text-to-video generation across seven categories (attribute binding, spatial relationships, motion binding, action binding, and object interactions), VSci-Judge focuses specifically on complex physics reasoning with a methodologically distinct approach. Using Llava-1.5-7B as the judge for the attribute binding category of T2V-CompBench, it achieves correlations $\tau$ = 0.62 and $\rho$ = 0.75. In contrast, VSci-Judge achieves substantially higher human agreement ($\tau$=0.81, $\rho$=0.89).  This critical architectural difference demonstrates that T2V-CompBench's VLM component struggles to capture nuanced physics violations, whereas VSci-Judge's evidence-grounded VLM approach can integrate multiple signals to assess physical plausibility holistically.

%\paragraph{\todo{add discussion on LMArena T2V} \lx{will do. or somebody can help me write some analysis here.} } data retrieved date Nov 5th, 2025.

\subsection{Qualitative Analysis}
\label{sec:exp-qualitative-analysis}

We provide qualitative examples illustrating how different models perform. As shown in the top portion of Figure~\ref{fig:qualitative_analysis_examples}, Sora-2 generates the expected physical phenomenon with strong prompt consistency and accurate motion. The video follows the described setup closely, exhibits coherent temporal evolution, and captures the ignition and color dynamics characteristic of the reaction. In contrast, Hailuo-2.3, despite maintaining reasonable dynamism, immutability and spatial coherence, fails to respect key elements of the prompt, resulting in a static video where the expected phenomenon never emerges. These differences underscore that correct experimental setup and prompt grounding are essential for achieving high scores on our benchmark, even when other dimensions appear satisfactory.

In contrast to the successful cases, simpler physical scenarios still pose significant challenges for current models. The bottom portion of Figure~\ref{fig:qualitative_analysis_examples}
%, solvable at a high school physics level,
reveals that even Sora-2 and Veo-3, the two highest-ranked models in our evaluation, struggle significantly with a scenario where text models can readily infer the expected phenomenon. Sora-2 generates a setup that partially aligns with the prompt, but contains significant errors and fails to capture the expected phenomenon. Despite this, Sora-2 maintains relatively good dynamism, immutability, and spatio-temporal coherence. Veo-3 depicts the setup poorly, though it momentarily suggests the expected outward movement of the balls.
% Hailuo's setup is comparatively more accurate, though still with flaws. Crucially, Hailuo fails to represent the expected phenomenon, showing the balls suspended in mid-air rather than pressed against the cup walls.
This failure highlights that even seemingly simple physical setups remain challenging under our benchmark.

\section{Related Work}
% Daniel Zhao: I have put original below in comments.
\subsection{Video Generation Models}  
Recent breakthroughs in T2V generation have been driven by large-scale diffusion and autoregressive architectures capable of synthesizing high-resolution, temporally coherent videos from prompts. 
State-of-the-art systems such as Sora-2~\citep{openai2025sora2} and Veo-3.1~\citep{deepmind2025veo3}, demonstrate impressive visual quality and scene diversity. However, despite the rapid advances, these models often struggle with temporal consistency and physical plausibility in generated videos.

To address these challenges, many efforts have focused on establishing comprehensive evaluation suites for video generation. \emph{Huang et al.} introduce VBench~\cite{huang2023VBench}, which decomposes video generation quality into multiple hierarchical dimensions containing subject identity consistency, motion smoothness, temporal flicker, and spatial relationships. VBench++~\cite{huang2024vbenchpp} extends this framework by incorporating additional dimensions such as compositionality, physical realism, and instruction following, offering broader coverage of both perceptual and reasoning-oriented aspects of video generation. Other works, such as WorldModelBench~\cite{li2025worldmodelbench}, VideoPhy~\cite{bansal2024videophy}, and Physion~\cite{bear2021physion}, further emphasize the need to measure dynamic coherence, physics adherence, and material interactions.

\subsection{VLM-as-a Judge for Video Model Evaluation}
The use of vision-language models (VLMs) and multimodal large language models as automated judges for video generation has become increasingly prevalent. \cite{he2024videoscore, he2025videoscore2}. Complementary to these judgers, frameworks like T2V-CompBench~\cite{sun2024t2vcompbench} propose a hybrid evaluation pipeline that integrates both VLM-based and computer vision-based metrics for compositional video understanding. Its VLM-as-a-Judge framework employs Grid-LLaVA to reason over sampled frames combined with detection-based and tracking-based modules (like GroundingDINO, SAM, Depth-Anything, and DOT) to create a structured rubric that aligns semantic, spatial and motion dimensions.

\subsection{Physical Commonsense Benchmarks for Video Models}  
Much of our inspiration for this work comes from past benchmarks designed to evaluate VLM's physical reasoning capabilities. Benchmarks like PhyGenBench~\cite{meng2025phygenbench}, VideoPhy~\cite{bansal2024videophy}, VideoPhy-2~\cite{bansal2025videophy2}, Physion~\cite{bear2021physion}, and IntPhys~\cite{riochet2018intphys}, all attempt to evaluate basic physical laws of video generation models, like mechanics, object interactions, and gravity. %\textbf{PHYRE}~\cite{bakhtin2019phyre}and \textbf{Physion++}~\cite{tung2023physionpp} push this idea further by requiring models to understand physics dynamics and use multi-step reasoning to solve puzzles.

However, there is currently a large gap in the space of video reasoning benchmarks that both require a synthesis of multiple concepts across multiple physics domains at a college undergraduate level and provide an auto-eval framework for evaluation. Our work, \sysnamenospace, fills that need by providing complex examples that require a fundamental understanding of the physical world. Our benchmark reveals that advancement is still needed for SOTA video models to truly understand physical phenomena.

\section{Conclusion}

%\paragraph{ \todo{(1) what problem we study. (2) benchmark overview. (3) auto eval. (4) human annotation and auto eval results. (5) quanlitative insights.} }

We introduce \sysnamenospace-Bench, the first benchmark designed to evaluate scientific reasoning in video models via the generation of scientific phenomena. Unlike prior benchmarks focused on physical commonsense, \sysnamenospace-Bench emphasizes understanding and reasoning across undergraduate-level topics and concepts in physics and chemistry. We ask expert annotators to rate video generations from the seven latest models along five dimensions. The results show that while the latest video models achieve high visual quality and temporal coherence, they still struggle with following instructions and adhering to physical laws. To further support automatic evaluation, we develop \sysnamenospace-Judge, a checklist-based VLM-as-a-Judge framework, and requires the VLM judge to reason over evidence extracted from key frames from the video and a CV-grounded evidence table. Correlation analysis between our auto-evaluation scores and human annotations
%, along with comparisons to other benchmarks, 
shows that \sysnamenospace-Judge achieves the strongest alignment with expert-rated rankings and best captures a video model’s scientific reasoning capability in comparison with existing benchmarks. 
%Looking ahead, challenges remain, particularly in automatically evaluating complex phenomena involving rapid, high-frequency physical or chemical changes, as well as in scaling to even more demanding scientific reasoning tasks. 
We hope \sysnamenospace-Bench will foster progress toward video models that not only produce visually compelling results, but also reason faithfully about the underlying scientific principles.

\newpage
{
    \small
    \bibliographystyle{ieeenat_fullname}
    \bibliography{sci-compass}
}

\newpage
\appendix
\onecolumn

\section{Test Suite Categorization}
\label{appendix:test_suite_categorization}

\subsection{Subcategory-to-Subject Mapping}

We define a set of physics and chemistry subcategories, each with a (two-letter code, subject) tuple:

\begin{verbatim}
SUBCATEGORY_TO_SUBJECT = {
    "Classical Mechanics": ("CM", "Physics"),
    "Optics": ("Op", "Physics"),
    "Thermodynamics": ("Th", "Physics"), 
    "Fluid Mechanics": ("FM", "Physics"),
    "Electromagnetism": ("El", "Physics"),
    "Wave": ("Wv", "Physics"),
    "Energy": ("En", "Physics"),
    "Material Mechanics": ("MM", "Physics"),
    "Modern Physics": ("MP", "Physics")

    
    "Redox Reactions": ("RR", "Chemistry"),
    "Liquid Chemistry": ("LC", "Chemistry"),
    "Acid-Base": ("AB", "Chemistry"),
    "Reaction Kinetics": ("RK", "Chemistry"),
    "Material Chemistry": ("MC", "Chemistry")
}
\end{verbatim}

\subsection{Concept-to-Subcategory Mapping}

Each fine-grained concept is mapped to \emph{exactly one} subcategory. Here is the complete mapping used: 
\begin{verbatim}
CONCEPTS_TO_SUBCATEGORY = {
    "Dispersion": "Optics",
    "Reflection": "Optics",
    "Refraction": "Optics",
    "Diffraction": "Optics",
    "Interference": "Optics",
    "Polarization": "Optics",
    "Total Internal Reflection": "Optics",
    "Light Spectrum": "Optics",
    "Light": "Optics",
    

    "Gravity": "Classical Mechanics",
    "Inertia": "Classical Mechanics",
    "Friction": "Classical Mechanics",
    "Momentum Conservation": "Classical Mechanics",
    "Projectile Motion": "Classical Mechanics",
    "Elastic Collision": "Classical Mechanics",
    "Inelastic Collision": "Classical Mechanics",
    "Pendulum": "Classical Mechanics",
    "Simple harmonic motion": "Classical Mechanics",
    "Hooke's law": "Classical Mechanics",
    "Rotational Dynamics": "Classical Mechanics",
    "Angular Momentum": "Classical Mechanics",
    "Gyroscopic Precession": "Classical Mechanics",
    "Newton's Laws": "Classical Mechanics",
    "Kinematics": "Classical Mechanics",
    "Impulse": "Classical Mechanics",
    "Impact Mechanics": "Classical Mechanics",
    "Relative Velocity": "Classical Mechanics",
    "Kinetic Energy Loss": "Classical Mechanics",
    "Centripetal Force": "Classical Mechanics",
    "Centrifugal Force": "Classical Mechanics",
    
  
    "Buoyancy": "Fluid Mechanics",
    "Bernoulli Effect": "Fluid Mechanics",
    "Coanda Effect": "Fluid Mechanics",
    "Capillary Action": "Fluid Mechanics",
    "Fluid Dynamics": "Fluid Mechanics",
    "Fluid Flow": "Fluid Mechanics",
    "Hydrostatic": "Fluid Mechanics",
    "Non-Newtonian Fluid": "Fluid Mechanics",
    "Density and Buoyancy": "Fluid Mechanics",
    "Air pocket": "Fluid Mechanics",
    

    "Heat": "Thermodynamics",
    "Heat Transfer": "Thermodynamics",
    "Thermal conduction": "Thermodynamics",
    "Thermal convection": "Thermodynamics",
    "Heat Capacity": "Thermodynamics",
    "Thermodynamics": "Thermodynamics",
    "Statistical equilibrium": "Thermodynamics",
    "Pressure": "Thermodynamics",
    "pressure equilibrium": "Thermodynamics",
    "Gas Laws": "Thermodynamics",
    "Vacuum": "Thermodynamics",
    "Phase transitions": "Thermodynamics",
    "Phase Change": "Thermodynamics",
    "Condensation": "Thermodynamics",
    

    "Electrostatics": "Electromagnetism",
    "Magnetics": "Electromagnetism",
    "Magnetism": "Electromagnetism",
    "Electromagnetism": "Electromagnetism",
    "Simple circuit": "Electromagnetism",
    "Simple circuit lighting": "Electromagnetism",
    "Eddy Currents": "Electromagnetism",
    "Static Electricity": "Electromagnetism",
    "Gas Ionization": "Electromagnetism",
    

    "Sound": "Wave",
    "Sound Waves": "Wave",
    "Acoustics": "Wave",
    "Standing waves": "Wave",
    "Vibration": "Wave",
    "Resonance": "Wave",
    "Wave Propagration": "Wave",
    

    "Energy Conversion": "Energy",
    "Energy Transfer": "Energy",
    "Energy Storage & Release": "Energy",
    

    "Structural Physics": "Material Mechanics",
    "stress‐concentration": "Material Mechanics",
    "Fracture": "Material Mechanics",
    "Deformation": "Material Mechanics",
    "Stability": "Material Mechanics",
    "Equlibrium": "Material Mechanics",
    

    "Quantum Mechanics": "Modern Physics",
    "Plasma Physics": "Modern Physics",
    

    "Combustion": "Redox Reactions",
    "flammability": "Redox Reactions",
    "Redox Reaction": "Redox Reactions",
    "Galvanic Cell": "Redox Reactions",
    "Electrolysis": "Redox Reactions",
    "Electrochemistry": "Redox Reactions",
    

    "Acid-Base Reaction": "Acid-Base",
    "Indicator color change": "Acid-Base",
    

    "Solubility": "Liquid Chemistry",
    "Emulsion": "Liquid Chemistry",
    "Surface Tension": "Liquid Chemistry",
    "Surfactants": "Liquid Chemistry",
    "Viscosity": "Liquid Chemistry",
    "Density": "Liquid Chemistry",
    "Tyndall Effect": "Liquid Chemistry",
    

    "Chemical Reaction": "Reaction Kinetics",
    "Catalysis": "Reaction Kinetics",
    "Exothermic Reaction": "Reaction Kinetics",
    "Precipitation reaction": "Reaction Kinetics",


    "Polymers": "Material Chemistry",
    "Nano Science": "Material Chemistry",
    "Fluorescence": "Material Chemistry",
    "Phosphorescence": "Material Chemistry",
    "Protein Denaturation": "Material Chemistry",
}
\end{verbatim}

% TEST SUITE
\section{Test Suite Examples}
\label{appendix:test_suite_examples}

\subsection{T2V Examples}
\begin{figure*}[htbp]
\centering
\setlength{\tabcolsep}{4pt}
\renewcommand{\arraystretch}{1.2}
\begin{tabular}{@{}p{0.15\textwidth}p{0.82\textwidth}@{}}
\toprule
\multicolumn{2}{c}{\textbf{Curved Refraction Gradient}}\\
\midrule
\textbf{Prompt} & 
A transparent water tank is filled with sugar water of layered concentrations, placed in a dark room where a laser beam enters from the side to make the light path visible.\\[0.3em]

\textbf{Expected} & 
The different densities of the two layers create distinct refractive indices. As the beam passes between them, it bends sharply (Snell’s law). Because the boundary is diffuse, the beam curves smoothly through the gradient, illustrating how refractive index changes with density.\\
\midrule

\multicolumn{2}{c}{\cellcolor{green!10}\textbf{Sora-2:} {\color{green!60!black} Successfully Generated Expected Phenomenon}}\\[0.2em]

\multicolumn{2}{c}{
  \includegraphics[height=1.3cm]{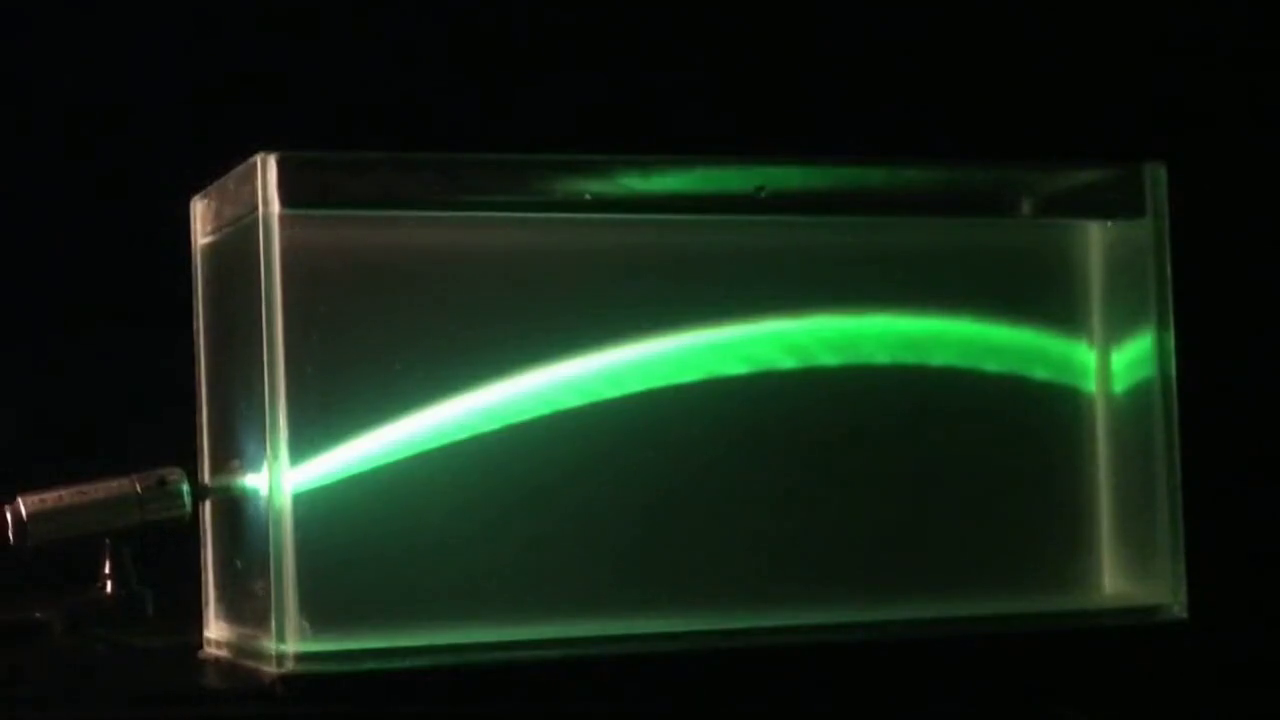}\hspace{1pt}%
  \includegraphics[height=1.3cm]{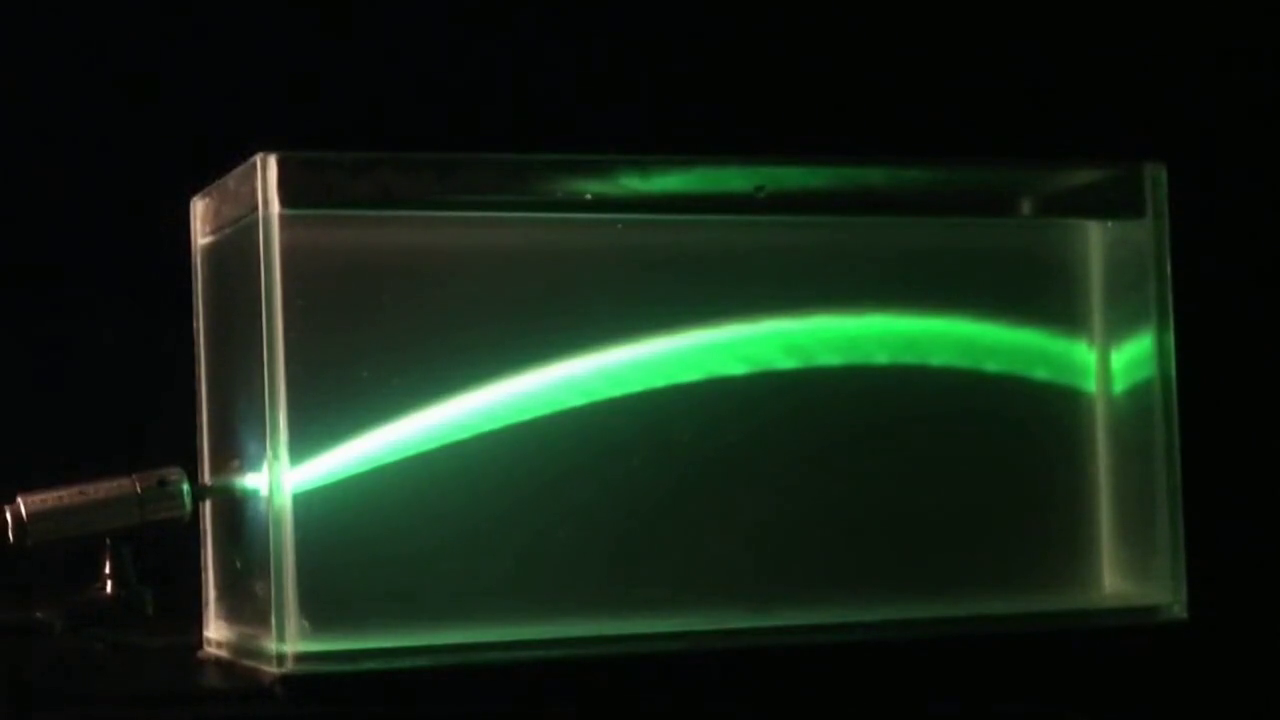}\hspace{1pt}%
  \includegraphics[height=1.3cm]{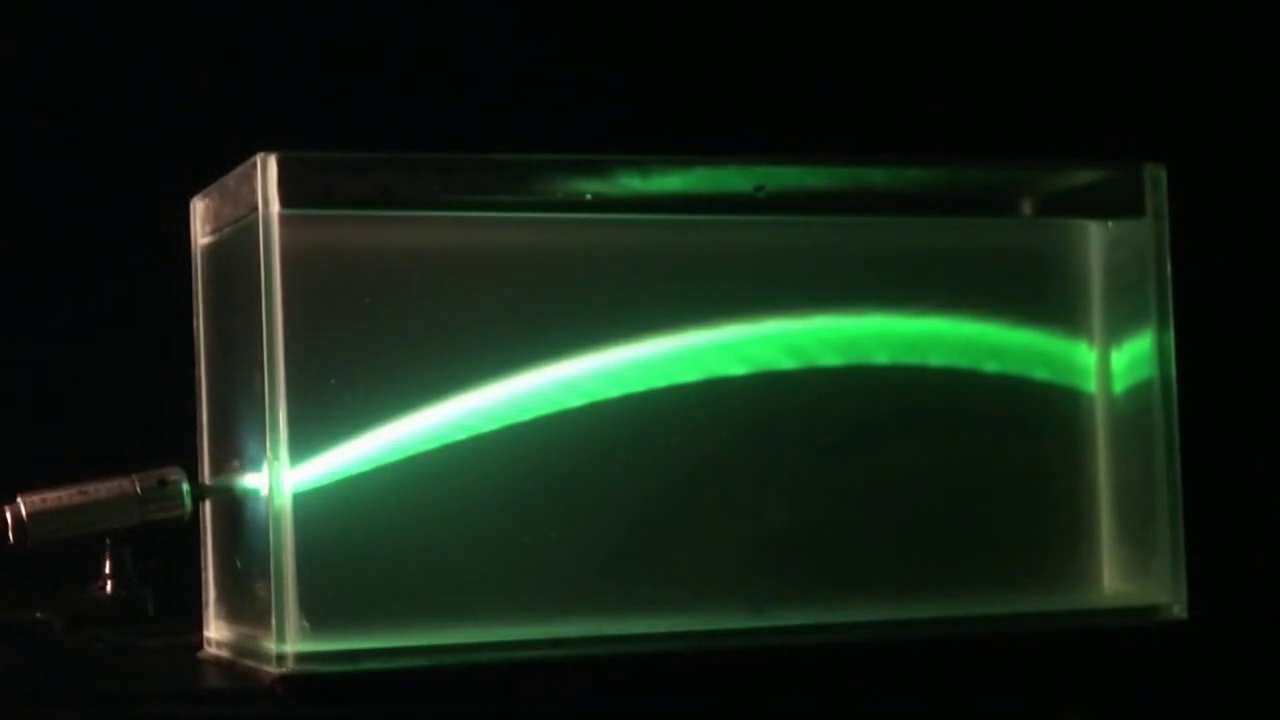}\hspace{1pt}%
  \includegraphics[height=1.3cm]{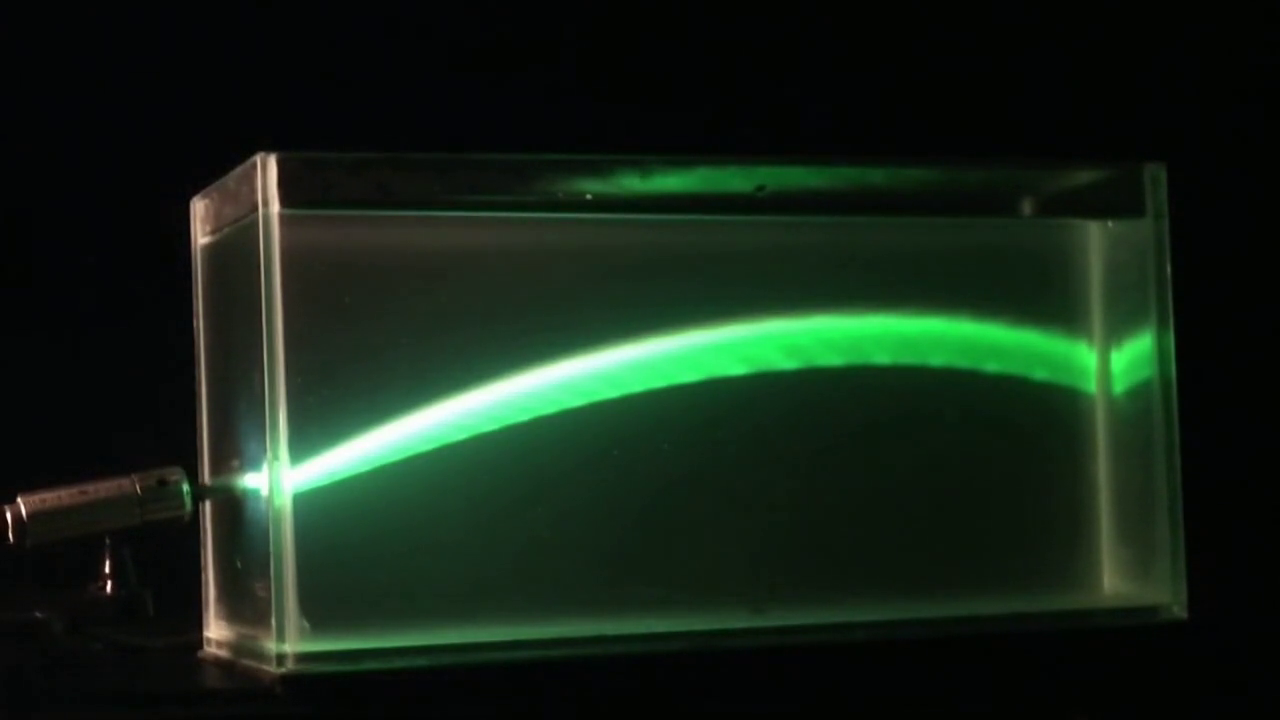}\hspace{1pt}%
  \includegraphics[height=1.3cm]{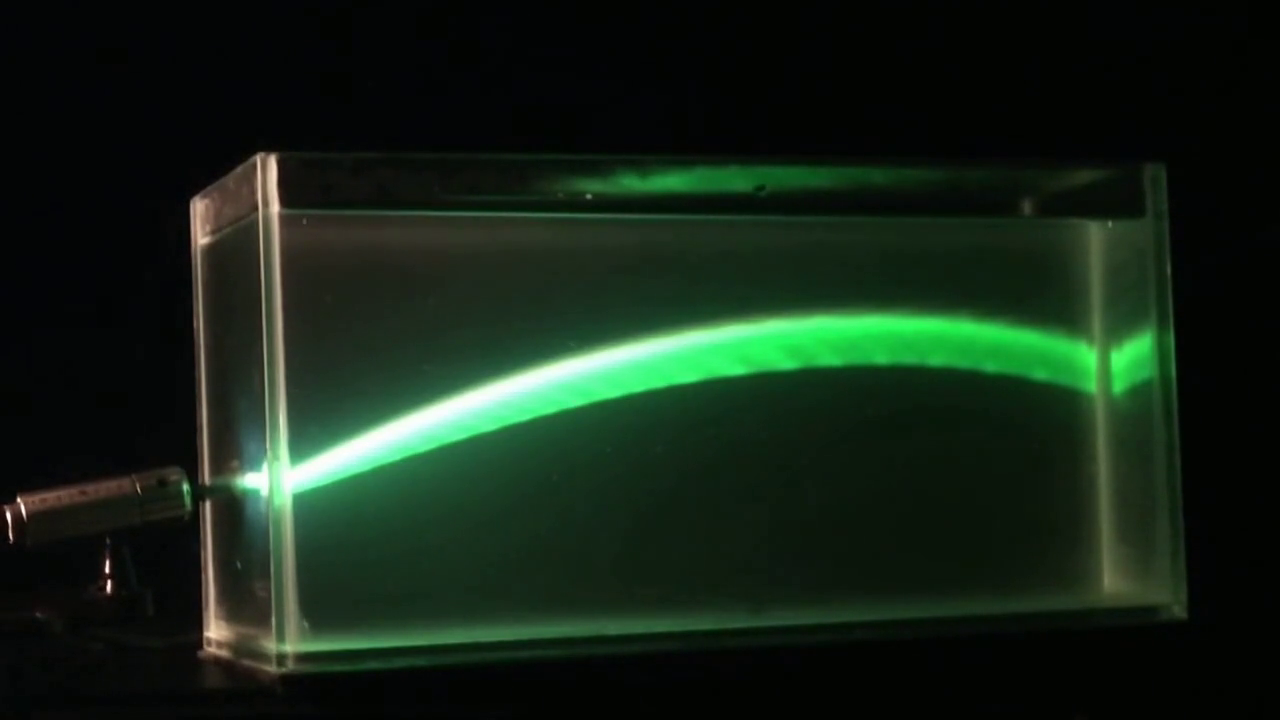}\hspace{1pt}%
  \includegraphics[height=1.3cm]{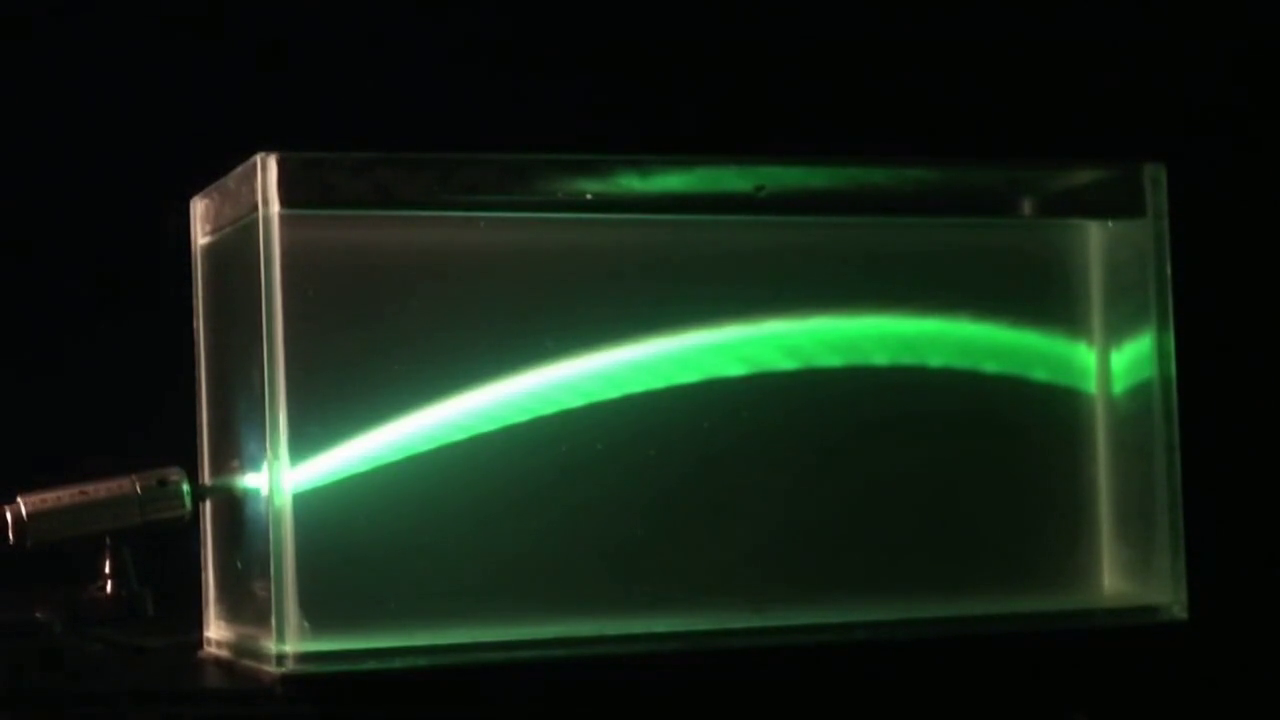}

}\\[0.2em]

\multicolumn{2}{c}{
  \textbf{PCS:} 4/4 \quad 
  \textbf{PCG:} 3/4 \quad 
  \textbf{CDN:} 4/4 \quad 
  \textbf{IMB:} 4/4 \quad 
  \textbf{STC:} 3/4
}\\
\bottomrule
\end{tabular}
\caption{Sora-2 generated video of Curved Refraction Gradient, where Sora-2 correctly depicts the setup and the expected phenomenon, despite the visual output being largely static. Human Annotation Rating: Prompt Consistency (PCS), Phenomenon Congruency (PCG), Correct Dynamism (CDN), Immutability (IMB), and Spatio-Temporal Coherence (STC).}
\label{fig:example175_t2v}
\end{figure*}

\begin{figure*}[htbp]
\centering
\setlength{\tabcolsep}{4pt}
\renewcommand{\arraystretch}{1.2}
\begin{tabular}{@{}p{0.15\textwidth}p{0.82\textwidth}@{}}
\toprule
\multicolumn{2}{c}{\textbf{Water Separation}}\\
\midrule
\textbf{Prompt} & 
Add red food coloring to a glass of hot water and blue food coloring to a glass of cold water. Cover the hot water glass with a thin plastic sheet. Carefully invert the glass of hot water, gently position it mouth-to-mouth on top of the glass of cold water and slowly remove the plastic sheet.\\[0.3em]

\textbf{Expected} & 
The red (hot) water remains above the blue (cold) water with minimal mixing, forming a stable boundary between the two layers.
\\
\midrule

\multicolumn{2}{c}{\cellcolor{red!10}\textbf{Sora-2:} {\color{red!60!black} Failed to Generated Expected Phenomenon}}\\[0.2em]

\multicolumn{2}{c}{
  \includegraphics[height=1.3cm]{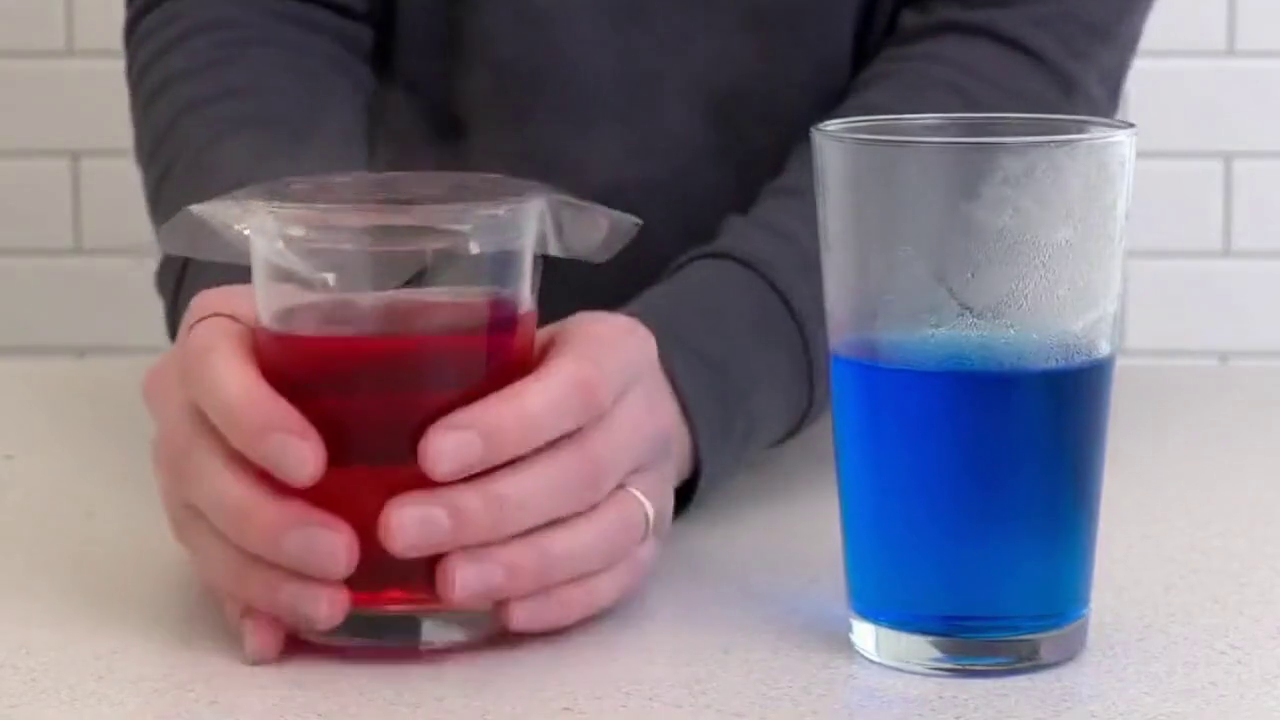}\hspace{1pt}%
  \includegraphics[height=1.3cm]{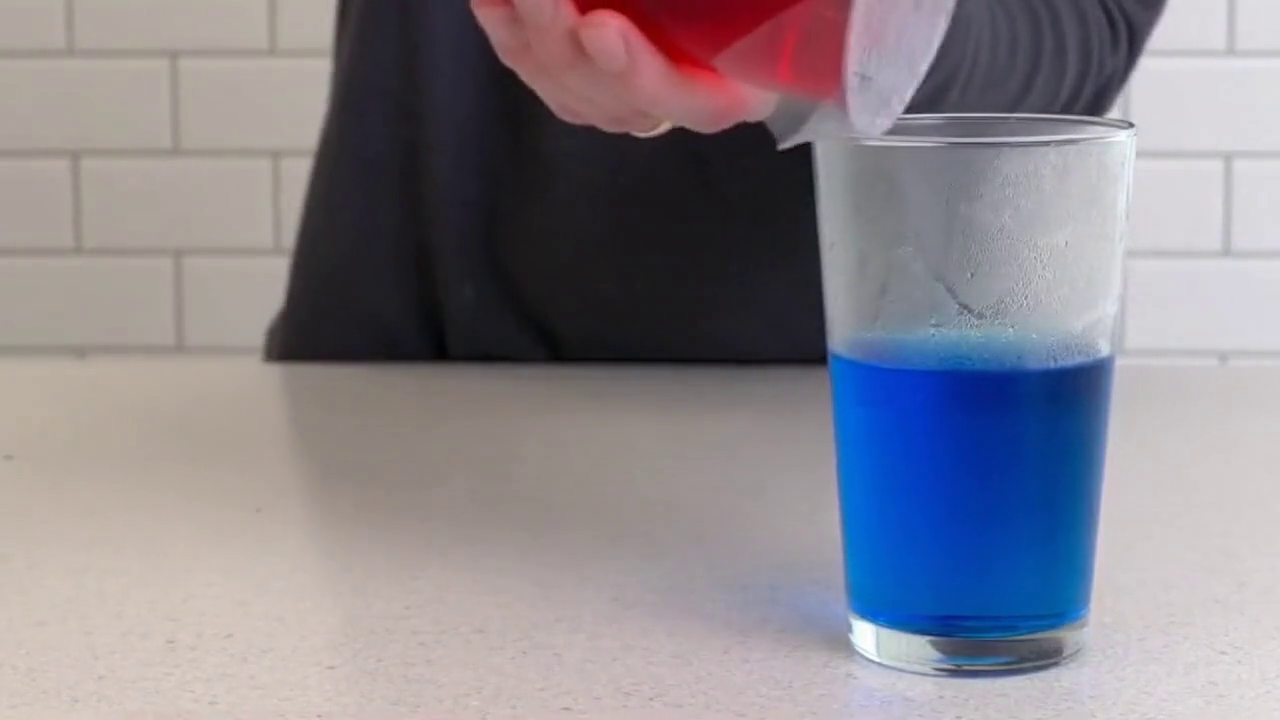}\hspace{1pt}%
  \includegraphics[height=1.3cm]{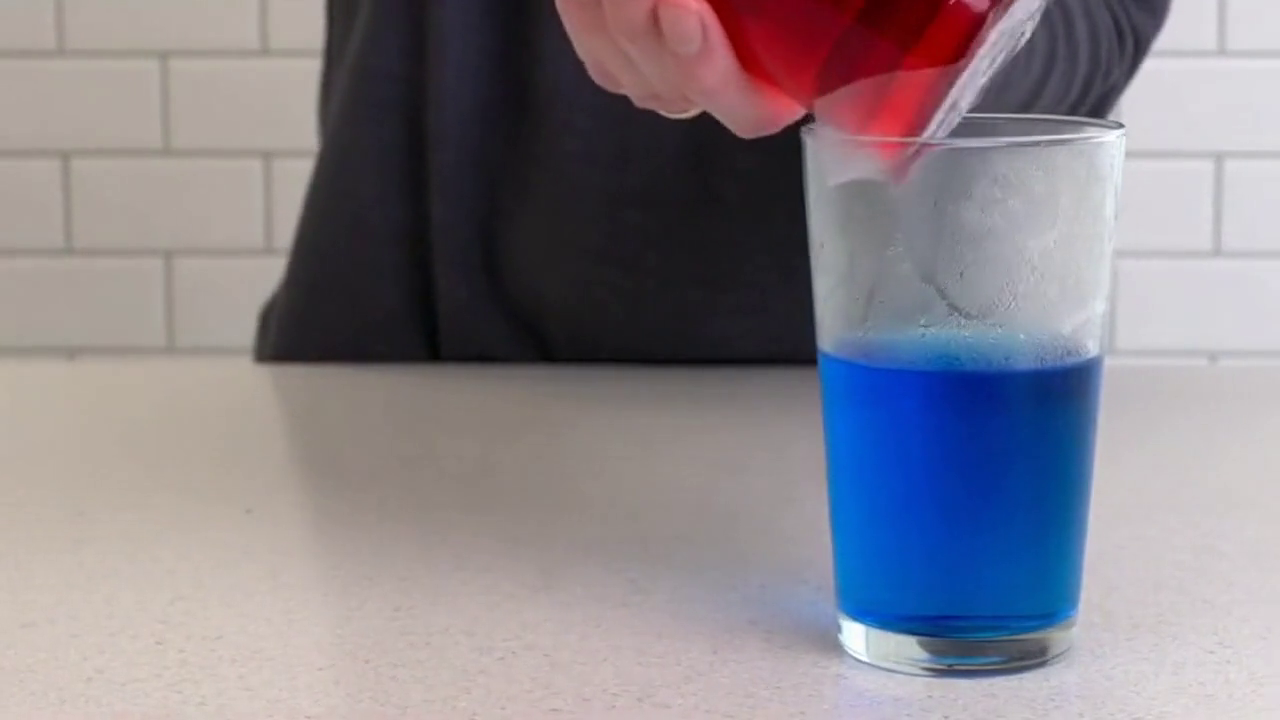}\hspace{1pt}%
  \includegraphics[height=1.3cm]{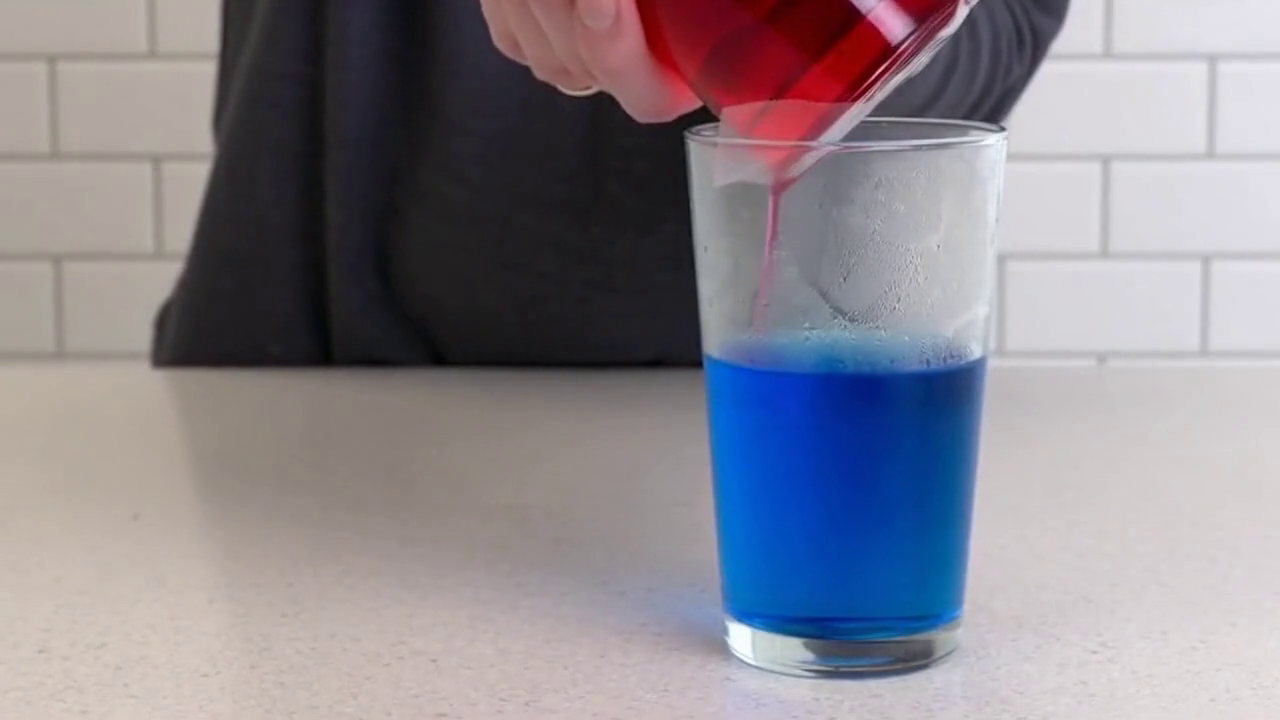}\hspace{1pt}%
  \includegraphics[height=1.3cm]{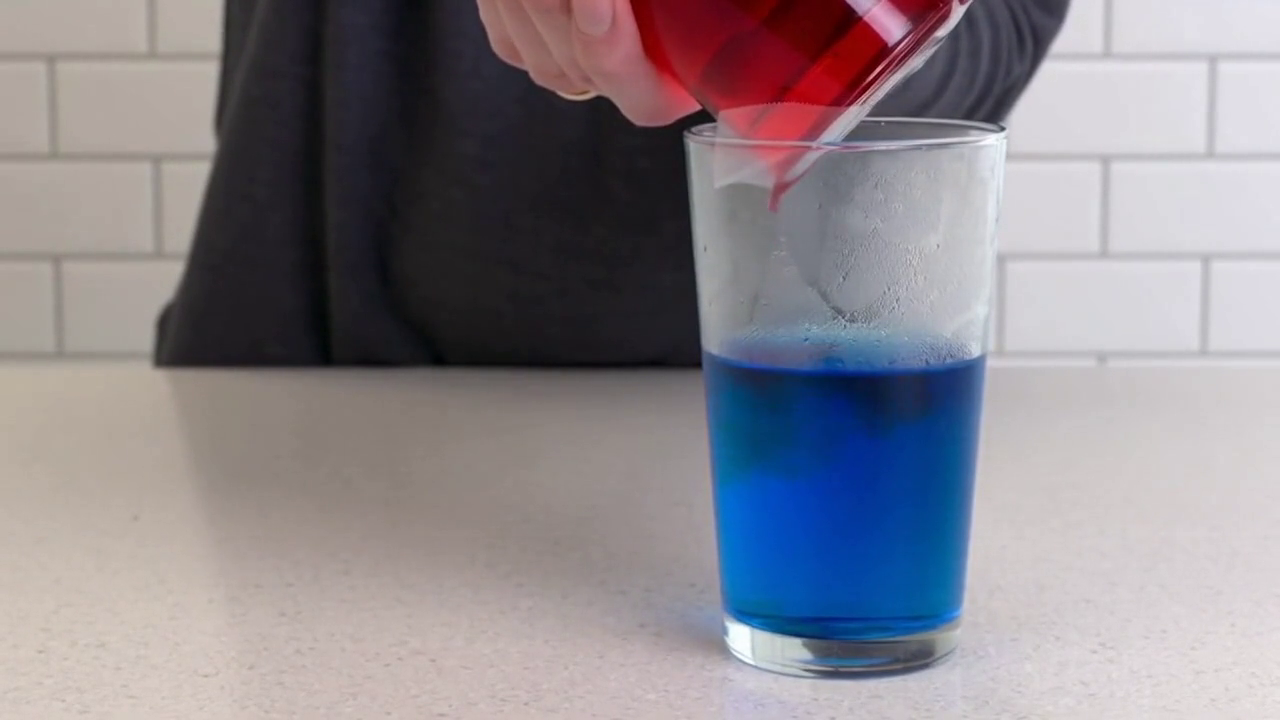}\hspace{1pt}%
  \includegraphics[height=1.3cm]{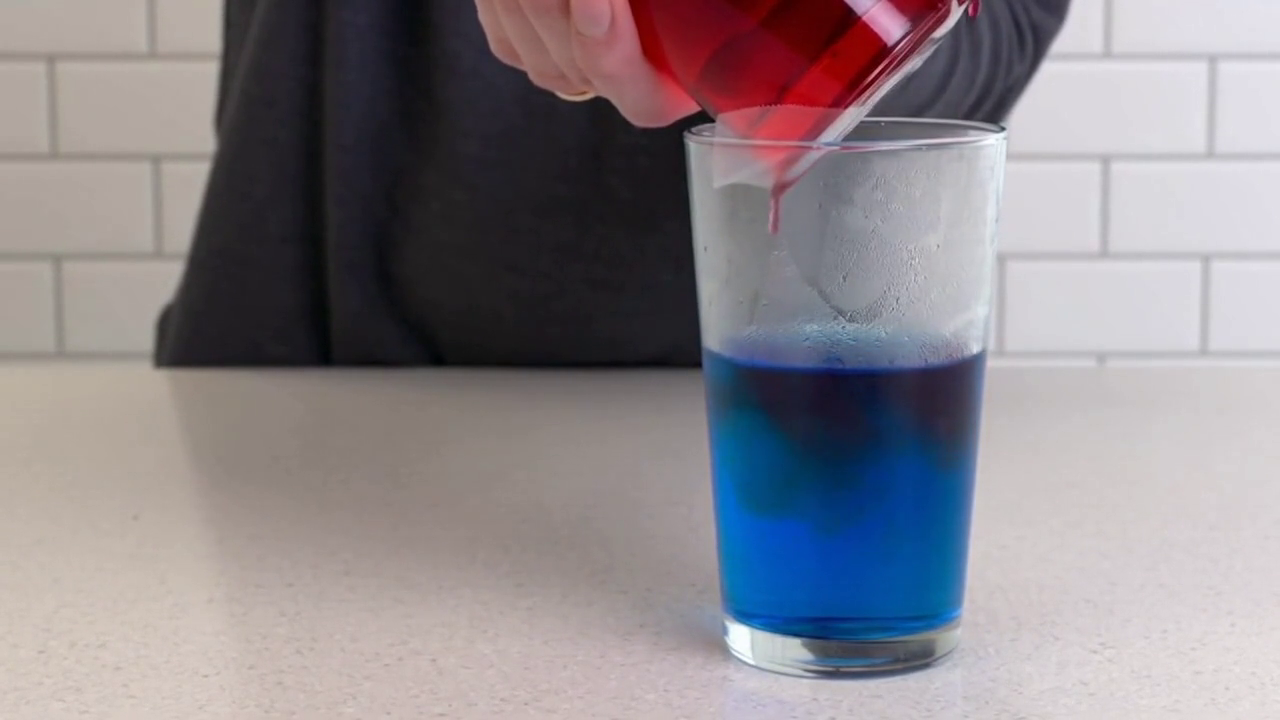}

}\\[0.2em]

\multicolumn{2}{c}{
  \textbf{PCS:} 3/4 \quad 
  \textbf{PCG:} 2/4 \quad 
  \textbf{CDN:} 3/4 \quad 
  \textbf{IMB:} 4/4 \quad 
  \textbf{STC:} 3/4
}\\
\bottomrule
\end{tabular}
\caption{Sora-2 generated video of Water Separation, where Sora-2 fails to depict the expected phenomenon (the two colored liquids should barely mix). Human Annotation Rating: Prompt Consistency (PCS), Phenomenon Congruency (PCG), Correct Dynamism (CDN), Immutability (IMB), and Spatio-Temporal Coherence (STC).}
\label{fig:example_water_separation_t2v}
\end{figure*}

\begin{figure*}[htbp]
\centering
\setlength{\tabcolsep}{4pt}
\renewcommand{\arraystretch}{1.2}
\begin{tabular}{@{}p{0.15\textwidth}p{0.82\textwidth}@{}}
\toprule
\multicolumn{2}{c}{\textbf{Dry Ice Bubbles}}\\
\midrule
\textbf{Prompt} & 
Place a small chunk of dry ice into a tall glass filled with clear warm soapy water.\\[0.3em]

\textbf{Expected} & 
A large amount of white, misty bubbles rapidly forms and overflows the rim of the glass.
\\
\midrule

\multicolumn{2}{c}{\cellcolor{red!10}\textbf{Ray-2:} {\color{red!60!black} Failed to Generated Expected Phenomenon}}\\[0.2em]

\multicolumn{2}{c}{
  \includegraphics[height=1.3cm]{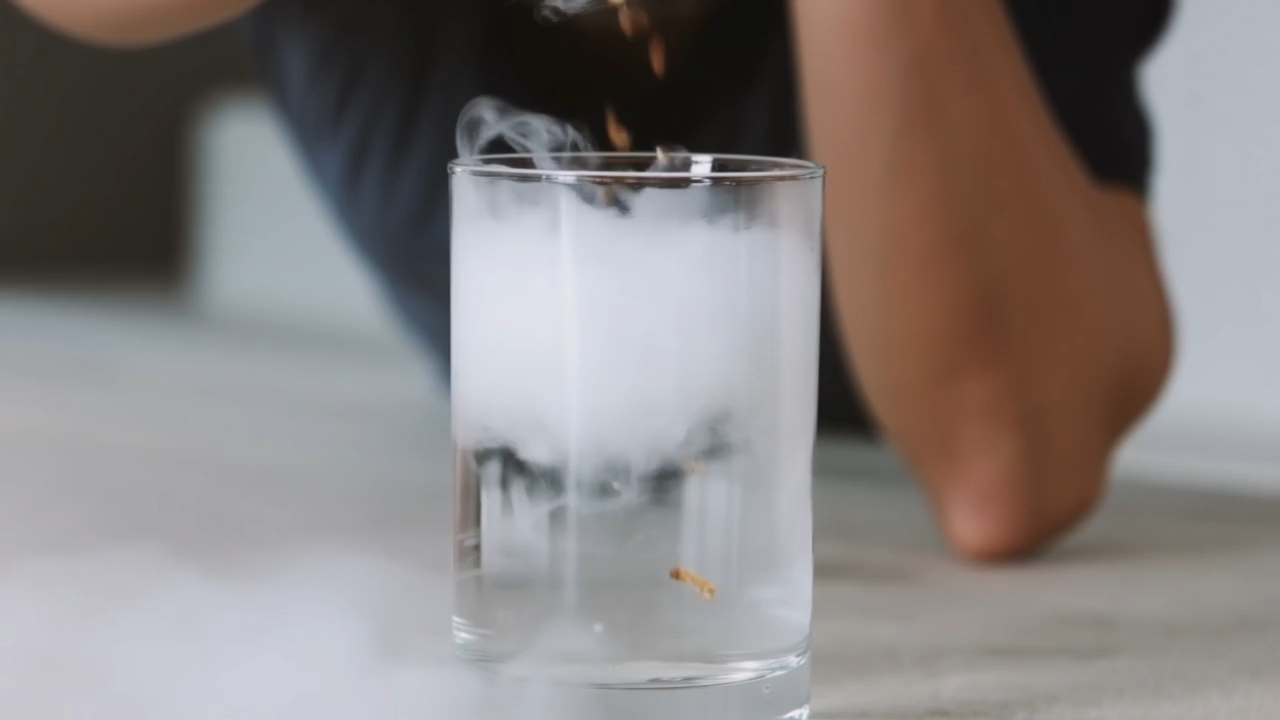}\hspace{1pt}%
  \includegraphics[height=1.3cm]{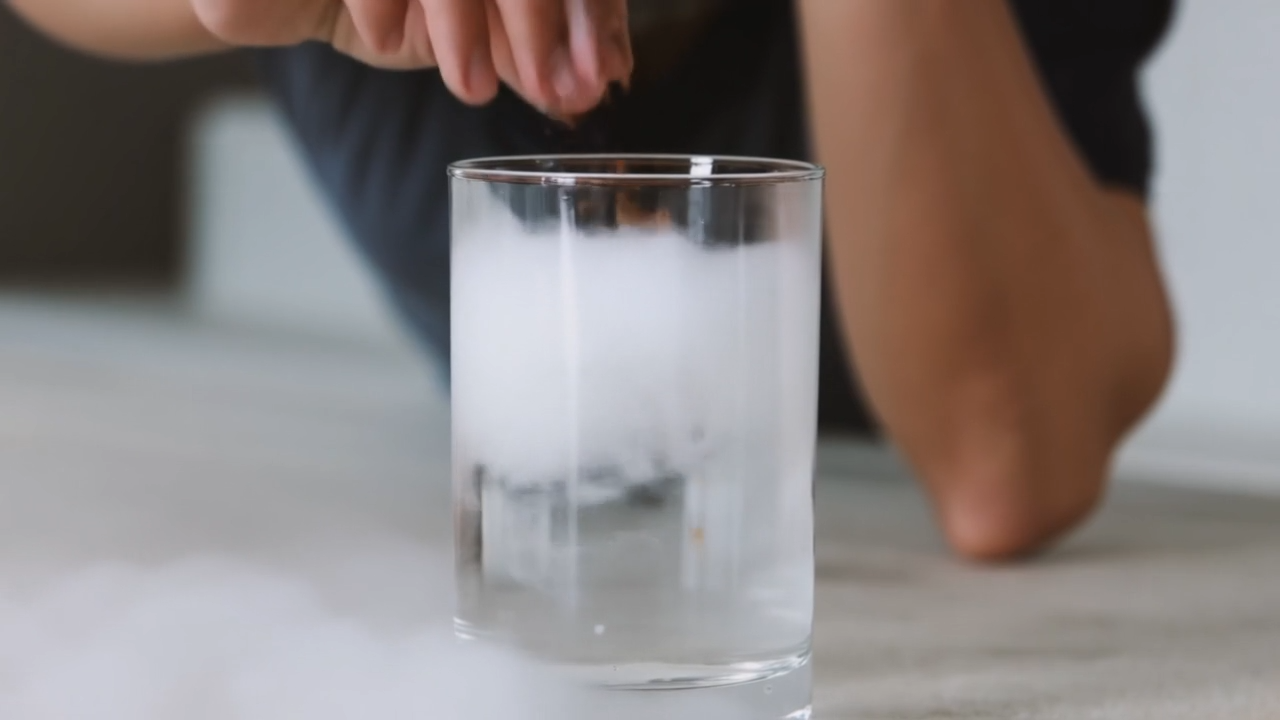}\hspace{1pt}%
  \includegraphics[height=1.3cm]{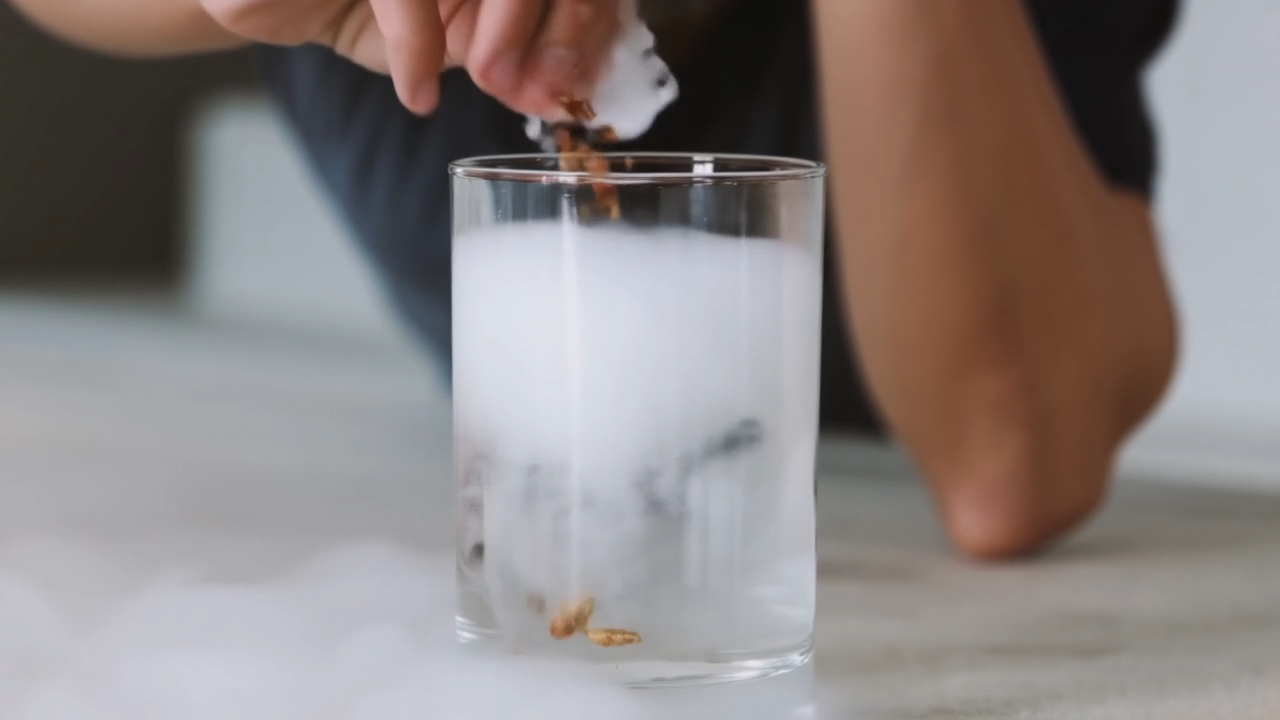}\hspace{1pt}%
  \includegraphics[height=1.3cm]{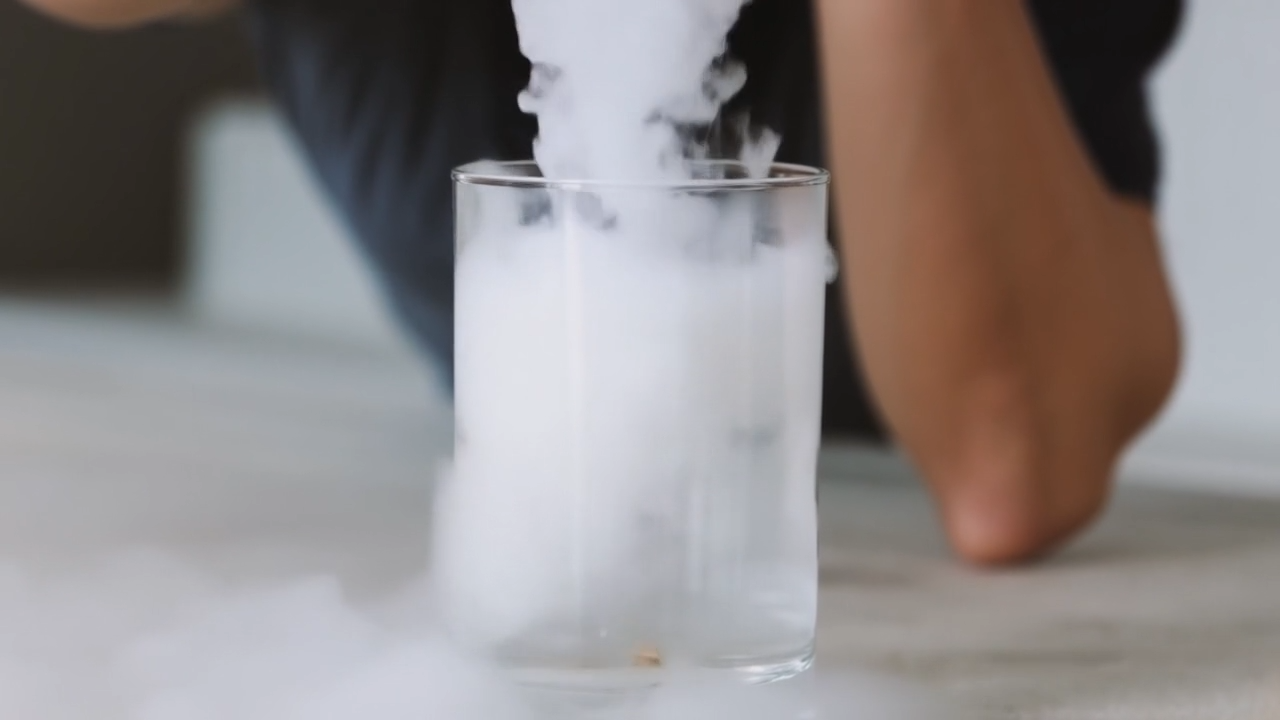}\hspace{1pt}%
  \includegraphics[height=1.3cm]{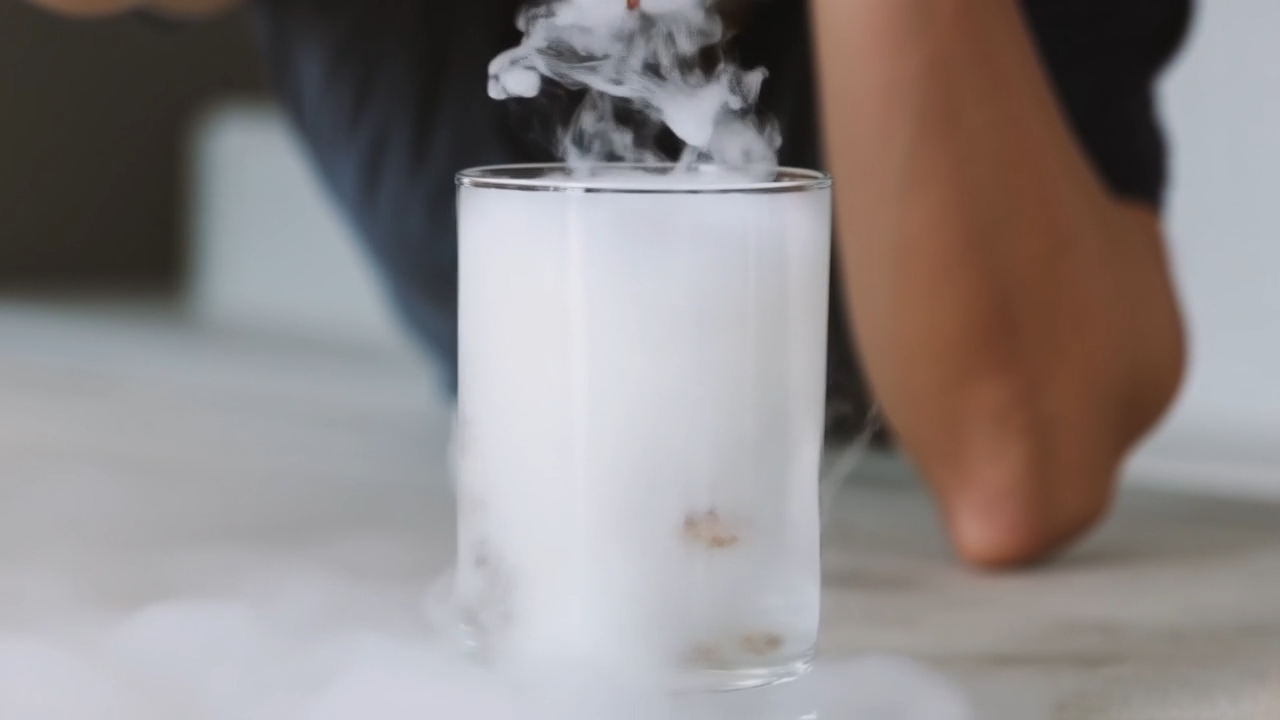}\hspace{1pt}%
  \includegraphics[height=1.3cm]{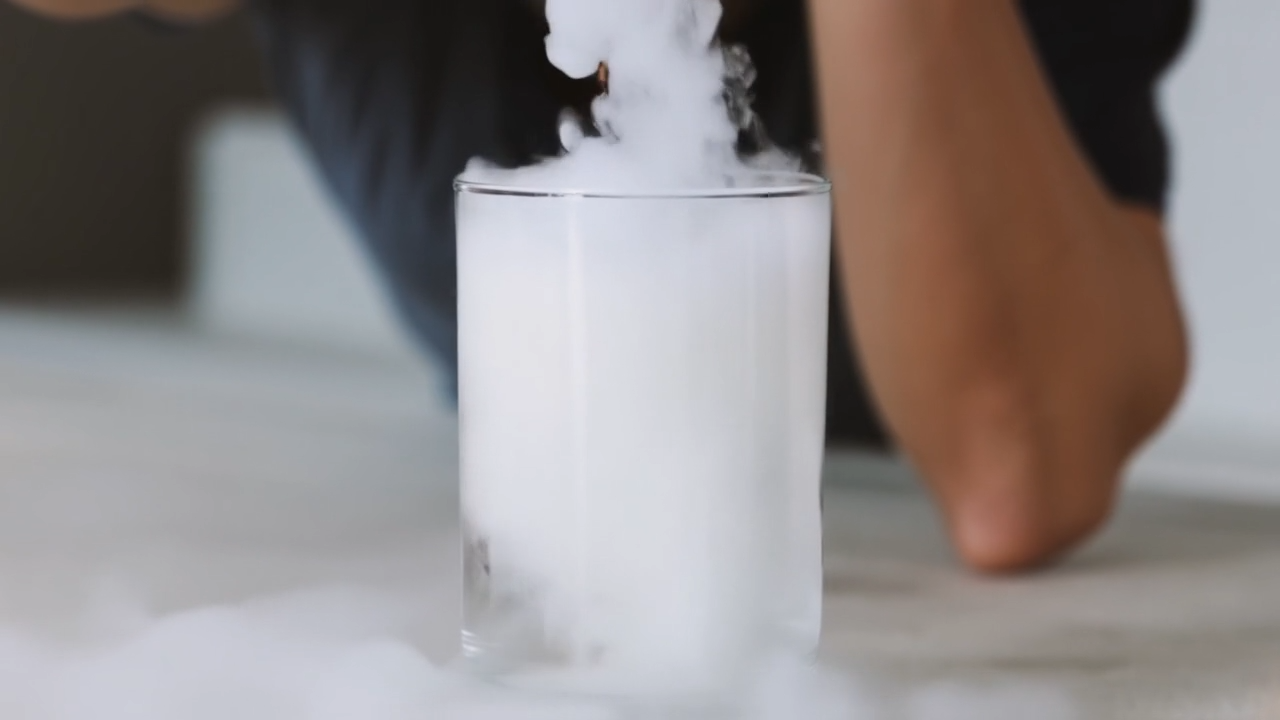}\hspace{1pt}%
  \includegraphics[height=1.3cm]{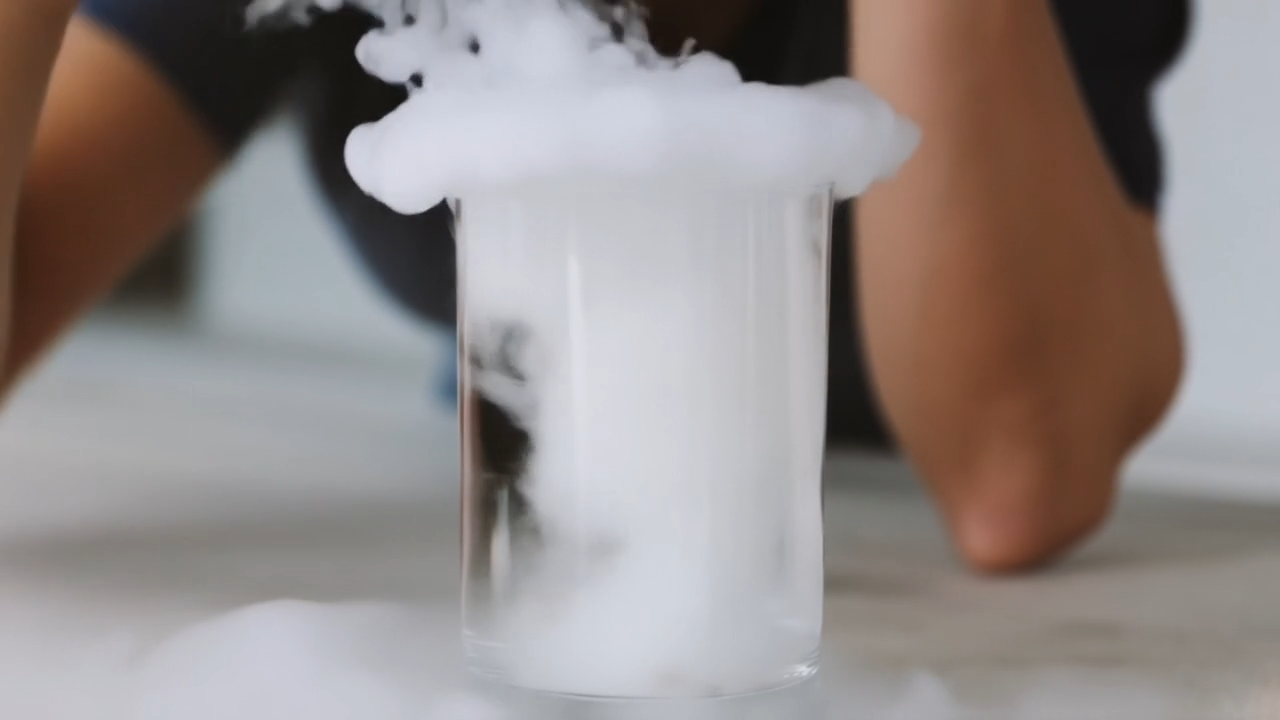}\hspace{1pt}%
  \includegraphics[height=1.3cm]{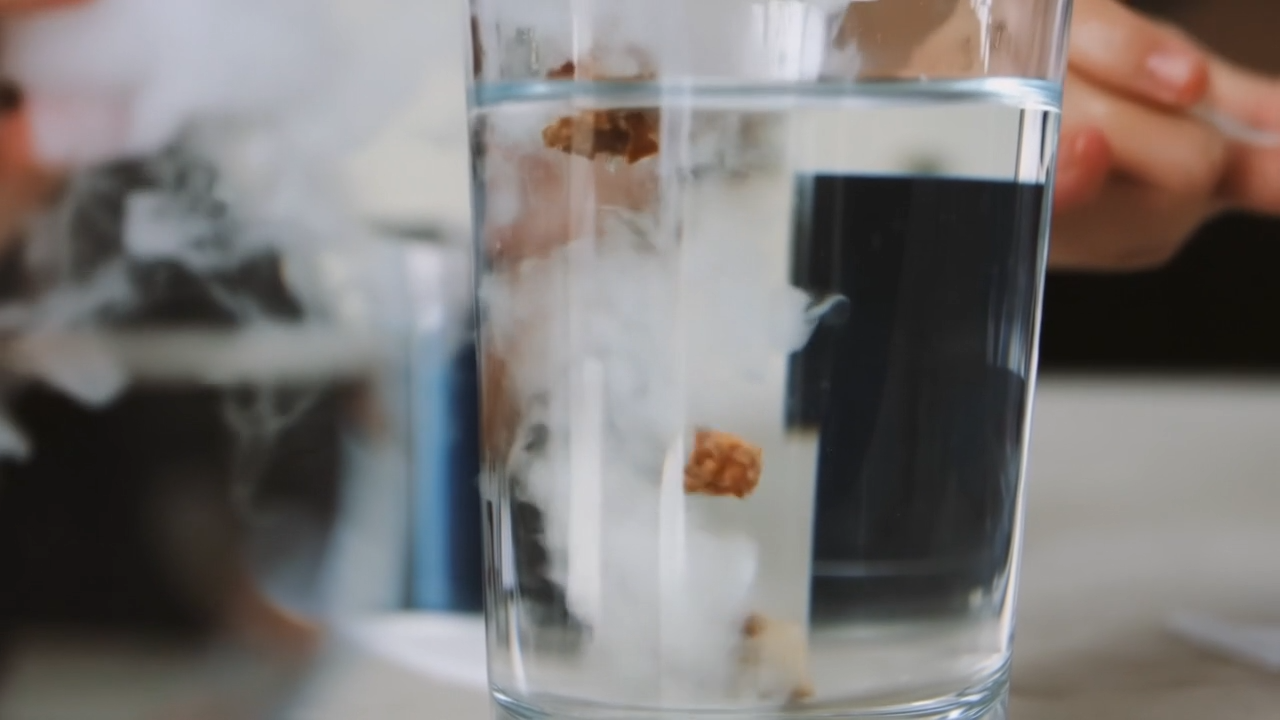}

}\\[0.2em]

\multicolumn{2}{c}{
  \textbf{PCS:} 1/4 \quad 
  \textbf{PCG:} 1/4 \quad 
  \textbf{CDN:} 4/4 \quad 
  \textbf{IMB:} 4/4 \quad 
  \textbf{STC:} 3/4
}\\
\bottomrule
\end{tabular}
\caption{Ray-2 generated video of water, which is not soapy, and instead of dry ice, some other substance is dropped into the water. Human Annotation Rating: Prompt Consistency (PCS), Phenomenon Congruency (PCG), Correct Dynamism (CDN), Immutability (IMB), and Spatio-Temporal Coherence (STC).}
\label{fig:example122_t2v_ray2}
\end{figure*}

\begin{figure*}[htbp]
\centering
\setlength{\tabcolsep}{4pt}
\renewcommand{\arraystretch}{1.2}
\begin{tabular}{@{}p{0.15\textwidth}p{0.82\textwidth}@{}}
\toprule
\multicolumn{2}{c}{\textbf{Milky Limewater}}\\
\midrule
\textbf{Prompt} & 
Two clear cups are shown side by side. The left cup contains fizzing vinegar and baking soda; the right cup contains clear limewater. A straw connects the two cups, guiding the gas from the left cup into the limewater in the right cup.\\[0.3em]

\textbf{Expected} & 
The limewater quickly turns milky white, and bubbles rise from the bottom.﻿
\\
\midrule

\multicolumn{2}{c}{\cellcolor{red!10}\textbf{Wan-2.5:} {\color{red!60!black} Failed to Generated Expected Phenomenon}}\\[0.2em]

\multicolumn{2}{c}{
  \includegraphics[height=1.3cm]{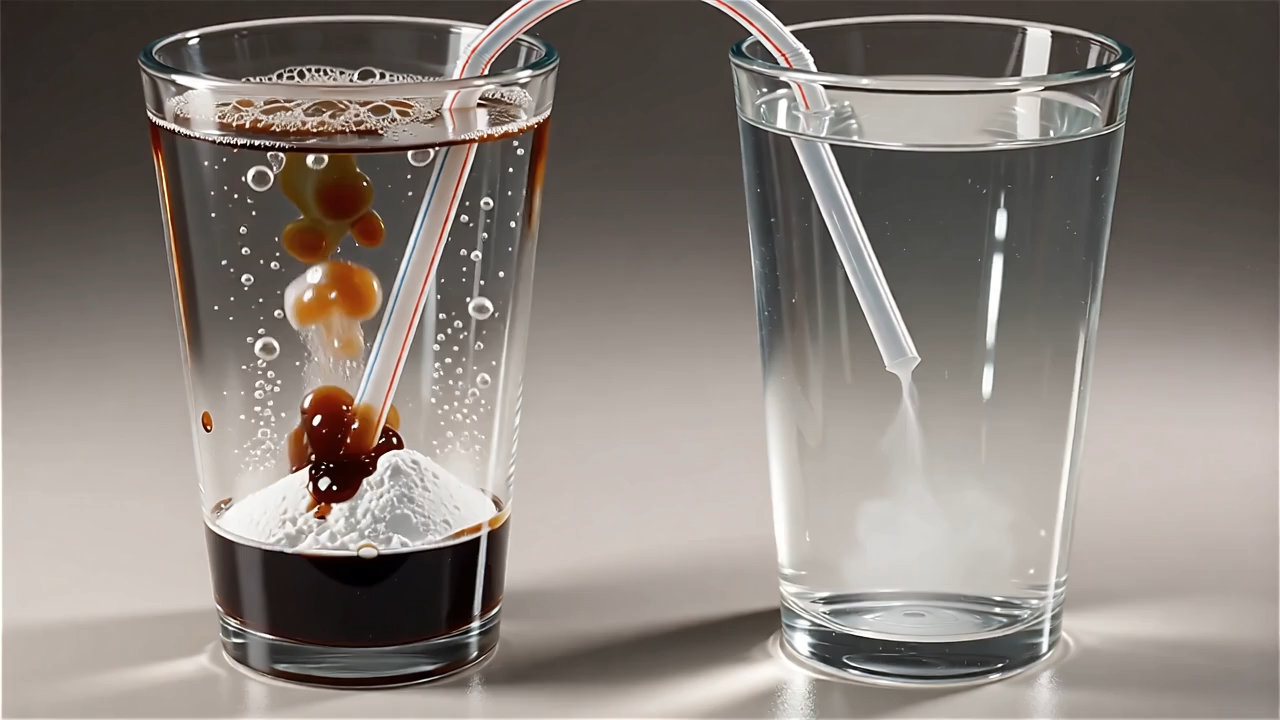}\hspace{1pt}%
  \includegraphics[height=1.3cm]{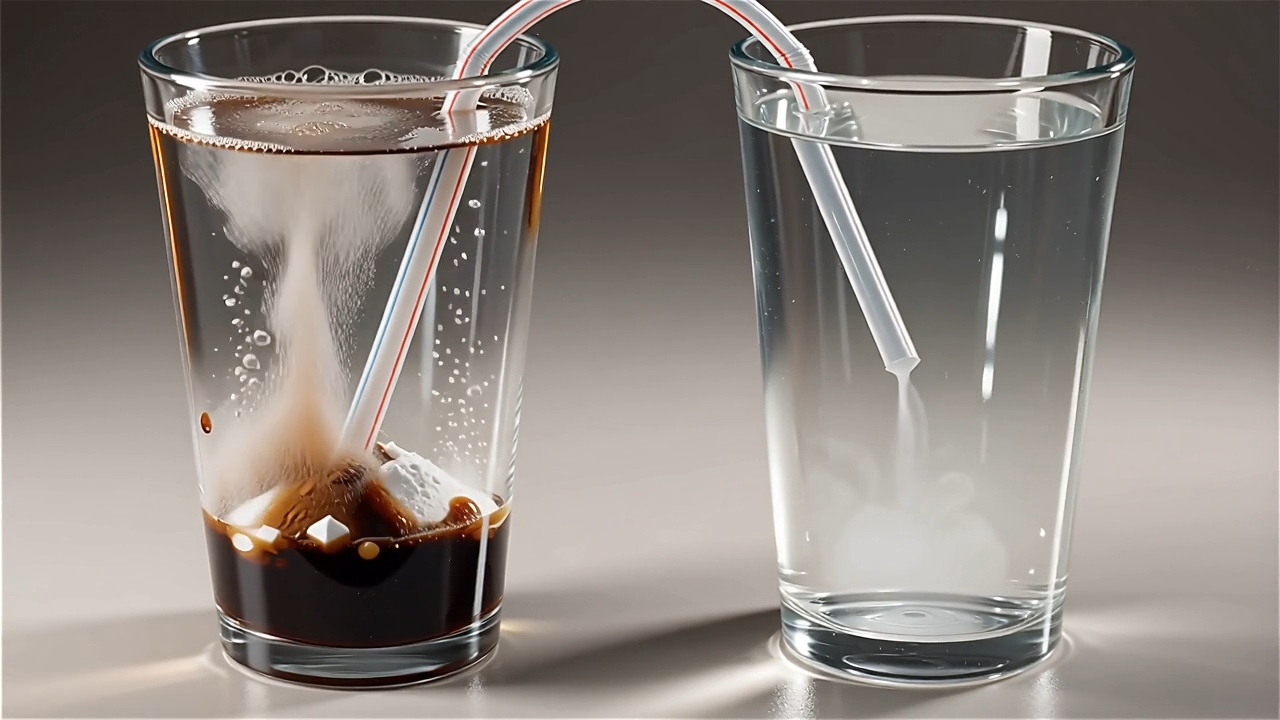}\hspace{1pt}%
  \includegraphics[height=1.3cm]{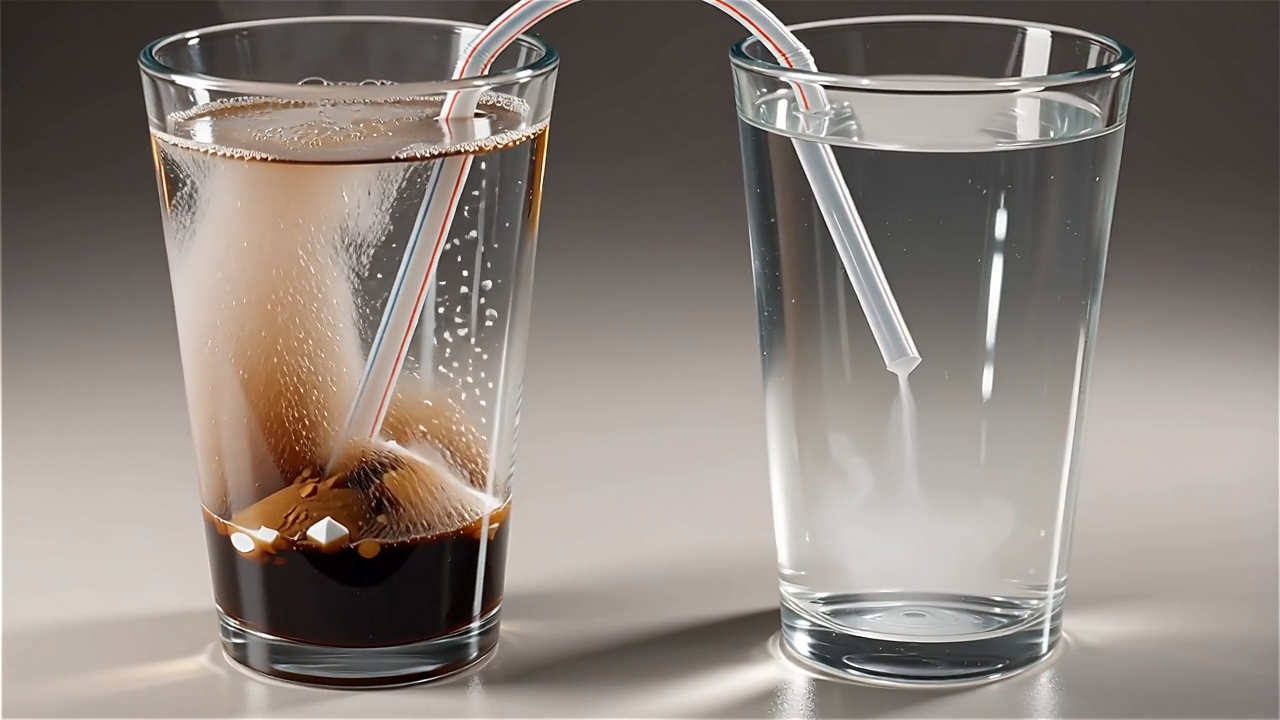}\hspace{1pt}%
  \includegraphics[height=1.3cm]{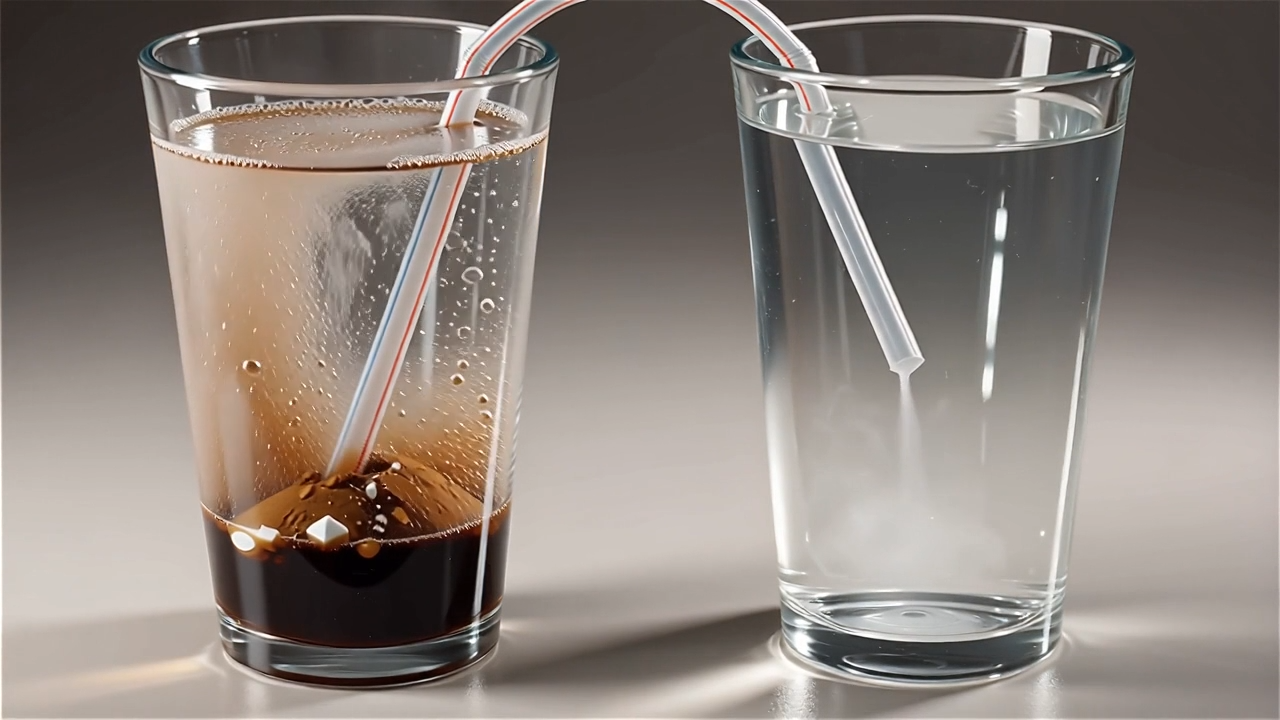}\hspace{1pt}%
  \includegraphics[height=1.3cm]{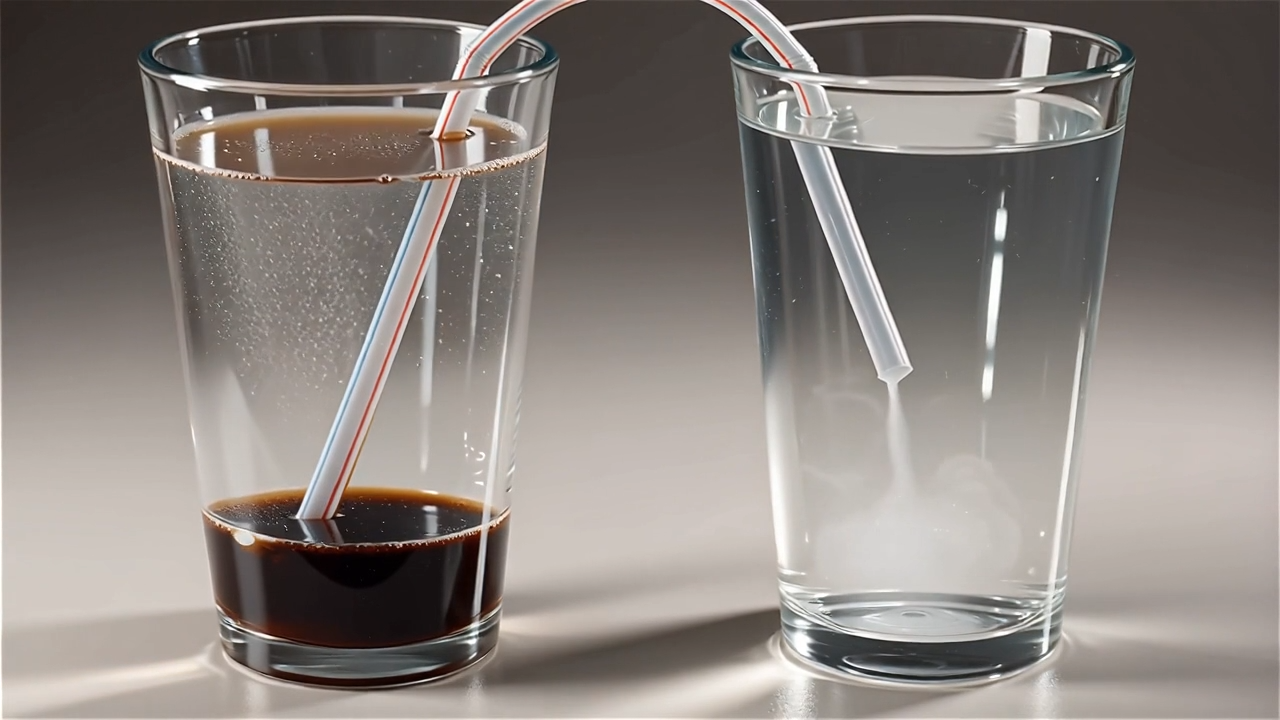}\hspace{1pt}%
  \includegraphics[height=1.3cm]{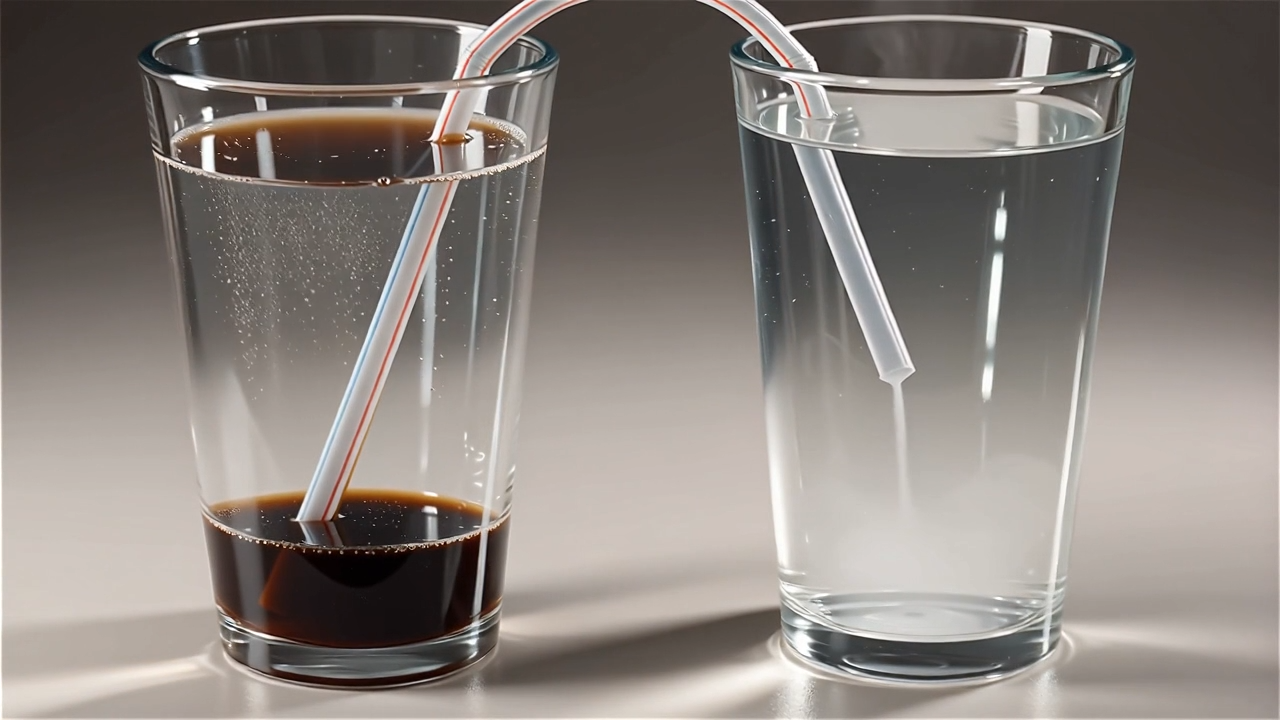}\hspace{1pt}%
  \includegraphics[height=1.3cm]{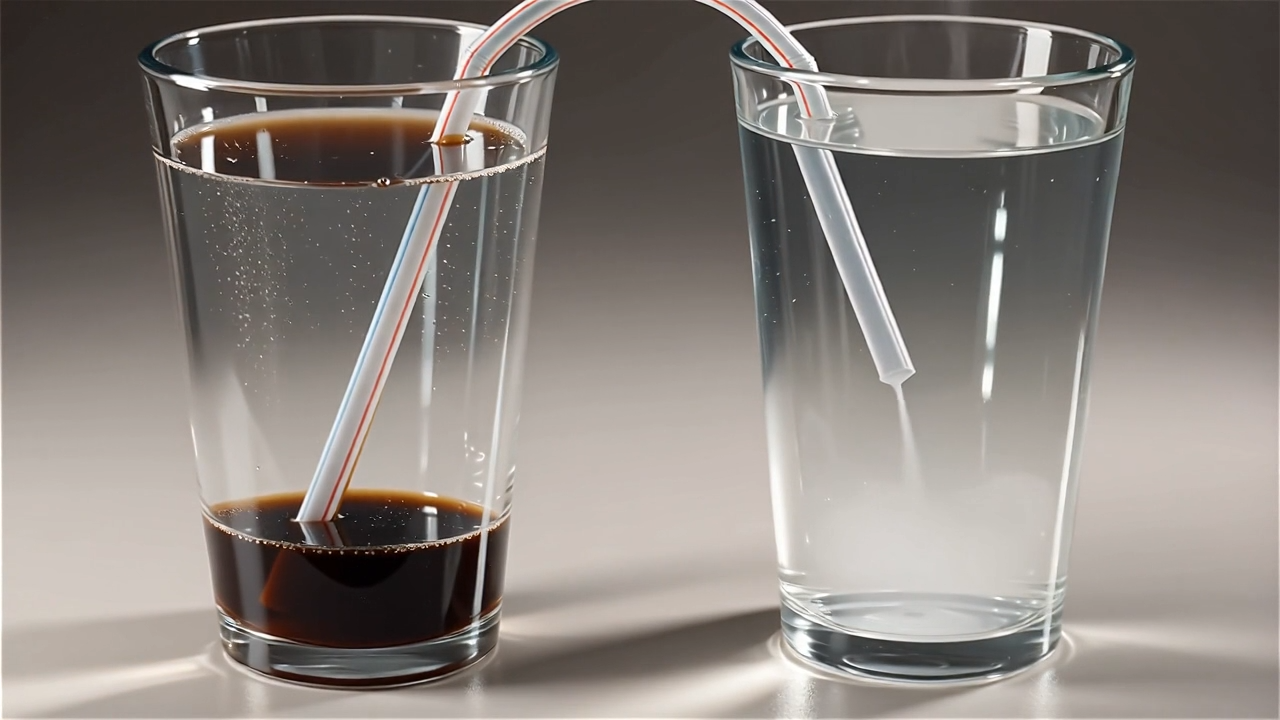}\hspace{1pt}%
  \includegraphics[height=1.3cm]{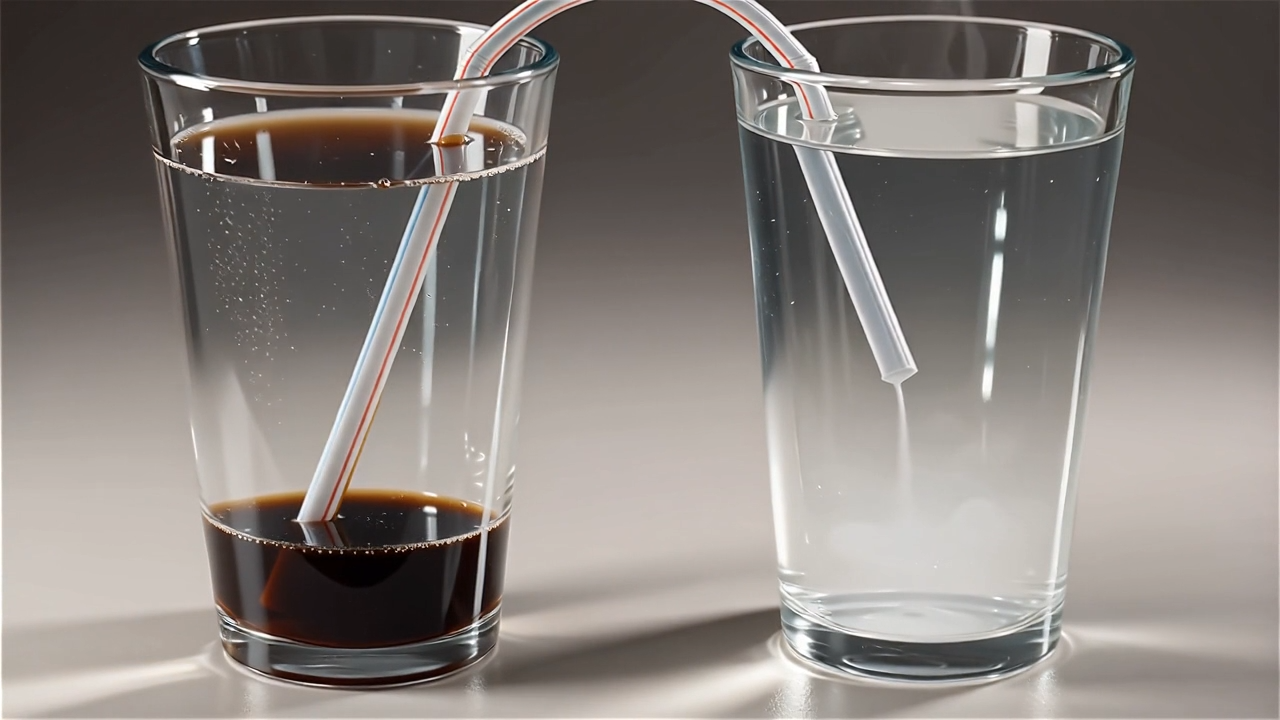}

}\\[0.2em]

\multicolumn{2}{c}{
  \textbf{PCS:} 3/4 \quad 
  \textbf{PCG:} 2/4 \quad 
  \textbf{CDN:} 3/4 \quad 
  \textbf{IMB:} 4/4 \quad 
  \textbf{STC:} 4/4
}\\
\bottomrule
\end{tabular}
\caption{In the generated video, the limewater doesn't turn white, and no bubbles appear. In the left glass (which should be transparent), bubbles appear suddenly and randomly. Human Annotation Rating: Prompt Consistency (PCS), Phenomenon Congruency (PCG), Correct Dynamism (CDN), Immutability (IMB), and Spatio-Temporal Coherence (STC).}
\label{fig:example120_t2v_wan2}
\end{figure*}

We present and discuss a few examples of T2V generation from Figure~\ref{fig:example175_t2v} to Figure~\ref{fig:example120_t2v_wan2}, where a prompt describing the experimental setup is provided to the video model. More examples can be found in supplementary materials.

\subsection{I2V Examples}
\begin{figure*}[htbp]
\centering
\setlength{\tabcolsep}{4pt}
\renewcommand{\arraystretch}{1.2}
\begin{tabular}{@{}p{0.15\textwidth}p{0.82\textwidth}@{}}
\toprule
\multicolumn{2}{c}{\textbf{Newton's Cradle}}\\
\midrule
\textbf{Prompt} & 
Five identical metal balls are suspended in a straight line by thin strings so that they hang just touching each other at rest. One ball on the end is pulled back and released, allowing it to swing and strike the others.\\
[0.3em]

\textbf{Expected} & 
When the lifted ball strikes the row, the ball on the opposite end swings outward while the others remain nearly stationary, demonstrating conservation of momentum and energy.\\
\midrule

\multicolumn{2}{c}{\cellcolor{red!10}\textbf{Veo-3:} {\color{red!70!black} Failed to Generate Expected Phenomenon}}\\[0.2em]

\multicolumn{2}{c}{
  \includegraphics[height=1.3cm]{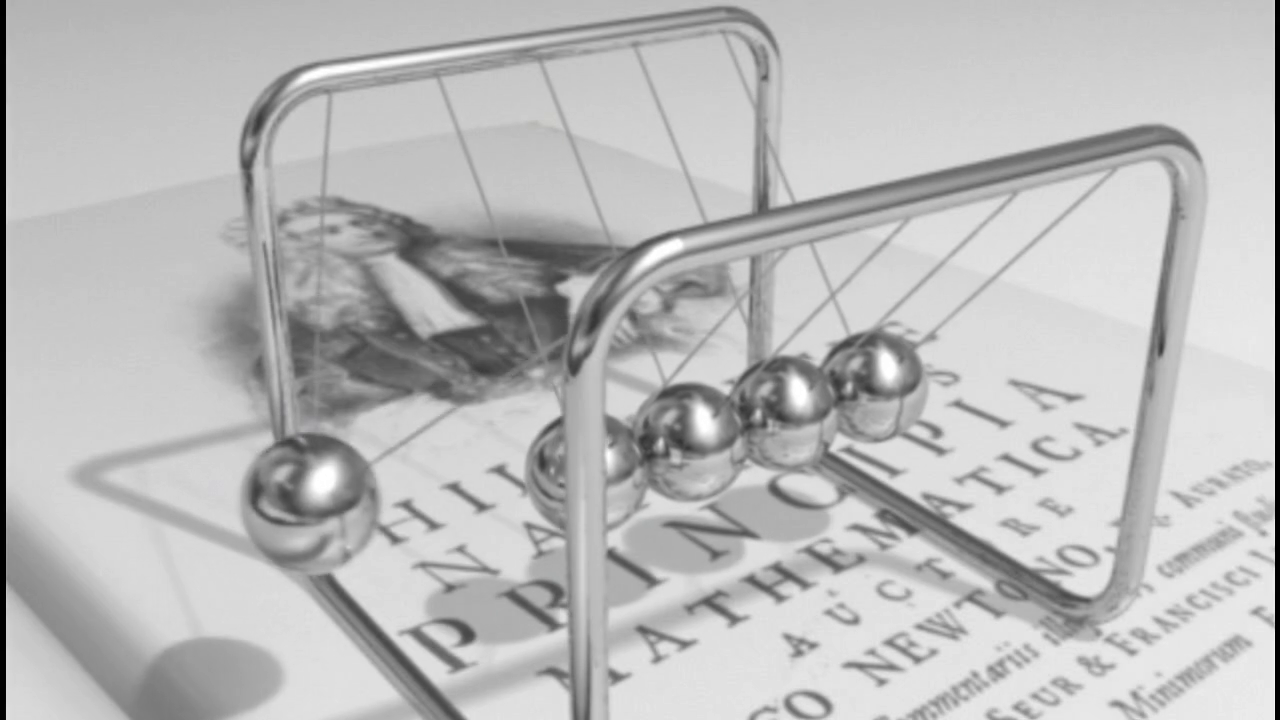}\hspace{1pt}%
  \hspace{0.25em}
 \raisebox{1.3em} {\Large$\boldsymbol{\longrightarrow}$}
  \hspace{0.25em}
  \includegraphics[height=1.3cm]{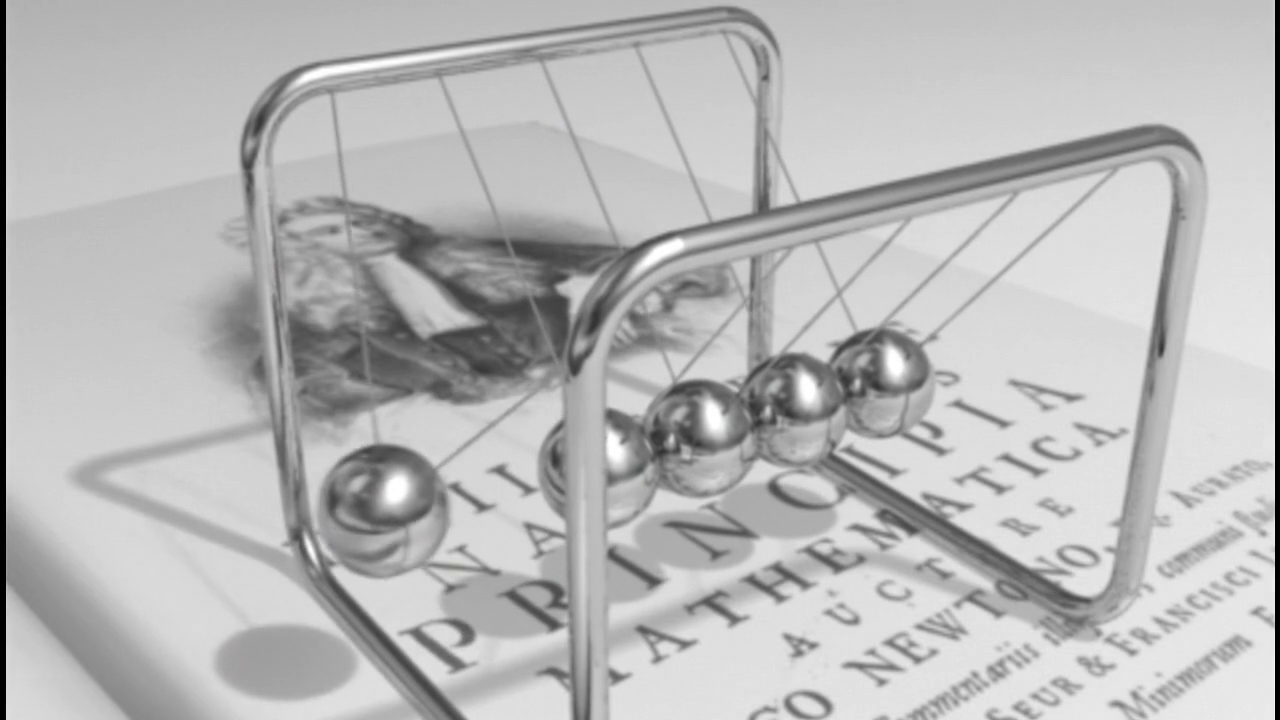}\hspace{1pt}%
  \includegraphics[height=1.3cm]{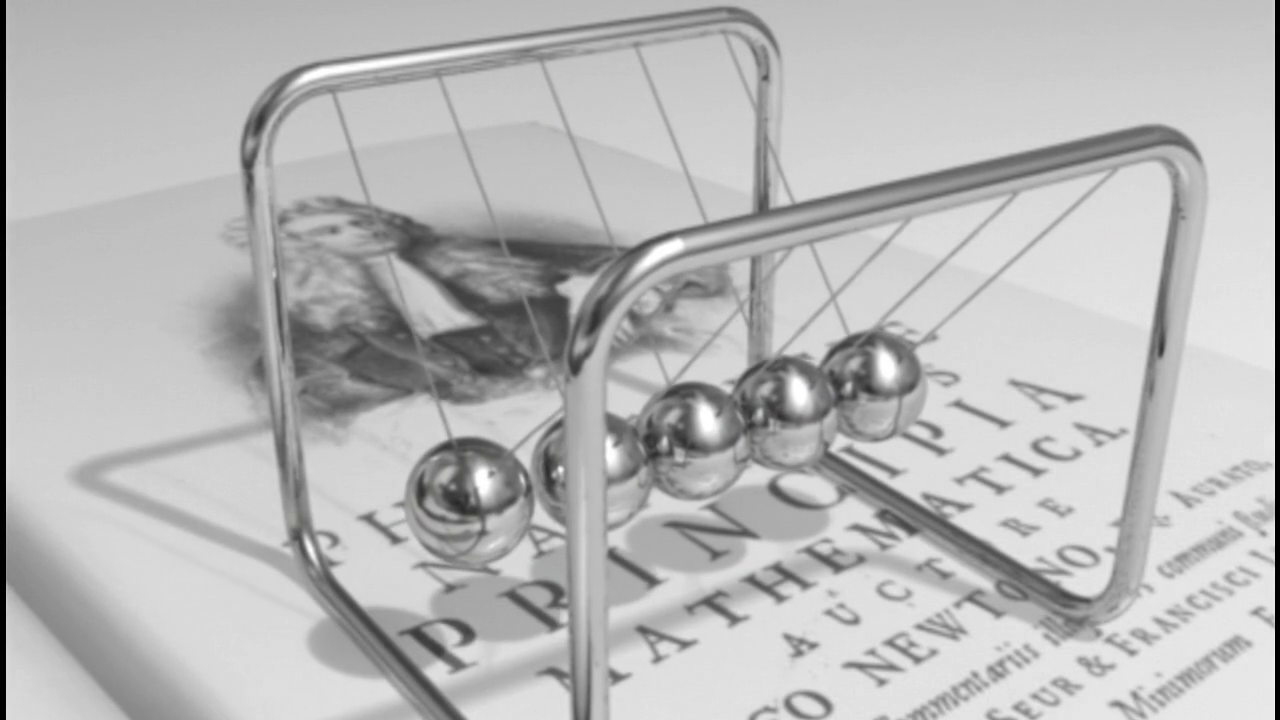}\hspace{1pt}%
  \includegraphics[height=1.3cm]{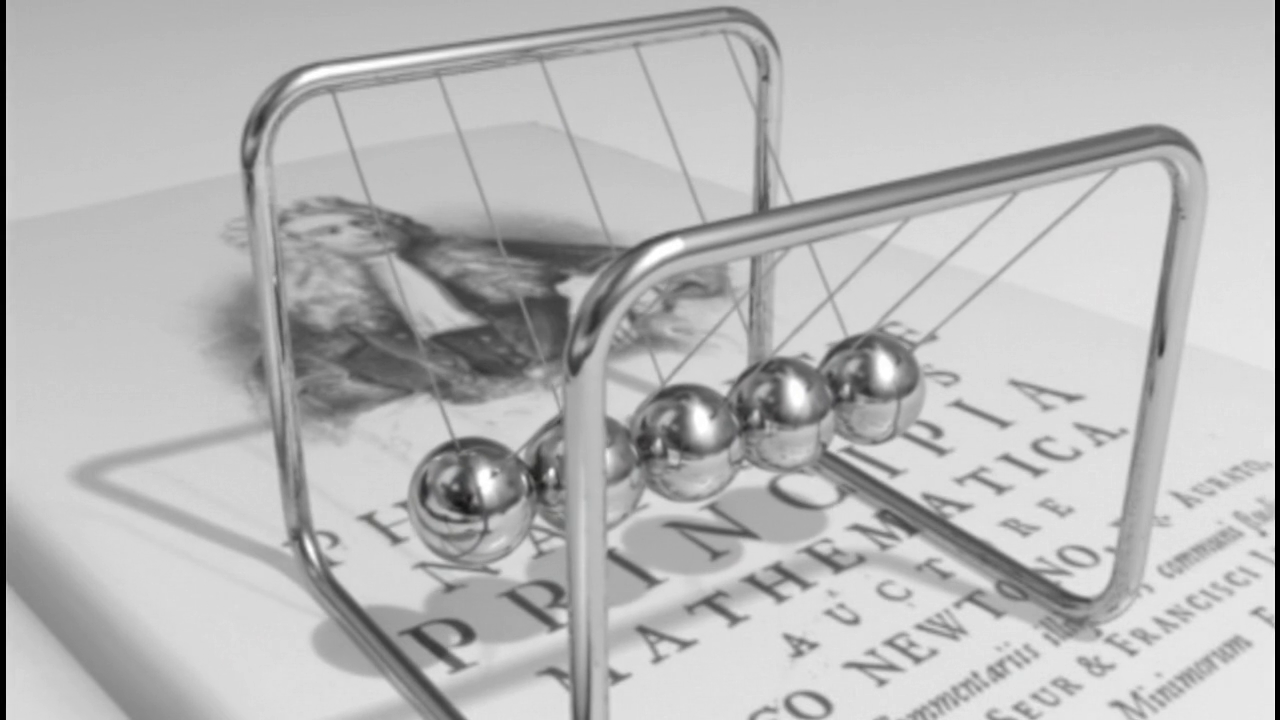}\hspace{1pt}%
  \includegraphics[height=1.3cm]{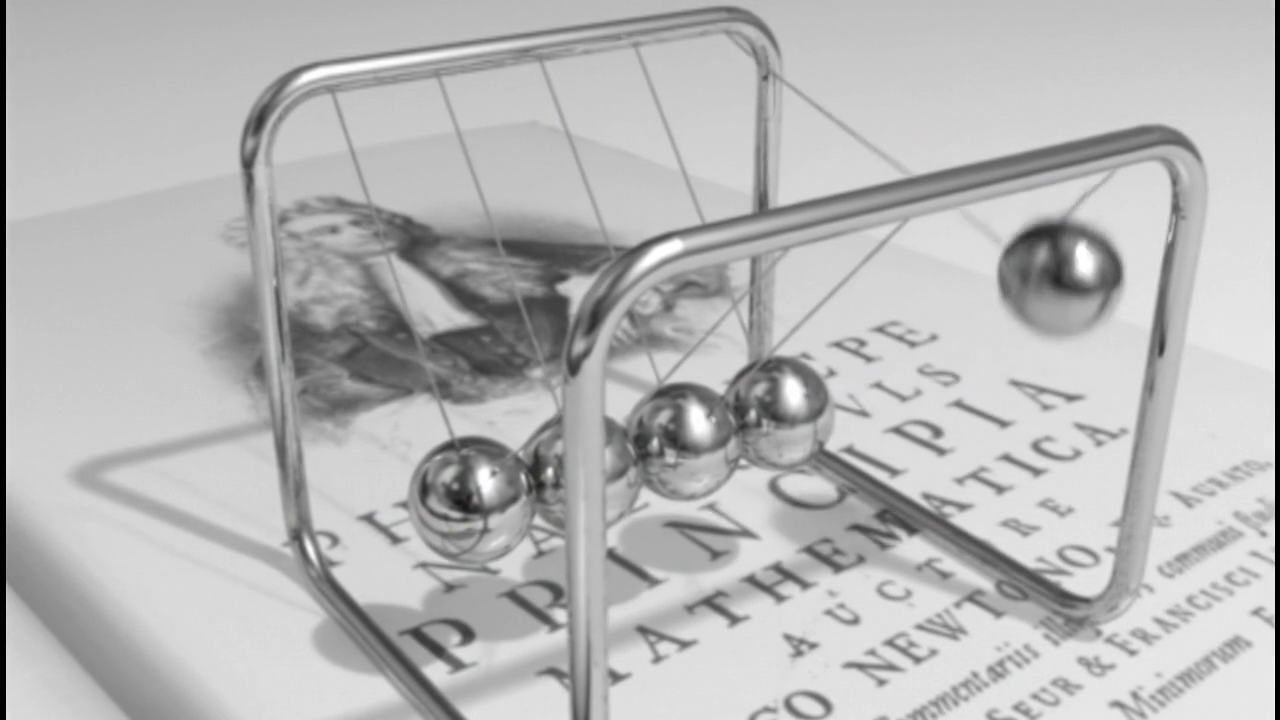}\hspace{1pt}%
  \includegraphics[height=1.3cm]{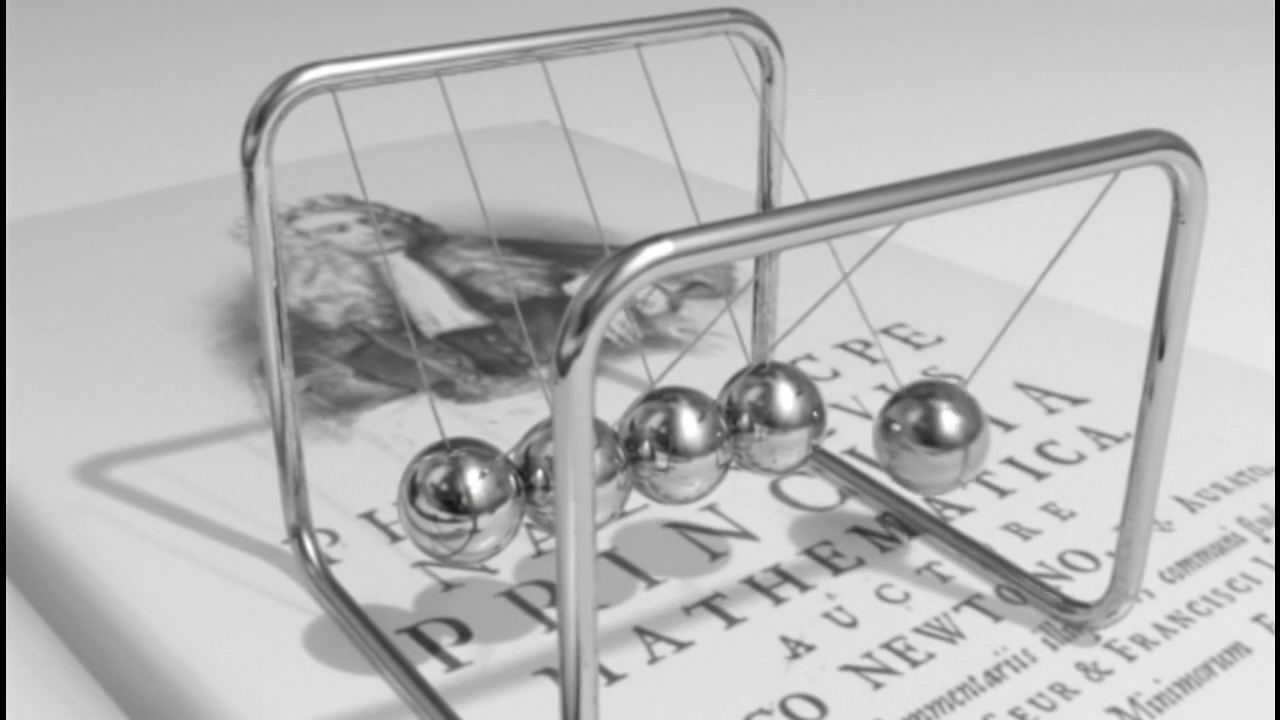}
  \includegraphics[height=1.3cm]{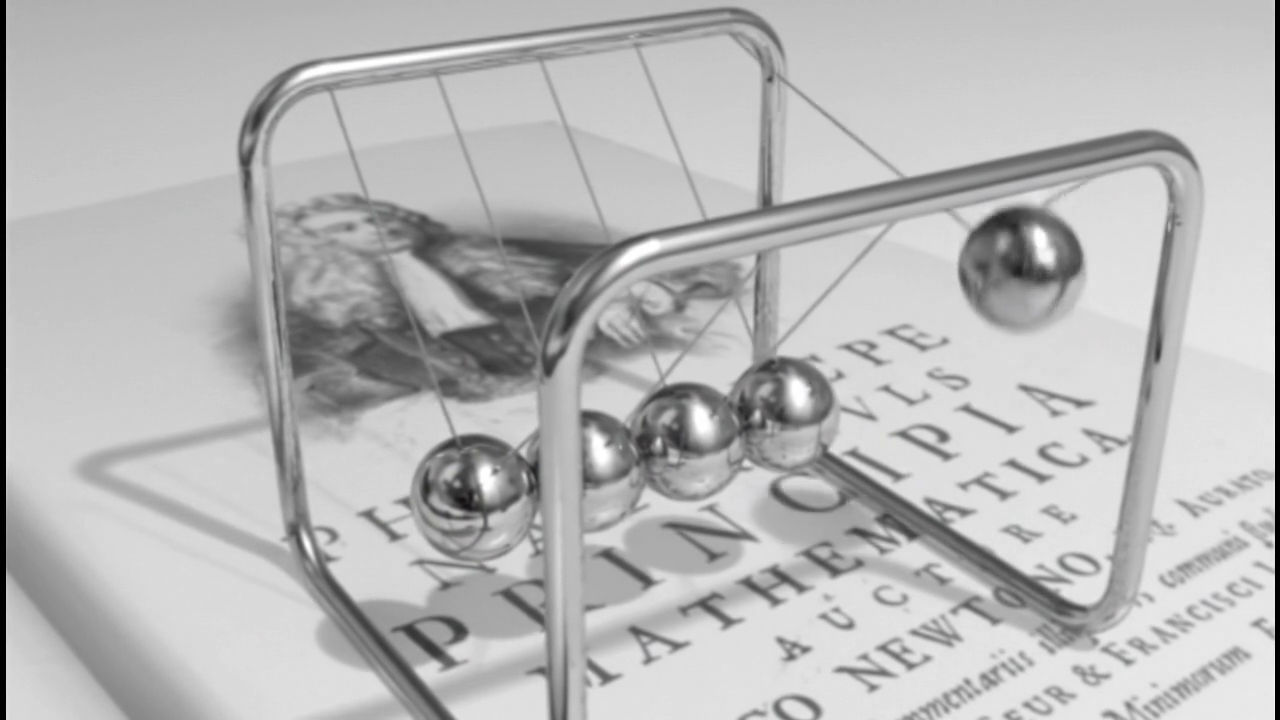}

}\\[0.2em]

\multicolumn{2}{c}{
  \textbf{PCS:} 4/4 \quad 
  \textbf{PCG:} 1/4 \quad 
  \textbf{CDN:} 1/4 \quad 
  \textbf{IMB:} 4/4 \quad 
  \textbf{STC:} 4/4
}\\
\bottomrule
\end{tabular}
\caption{Veo-3 generated video of Newton's Cradle conditioned on the first-frame image (from Wikipedia), where the video model fails to generate the expected phenomenon after the first strike to demonstrate conservation of energy.}
\label{fig:example_16_i2v}
\end{figure*}

We present and discuss one example of I2V generation in Figure~\ref{fig:example_16_i2v}, where both a prompt describing the experimental setup, and the first frame image are provided to the video model. More examples can be found in supplementary materials. 

% PROMPT FOR VLM-AS-A-JUDGE
\section{Prompts for VLM-as-a-Judge}
\label{appendix:prompt}

% -----------------  CUSTOMIZED  -----------------

% a common style that both boxes inherit
\tcbset{
  reasoningstyle/.style={
     colback=gray!5,
     colframe=purple,          
     coltitle=green,             
     fonttitle=\bfseries,      
     boxrule=1.2pt,   
     arc=4pt,              
     left=1em,right=1em,top=1em,bottom=1em,
  }
}
% -----------------  CUSTOMIZED  -----------------

\newtcolorbox{checklist-prompt}{
  reasoningstyle,
  title={Prompt for Checklist Generation}
}

\newtcolorbox{scoring-prompt}{
  reasoningstyle,
  title={Prompt for VLM as a Judge}
}

\subsection{Checklist Generation}
\label{appendix:prompt_checklist}

The checklist generation prompt template takes in the question prompt and the ground truth expected phenomenon to generate an itemized checklist for the VLM judge to pay attention to during the scoring process, and the VLM judge will be asked to reason and find to justify the scoring.

\begin{checklist-prompt}

From my evaluation of text2video models I have generated a video using the prompt\\
Prompt: \textcolor{blue}{\{{prompt\_text}\}}\\

\textcolor{blue}{\{{reference\_source}\}}\\

Use reference source (ground truth phenomenon and video reference if available) to create the checklist (mandatory).\\

Create a comprehensive checklist targeting the following categories (note: not all categories are required for a prompt).\\

1. PHENOMENON CONGRUENCY: Does the video show the correct expected phenomenon?\\

2. CORRECT DYNAMISM Are the physics dynamics and motion behaviors accurate?\\

3. SPATIO-TEMPORAL CONTINUITY Are spatial relationships and temporal sequences physically consistent?\\

4. IMMUTABILITY Do object properties remain physically consistent?\\

5.  INTERACTION REALISM Do object interactions follow physical laws?\\

Guidelines for checklist creation:\\
- only target things which are visually observable in the video\\
- the statements in checklist needs to be assertive statements instead of questions\\
\end{checklist-prompt}

\subsection{Scoring}
\label{appendix:prompt_judge}

The scoring prompt template follows the setup in Figure \ref{fig:judge_pipeline}, where the checklist and CV-based evidence table can be selectively enabled. The VLM judge is required to justify each rating using salient video frames, and when available, corresponding checklist items and CV-derived evidence.

\begin{scoring-prompt}

\textbf{\textcolor{blue}{RUBRIC\_TEXT}}\\
You are VLM-Judge evaluating a generated science video.\\
Score each rubric from 1–4 (1=absent/contradictory, 2=weak/partly wrong, 3=mostly correct, 4=clearly correct):\\
a) prompt\_consistency — follows instructions: correct setup and correct experiment execution.\\
b) expected\_phenomenon — expected physical/chemical outcome is present and correct.\\
c) immutability — objects remain intact/unchanged unless changes are explicitly expected.\\
d) dynamism — other physical laws are obeyed.\\
e) coherence — natural transitions across frames; no flicker/teleport/identity swap.\\

Each rating must be supported with clear justification, drawing on salient video frames and, when provided, the corresponding checklist items and CV-based evidence table.\\

\textbf{\textcolor{blue}{OUTPUT\_SCHEMA\_FORMAT}}\\
Return JSON with fields:\\
\{ 
    "scores": \{\\
    "prompt\_consistency":1-4,\\
    "expected\_phenomenon":1-4,\\
    "immutability":1-4,\\
    "dynamism":1-4,\\
    "coherence":1-4\\
  \},\\
  "explanations": \{"summary": string, "issues": [string]\},\\
  "evidence": \{"candidate": [\{"t":"0.0s","observation":""\}],\\
               "reference": [\{"t":"0.0s","observation":""\}]\}\\
\}\\

\textbf{\textcolor{blue}{Scoring Prompt:}}\\
Question description {question\_description}\\
Ground-truth phenomenon: {phenomenon}\\
\textcolor{blue}{\{{OPTIONAL\_CHECKLIST}\}}\\
\textcolor{blue}{\{{OPTIONAL\_CV\_BASED\_EVIDENCE\_TABLE}\}}\\
\textcolor{blue}{\{{RUBRIC\_TEXT}\}}\\
\textcolor{blue}{\{{OUTPUT\_SCHEMA\_FORMAT}\}}\\

\end{scoring-prompt}

\section{Baseline Evaluation Results}

\paragraph{PhyGenEval.}While both \sysnamenospace-Judge and PhyGenEval~\cite{meng2025phygenbench} address the critical challenge of evaluating physical commonsense in text-to-video models, they adopt complementary evaluation philosophies. PhyGenEval employs a hierarchical three-tier framework that progressively assesses key physical phenomena detection, physics order verification, and overall naturalness using multiple specialized VLMs with GPT-4o-generated questions. In contrast, \sysnamenospace-Judge grounds its evaluations in explicit, quantifiable evidence by integrating deterministic computer vision modules for entity verification, temporal coherence, motion analysis, and semantic alignment. They produce frame-level JSON records to support checklist-based deductions. This evidence-centric approach addresses the under-justified assessments and false-positive ratings we observed when directly prompting VLMs (Appendix~\ref{appendix:judge_model_choice}). Our evaluation of state-of-the-art models on \sysnamenospace using the PhyGenEval framework (Table~\ref{tab:phygenbench_baseline_results}) reveals that even the best-performing model, Veo-3, achieves only an overall score of 0.47. According to PhyGenEval scores, all the models perform better on physics than on chemistry-related topics.

Additionally, we also calculated $\tau$ and $\rho$ by giving equal weightage to all five dimensions (See section~\ref{para:rubrics}) but the $\tau$ and $\rho$ changed very insignificantly when compared to Table~\ref{tab:auto_eval_results}.

\begin{table}[!ht]
    \centering
    \small{
    \begin{tabular}{l c c c}
    \toprule
         Model  & Total & Physics & Chemistry \\ \midrule
        Ray2 & 0.29 & 0.29 & 0.26 \\ 
        Hailuo 2.3 & 0.30 & 0.31 & 0.22 \\ 
        Seedance 1.0 Pro & 0.33 & 0.34 & 0.29 \\ 
        Veo-3 & \textbf{0.47} & \textbf{0.47} & \textbf{0.43} \\
        Kling-v2.5-Turbo-Pro & 0.29 & 0.30 & 0.21 \\ 
        Wan-2.5-T2V-Preview & 0.41 & 0.42 & 0.35 \\ 
        Sora-2 & 0.43 & 0.43 & 0.40 \\ \bottomrule
    \end{tabular}}
    \caption{Physical commonsense evaluation scores across different models. Higher scores indicate better adherence to physical laws. Evaluated using the PhyGenEval framework \cite{meng2025phygenbench} on \sysnamenospace.}
    \label{tab:phygenbench_baseline_results}
\end{table}

\paragraph{T2VBench.}
We evaluate our full set of text-to-video models under the Consistent Attribute Binding metric from T2V-CompBench. Videos are generated with three independent runs per model, and Llava-1.5-7B is used as the automatic judge. The scores are normalized at the end for correlation analysis. The evaluation reveals normalized scores ranging from 0.75 to 0.8, with Veo-3.1 achieving the highest score and Ray2 achieving the lowest. We find that the automatic metric aligns well with human annotations, achieving moderate-to-strong rank correlation. These results suggest that although current models perform somewhat similarly on controlled attribute-binding tasks, SOTA models such as Veo-3 exhibit more stable object-attribute rendering, whereas models like Ray2 and Hailuo 2.3 struggle with maintaining attribute fidelity across frames for different runs. Refer to Table \ref{tab:t2v_baseline_results} for more details. 

\begin{table}[!ht]
    \centering
    \small % reduce font slightly to fit nicely
    \begin{tabular}{l|c|c}
    \hline
        Model & Mean Raw Score & Mean Normalized \\\hline
        Veo3 & \textbf{12.1998} & \textbf{0.7999} \\
        Wan-2.5-T2V-Preview & 12.0894 & 0.7921 \\
        Sora-2 & 12.0583 & 0.7899 \\
        Seedance 1.0 Pro & 11.9625 & 0.7830 \\
        Kling-v2.5-Turbo-Pro & 11.9450 & 0.7818 \\
        Hailuo 2.3 & 11.8021 & 0.7716 \\
        Ray2 & 11.6900 & 0.7636 \\ \hline
    \end{tabular}
    \caption{Consistent attribute binding evaluation scores across different models in descending order, with Veo3 achieving the highest score and Ray2 the lowest. Higher scores indicate more consistent object-attribute binding in the videos. Evaluated using the T2VBench framework \cite{sun2024t2vcompbench} on ScienceCompass.}
    \label{tab:t2v_baseline_results}
\end{table}

\paragraph{VideoScore2.}
We evaluate our full set of text-to-video models using the comprehensive multi-dimensional framework of VideoScore2. We report per-dimension scores and an overall aggregate (simple average across dimensions), min-max normalized to [0,1] for correlation analysis. This evaluation yields normalized aggregate scores ranging from 0.67 to 0.79, with Wan-2.5-T2V-Preview achieving the highest score and Ray2 achieving the lowest. To manage GPU memory constraints, we performed inference for VideoScore2 using half-precision float, a standard practice that prioritizes resource efficiency without substantially affecting the model's comparative ranking capability. A notable disparity is observed between these automated scores and our human annotations. This weak alignment is attributed to the fact that VideoScore2 was primarily designed to assess everyday common-sense plausibility and simple physical rules, capturing surface-level, visually observable anomalies. It does not require deeper scientific reasoning or evaluation of the non-trivial physical dynamics emphasized by our dataset. Refer to Table \ref{tab:vs2_baseline_results} for the full score breakdown.

% \begin{table}[!ht]
%     \centering
%     \small{
%     \begin{tabular}{c c c c}
%     \toprule
%         Model & PhyGenEval ($\uparrow$) & $\tau$ ($\uparrow$)& $\rho$ ($\uparrow$)\\ \midrule
%         Ray2 & 0.29 & 0.03 & 0.04 \\
%         Hailuo 2.3 & 0.29 & -0.05 & -0.06 \\
%         Seedance 1.0 Pro & 0.33 & -0.02 & -0.02 \\
%         Veo-3 & \textbf{0.47} & \textbf{0.15} & \textbf{0.19} \\ 
%         Kling-v2.5-Turbo-Pro & 0.29 & 0.05 & 0.06 \\
%         Wan-2.5-T2V-Preview & 0.41 & 0.11 & 0.14 \\
%         Sora-2 & 0.43 & 0.07 & 0.08 \\ \bottomrule
%     \end{tabular}}
%     \caption{Comparison of Physical Commonsense Evaluation scores across different models with human annotations. Evaluated using the PhyGenEval framework \cite{meng2024phygenbench} on \sysnamenospace-Bench. Kendall Tau ($\tau$) and Spearman correlations ($\rho$) measure alignment between PhyGenEval scores and human annotations.}
%     \label{tab:phygenbench_baseline_results}
% \end{table}

\begin{table*}[t]
\centering
\small
\begin{tabular}{@{}cccccc@{}}
\toprule
\textbf{Rank} & \textbf{Model} & \textbf{Visual Quality} & \textbf{Text–Video Alignment} & \textbf{Physical Consistency} & \textbf{Overall} \\
\midrule
1 & Wan-2.5-T2V-Preview & $0.93 \text{\scriptsize$\pm 0.18$}$ & $0.72 \text{\scriptsize$\pm 0.23$}$ & $0.73 \text{\scriptsize$\pm 0.27$}$ & $0.79 \text{\scriptsize$\pm 0.17$}$ \\
2 & Veo-3 & $0.90 \text{\scriptsize$\pm 0.18$}$ & $0.73 \text{\scriptsize$\pm 0.23$}$ & $0.70 \text{\scriptsize$\pm 0.28$}$ & $0.78 \text{\scriptsize$\pm 0.17$}$ \\
3 & Seedance 1.0 Pro & $0.92 \text{\scriptsize$\pm 0.19$}$ & $0.67 \text{\scriptsize$\pm 0.25$}$ & $0.73 \text{\scriptsize$\pm 0.27$}$ & $0.77 \text{\scriptsize$\pm 0.18$}$ \\
4 & Kling-v2.5-Turbo-Pro & $0.91 \text{\scriptsize$\pm 0.21$}$ & $0.68 \text{\scriptsize$\pm 0.24$}$ & $0.71 \text{\scriptsize$\pm 0.28$}$ & $0.77 \text{\scriptsize$\pm 0.19$}$ \\
5 & Hailuo 2.3 & $0.89 \text{\scriptsize$\pm 0.20$}$ & $0.60 \text{\scriptsize$\pm 0.25$}$ & $0.68 \text{\scriptsize$\pm 0.28$}$ & $0.72 \text{\scriptsize$\pm 0.18$}$ \\
6 & Sora-2 & $0.86 \text{\scriptsize$\pm 0.20$}$ & $0.65 \text{\scriptsize$\pm 0.23$}$ & $0.65 \text{\scriptsize$\pm 0.28$}$ & $0.72 \text{\scriptsize$\pm 0.18$}$ \\
7 & Ray2 & $0.85 \text{\scriptsize$\pm 0.21$}$ & $0.52 \text{\scriptsize$\pm 0.24$}$ & $0.64 \text{\scriptsize$\pm 0.28$}$ & $0.67 \text{\scriptsize$\pm 0.18$}$ \\
\bottomrule
\end{tabular}
\caption{Overall quantitative results of VideoScore2 (mean $\pm$ standard deviation).}
\label{tab:vs2_baseline_results}
\end{table*}

\section{\sysnamenospace-Judge Model Choice}
\label{appendix:judge_model_choice}

In Table~\ref{tab:sc_judge_gaps}, we compare the performance of \sysnamenospace-Judge when using a non-reasoning model (GPT-4o) versus a state-of-the-art reasoning model (GPT-5-Pro), both from OpenAI. The results reveal that GPT-4o’s ratings are heavily positively skewed, consistently overestimating video quality. Even after min–max normalization, its absolute deviations remain high, as much as 0.384 and 0.417 in dimensions like \emph{Coherence} and \emph{Immutabiilty}. Relacing 4o with GPT-5-pro reduces this discrepancy by up to 57\%, demonstrating employing strong reasoning model is critical to produce more careful, less inflated judgments.

\begin{table}[t]
\centering
\begin{tabular}{lcc}
\toprule
\textbf{Score dimension} & \textbf{VSci-Judge-4o} & \textbf{VSci-Judge-GPT-5-pro} \\
\midrule
Dynamism              & $+0.372\,\text{\scriptsize$\pm 0.129$}$ & $+0.315\,\text{\scriptsize$\pm 0.131$}$  \\
Prompt consistency    & $+0.223\,\text{\scriptsize$\pm 0.120$}$ & $-0.096\,\text{\scriptsize$\pm 0.098$}$ \\
Phenomenon congruency & $+0.125\,\text{\scriptsize$\pm 0.076$}$ & $-0.218\,\text{\scriptsize$\pm 0.103$}$ \\
Immutability          & $+0.417\,\text{\scriptsize$\pm 0.145$}$ & $+0.273\,\text{\scriptsize$\pm 0.120$}$ \\
Coherence             & $+0.384\,\text{\scriptsize$\pm 0.091$}$ & $+0.278\,\text{\scriptsize$\pm 0.083$}$ \\
\bottomrule
\end{tabular}
\caption{Gap to human annotations (mean $\Delta \pm$ std) for each score dimension and VSci-Judge variants using 4o and GPT-5 pro respectively.}
\label{tab:sc_judge_gaps}
\end{table}

\section{Details of Expert Annotation Results on \sysnamenospace-Bench}
\label{appendix:annotation_scores}

We present a detailed human annotation rating breakdowns in Table~\ref{tab:raw_human_annotation_scores_appendix} and normalized rating in Table~\ref{tab:normalized_annotation_scores}, where min-max normalization is applied to each rating instance before averaging them across runs and test instances for each model.

\begin{table}[htbp]
\centering
\small
\begin{tabular}{lccccc}
\toprule
\textbf{Model} &
\textbf{Prompt Consistency} &
\textbf{Phenomenon Congruency} &
\textbf{Dynamism} &
\textbf{Immutability} &
\textbf{Coherence} \\
\midrule
 Sora-2    & $3.32\,\text{\scriptsize$\pm 0.91$}$ & $2.56\,\text{\scriptsize$\pm 1.10$}$ & $3.33\,\text{\scriptsize$\pm 0.96$}$ & $3.73\,\text{\scriptsize$\pm 0.63$}$ & $3.71\,\text{\scriptsize$\pm 0.71$}$ \\
 Veo-3       & $3.01\,\text{\scriptsize$\pm 1.01$}$ & $2.35\,\text{\scriptsize$\pm 1.09$}$ & $2.83\,\text{\scriptsize$\pm 1.14$}$ & $3.30\,\text{\scriptsize$\pm 0.98$}$ & $3.42\,\text{\scriptsize$\pm 0.90$}$ \\
Kling-v2.5-Turbo-Pro  & $2.77\,\text{\scriptsize$\pm 1.06$}$ & $1.91\,\text{\scriptsize$\pm 1.03$}$ & $2.75\,\text{\scriptsize$\pm 1.06$}$ & $3.36\,\text{\scriptsize$\pm 0.94$}$ & $3.60\,\text{\scriptsize$\pm 0.72$}$ \\
Wan-2.5-T2V-Preview     & $2.87\,\text{\scriptsize$\pm 1.02$}$ & $1.84\,\text{\scriptsize$\pm 1.03$}$ & $2.83\,\text{\scriptsize$\pm 1.07$}$ & $3.36\,\text{\scriptsize$\pm 0.93$}$ & $3.46\,\text{\scriptsize$\pm 0.89$}$ \\
Seedance 1.0 Pro & $2.56\,\text{\scriptsize$\pm 1.04$}$ & $1.78\,\text{\scriptsize$\pm 1.03$}$ & $2.52\,\text{\scriptsize$\pm 1.13$}$ & $3.15\,\text{\scriptsize$\pm 1.06$}$ & $3.46\,\text{\scriptsize$\pm 0.84$}$ \\
Hailuo 2.3  & $2.39\,\text{\scriptsize$\pm 1.10$}$ & $1.67\,\text{\scriptsize$\pm 0.98$}$ & $2.57\,\text{\scriptsize$\pm 1.18$}$ & $3.16\,\text{\scriptsize$\pm 1.11$}$ & $3.46\,\text{\scriptsize$\pm 0.82$}$ \\
Ray2      & $1.65\,\text{\scriptsize$\pm 0.85$}$ & $1.26\,\text{\scriptsize$\pm 0.57$}$ & $2.13\,\text{\scriptsize$\pm 1.06$}$ & $2.44\,\text{\scriptsize$\pm 1.23$}$ & $2.92\,\text{\scriptsize$\pm 1.13$}$ \\
\bottomrule
\end{tabular}
\caption{Raw human annotation scores (mean $\pm$ std) per model and score dimension (1--4 Likert scale).}
\label{tab:raw_human_annotation_scores_appendix}
\end{table}

\begin{table}[ht]
\centering
\small
\begin{tabular}{lccccc}
\toprule
\textbf{Model} &
\textbf{Prompt Consistency} &
\textbf{Phenomenon Congruency} &
\textbf{Dynamism} &
\textbf{Immutability} &
\textbf{Coherence} \\
\midrule
 Sora-2      & $0.796$ & $0.697$ & $0.775$ & $0.818$ & $0.732$ \\
 Veo-3        & $0.677$ & $0.586$ & $0.551$ & $0.638$ & $0.601$ \\
Kling-v2.5-Turbo-Pro  & $0.567$ & $0.419$ & $0.530$ & $0.661$ & $0.691$ \\
Wan-2.5-T2V-Preview      & $0.602$ & $0.384$ & $0.558$ & $0.663$ & $0.614$ \\
Seedance 1.0 Pro  & $0.507$ & $0.354$ & $0.438$ & $0.571$ & $0.616$ \\
Hailuo 2.3   & $0.425$ & $0.298$ & $0.462$ & $0.602$ & $0.610$ \\
Ray2       & $0.129$ & $0.131$ & $0.268$ & $0.302$ & $0.385$ \\
\bottomrule
\end{tabular}
\caption{Human annotation scores (normalized) per model and per score dimension.}
\label{tab:normalized_annotation_scores}
\end{table}

\end{document}